\documentclass[twoside,11pt]{article}
\usepackage{jmlr2e}

\pdfoutput=1

\usepackage{mycommands1,amsmath,color,outlines,cite,subfigure}
\usepackage[small]{caption}
\usepackage{stackrel}
\usepackage[utf8x]{inputenc} 

\DeclareMathOperator*{\ve}{vec}
\DeclareMathOperator*{\diag}{diag }
\DeclareMathOperator*{\supp}{supp }
\DeclareMathOperator*{\Tr}{Tr}
\DeclareMathOperator*{\argmin}{arg\,min}

\DeclareMathOperator*{\Th}{^{\text{th}}}

\usepackage{chngcntr}
\usepackage{apptools}
\hypersetup{colorlinks = true, citecolor = blue, urlcolor = blue}

\begin{document}

\ShortHeadings{Joint Multiple Multi-layered Gaussian Graphical Models}{Majumdar and Michailidis}
\firstpageno{1}

\newtheorem{Algorithm}{Algorithm}

\title{Joint Estimation and Inference for Data Integration Problems based on Multiple Multi-layered Gaussian Graphical Models}
\date{}
\author{\name Subhabrata Majumdar\thanks{Currently in Splunk.} \email smajumdar@splunk.com \\
       \addr University of Florida Informatics Institute\\
       Gainesville, FL, 32611, USA
      \AND
      \name George Michailidis\thanks{Corresponding Author. Post Address: 205 Griffin Floyd Hall, 1 University Ave, Gainesville, FL, 32611.} \email gmichail@ufl.edu \\
      \addr Department of Statistics and Computer \& Information Science \& Engineering \\
      University of Florida \\
      Gainesville, FL 32611, USA 
       }

\editor{}
\maketitle

\begin{abstract} 
The rapid development of high-throughput technologies has enabled the generation of data from biological or disease processes that span multiple layers, like genomic, proteomic or metabolomic data, and further pertain to multiple sources, like disease subtypes or experimental conditions. In this work, we propose a general statistical framework based on Gaussian graphical models for horizontal (i.e. across conditions or subtypes) and vertical (i.e. across different layers containing data on molecular compartments) integration of information in such datasets. We start with decomposing the multi-layer problem into a series of two-layer problems. For each two-layer problem, we model the outcomes at a node in the lower layer as dependent on those of other nodes in that layer, as well as all nodes in the upper layer. We use a combination of neighborhood selection and group-penalized regression to obtain sparse estimates of all model parameters. Following this, we develop a debiasing technique and asymptotic distributions of inter-layer directed edge weights that utilize already computed neighborhood selection coefficients for nodes in the upper layer. Subsequently, we establish global and simultaneous testing procedures for these edge weights. Performance of the proposed methodology is evaluated on synthetic and real data.
\end{abstract}

\vspace{1em}

\section{Introduction}

Aberrations in complex biological systems develop in the background of diverse genetic and environmental factors and are associated with multiple complex molecular events. These include changes in the genome, transcriptome, proteome and metabolome, as well as epigenetic effects. Advances in high-throughput profiling techniques have enabled a systematic and comprehensive exploration of the genetic and epigenetic basis of various diseases, including cancer \citep{lee2016integrated, kaushik2016inhibition}, diabetes \citep{yuan2014integrated,sas2018shared}, chronic kidney disease \citep{atzler2014integrated}, etc. Further, such multi-Omics collections have become available for patients belonging to different, but related disease subtypes, with The Cancer Genome Atlas (TCGA: \citet{Tomczak15}) being a prototypical one. Hence, there is an increasing need for models that can {\em integrate} such complex data both {\em vertically} across multiple modalities and {\em horizontally} across different disease subtypes.

\begin{figure}
\centering
\includegraphics[]{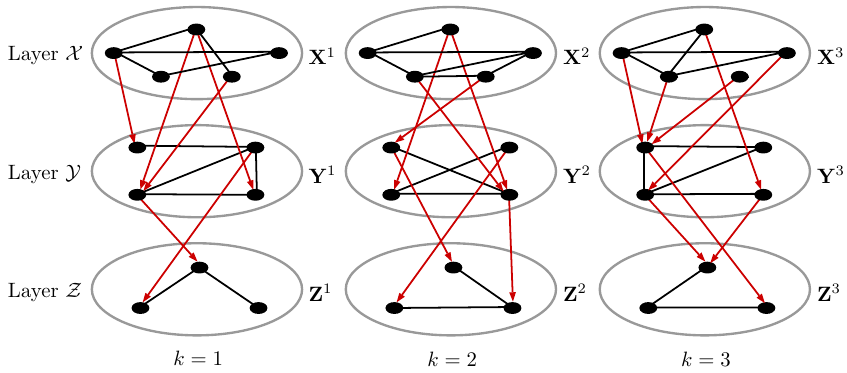}
\caption{Multiple multilayer graphical models. The matrices $(\bfX^k, \bfY^k, \bfZ^k), k = 1,2,3$ indicate data for each layer and category $k$. Within-layer connections (black lines) are undirected, while between-layer connections (red lines) go from an upper layer to the successive lower layer. For each type of edges (i.e. within $\cX, \cY, \cZ$ and $\cX \rightarrow \cY, \cY \rightarrow \cZ$), there are common edges across some or all $k$. }
\label{fig:multi2layer}
\end{figure}

Figure~\ref{fig:multi2layer} provides a schematic representation of the horizontal and vertical structure of such heterogeneous multi-modal Omics data as outlined above. A simultaneous analysis of all components in this complex layered structure has been coined in the literature as {\it data integration}. While it is common knowledge that this will result in a more comprehensive picture of the regulatory mechanisms behind diseases, phenotypes and biological processes in general, there is a dearth of rigorous methodologies that satisfactorily tackle all challenges that stem from attempts to perform data integration \citep{JoycePalsson06,GomezCabreroEtal14,GligPrzulj15}. A review of the present approaches towards achieving this goal, which are based mostly on specific case studies, can be found in \citet{GligPrzulj15} and \citet{ZhangOuyangZhao17}.

\paragraph{Contributions}
of this paper are two-fold. Firstly, we propose an integrative framework to conduct simultaneous inference for all parameters in {\it multiple} and {\it multi-layer} graphical models, essentially formalizing the structure in Figure~\ref{fig:multi2layer}. We decompose the multi-layer problem into a series of two-layer problems, propose an estimation algorithm for them based on group penalization, and derive theoretical properties of the estimators. Generalizing to group structures on the model parameters allows us to incorporate prior information, \textit{as and when available}, on within-layer or between-layer sub-graph components shared across \textit{some or all} $k=1,\cdots,K$. For biological processes, such information can stem from experimental or mechanistic knowledge (for example a pathway-based grouping of genes). Secondly, we obtain \textit{debiased} versions of within-layer regression coefficients in this two-layer model, and derive their asymptotic distributions using estimates of model parameters that satisfy generic convergence guarantees. Subsequently, we formulate a global test, as well as a simultaneous testing procedure that controls for False Discovery Rate (FDR) to detect important pairwise differences among directed edges between layers.

The novel techniques developed are based on a small number of technical assumptions that are quite general. For example, the model quantities used in our global testing procedure do not necessarily need to be sparse, and instead are only required to have $O(\sqrt{ \text{parameter dimension}/n})$ finite sample error bounds that have become standard in the high-dimensional literature (for example see \citet{LohWainwright12, BasuMichailidis15, BasuEtal19}). The advantage of this 
fiexibility is that components can be switched out to adapt the framework to other technical assumptions. The optional group sparsity assumptions in our estimation technique can be replaced by other structural restrictions (or no restrictions), for example low-rank or low-rank-plus-sparse, as deemed appropriate by the prior dependency assumptions across parameters. As long as these resulting estimates converge to the true parameters at the specified finite-sample rates, they can be used by the developed testing methodology.

\paragraph{Related work}
Gaussian Graphical Models (GGM) have been extensively used to model biological networks in the last few years. While the initial work on GGMs focused on estimating undirected edges within a single network through obtaining sparse estimates of the inverse covariance matrix from high-dimensional data (e.g. see references in \citet{BuhlmannvandeGeer11}), attention has shifted to estimating parameters from more complex structures. This includes (1) analyzing multiple related but not identical graphical models simultaneously, and (2) stacking up more multiple graphical models to form hierarchical multilayer networks, with both directed and undirected edges. For the first class of problems, \citet{GuoEtal11} and \citet{XieLiuValdar16} assumed perturbations over a common underlying structure to model multiple precision matrices, while \citet{DanaherEtal14} proposed using fused/group lasso type penalties for the same task. To incorporate prior information on the group structures across several graphs, \citet{MaMichailidis15} proposed the Joint Structural Estimation Method (JSEM), which uses group-penalized neighborhood regression and subsequent refitting for estimating precision matrices.
For the second problem, a two-layered structure can be modeled by interpreting directed edges between the two layers as elements of a multitask regression coefficient matrix, while undirected edges inside either layer correspond to the precision matrix of predictors in that layer. While several methods exist in the literature for joint estimation of both sets of parameters \citep{LeeLiu12, CaiEtal12}, only recently \citet{LinEtal16} made the observation that a multi-layer model can, in fact, be decomposed into a series of two-layer problems. Subsequently, they proposed an estimation algorithm and derived theoretical properties of the resulting estimators.

All the above approaches focus either on the horizontal or the vertical dimensions of the full hierarchical structure depicted in Figure~\ref{fig:multi2layer}. Hence, multiple related groups of heterogeneous data sets have to be modeled by analyzing all data in individual layers (i.e. models for $\{ \bfX^k \}$, $\{ \bfY^k \}$, $\{ \bfZ^k \}$), and then separately analyzing individual hierarchies of datasets (i.e. separate models for $(\bfX^k, \bfY^k, \bfZ^k), k = 1,2,3$). In another line of work, \citet{KlingEtal15,ZhangOuyangZhao17} model all undirected edges within all nodes together using penalized log-likelihoods. The advantage of this approach is that it can incorporate feedback loops and connections between nodes in non-adjacent layers. However, it has two considerable caveats. Firstly, it does not distinguish between hierarchies, hence delineating the direction of a connection between two nodes across two
different Omics modalities is not possible in such models. Secondly, computation becomes difficult when data from different Omics modalities are considered, since the number of estimable parameters increases at a faster late compared to a hierarchical model.

While there has been some progress for parameter estimation in multilayer models, little is known about the sampling distributions of resulting estimates. Current research on such distributions and related testing procedures for estimates from high-dimensional problems has been limited to single-response regression using lasso \citep{ZhangZhang14,JavanmardMontanari14,JavanmardMontanari18,vanDeGeerEtal14} or group lasso \citep{MitraZhang16} penalties, and partial correlations of single \citep{CaiLiu16} or multiple \citep{BelilovskyEtal16,Liu17} GGMs. From a systemic perspective, testing and identifying downstream interactions that differ across experimental conditions or disease subtypes can offer important insights on the underlying biological process \citep{MaoEtal17,LiEtal15}. In our proposed integrative framework, this can be accomplished by developing a hypothesis testing procedure for entries in the within-layer regression matrices.

\paragraph{Organization of paper}
We start with the model formulation in Section~\ref{sec:sec2}, then introduce our computational algorithm for a two-layer model, and derive theoretical convergence properties of the algorithm and resulting estimates. In section~\ref{sec:sec3}, we start by introducing the debiased versions of rows of the regression coefficient matrix estimates in our model, then use already computed parameter estimates that satisfy some general consistency conditions to obtain its asymptotic distribution. We then move on to pairwise testing, and use sparse estimates from our algorithm to propose a global test to detect overall differences in rows of the coefficient matrices, as well as a multiple testing procedure to detect elementwise differences and perform within-row thresholding of estimates in presence of moderate misspecification of the group sparsity structure. Sections~\ref{sec:sec4} and \ref{sec:secreal} are devoted to implementation of our methodology. In Section~\ref{sec:sec4}, we evaluate the performance of our estimation and testing procedure through several simulation settings, and give strategies to speed up the computational algorithm for high data dimensions. Section~\ref{sec:sec5} presents a real data example, where we illustrate how the application of our framework leads to knowledge discovery in complex biological networks. We conclude the paper with a discussion in Section~\ref{sec:secreal}. Proofs of all theoretical results, as well as some auxiliary results, are given in the Appendix.

\paragraph{Notation}
We denote scalars by small letters, vectors by bold small letters and matrices by bold capital letters. For any matrix $\bfA$, $(\bfA)_{ij}$ denote its element in the $(i,j)\Th$ position. For $a,b \in \BN$, we denote the set of all $a \times b$ real matrices by $\BM(a,b)$. For a positive semi-definite matrix $\bfP$, we denote its smallest and largest eigenvalues by $\Lambda_{\min} (\bfP)$ and $\Lambda_{\max} (\bfP)$, respectively. For any positive integer $c$, define $\cI_c = \{ 1, \ldots, c\}$. For vectors $\bfv$ and matrices $\bfM$, $\| \bfv \|$, $\|\bfv \|_1$ or $\|\bfM \|_1$ and $\|\bfv \|_\infty$ or $\|\bfM \|_\infty$ denote euclidean, $\ell_1$ and $\ell_\infty$ norms, respectively. The notation $\supp(\bfA)$ indicates the non-zero edge set in a matrix (or vector) $\bfA$, i.e. $\supp(\bfA) = \{(i,j): (\bfA)_{ij} \neq 0\}$. For any set $\cS$, $| \cS|$ denotes the number of elements in that set. For positive real numbers $A, B$ we write $A \succsim B$ if there exists $c>0$ independent of model parameters such that $A \geq cB$. We use the `$:=$' notation to define a quantity for the first time.
\section{The Joint Multiple Multilevel Estimation Framework}
\label{sec:sec2}

\subsection{Formulation}
Suppose there are $K$ independent data sets, each pertaining to an $M$-layered Gaussian Graphical Model (GGM) that has $p_m$ nodes in the $m$-th layer ($1 \leq m \leq M$). The $k^{\Th}$ model has the following structure:

\vspace{1em}
\begin{tabular}{ll}
{\it Layer 1}- &
$\BD_1^k = (D_{1 1}^k, \ldots, D^k_{1 p_1}) \sim
\cN (0, \Sigma_1^k); \quad k \in \cI_K,$\\
{\it Layer $m$} $(1< m \leq M)$-  &
$ \BD_m^k = \BD_{m-1}^k \bfB_m^k + \BE_m^k$, with $\bfB_m^k \in \BM(p_{m-1}, p_m) $\\
& and $\BE_m^k = (E_{m 1}^k, \ldots, E^k_{m p_m}) \sim
\cN (0, \Sigma_m^k); \quad k \in \cI_K $.\\
\end{tabular}
\vspace{1em}

\noindent In addition, information is available on horizontal (across $k$) or vertical (across $m$) dependencies among nodes within a layer or between nodes of adjacent layers. These are represented by known structured sparsity (i.e. grouping) patterns, denoted by $\cG_m$ and $\cH_m$, for the parameters of interest in the above model, i.e. the precision matrices $\Omega_m^k := (\Sigma_m^k)^{-1}$ and the regression coefficient matrices $\bfB_m^k$. Our goal is to leverage this side-information to estimate the full hierarchical structure of the network- specifically to obtain the undirected edges for the nodes inside a single layer, and the directed edges between two successive layers through jointly estimating $\{ \Omega_m^k \}$ and $\{ \bfB_m^k \}$.

Next, consider a two-layer model, which is a special case of the above model with $M=2$:
\begin{eqnarray}
\BX^k = (X^k_1, \ldots, X^k_p)^T \sim \cN (0, \Sigma^k_{x});\\
\BY^k = \BX^k \bfB^k + \BE^k; \quad \BE^k = (E^k_1, \ldots, E^k_p)^T \sim \cN (0, \Sigma^k_{y});\\
\bfB^k \in \BM(p,q), \quad \Omega^k_{x} = (\Sigma^k_{x})^{-1}; \quad \Omega^k_y = (\Sigma^k_{y})^{-1};
\end{eqnarray}
wherein we want to estimate $\{ (\Omega^k_{x}, \Omega^k_{y}, \bfB^k); k \in \cI_K \}$ from data $\cZ^k = \{ (\bfY^k, \bfX^k); \bfY^k \in \BM(n,q), \bfX^k \in \BM(n,p), k \in \cI_K\}$ in presence of known grouping structures $\cG_x, \cG_y, \cH$ respectively and assuming $n_k = n$ for all $k \in \cI_K$ for simplicity. We focus the theoretical discussion in the remainder of the paper on jointly estimating $\Omega_{y}:= \{ \Omega_{y}^k \}$ and $\cB := \{ \bfB^k \}$. This is because for $M>2$, within-layer undirected edges of any $m{\Th}$ layer $(m>1)$ and between-layer directed edges from the $(m-1){\Th}$ layer to the $m{\Th}$ layer can be estimated from the corresponding data matrices in a similar fashion (see details in \citet{LinEtal16}). On the other hand, parameters in the very first layer are analogous to $\Omega_{x} := \{ \Omega_{x}^k \}$, and can be estimated from $\{ \bfX^k\}$ using any method for joint estimation of multiple graphical models (e.g. \citet{GuoEtal11, MaMichailidis15}). This provides all building blocks for recovering the full hierarchical structure of our $M$-layered multiple GGMs.


\subsection{Algorithm}
\label{sec:algosection}

\begin{figure}
\centering
\includegraphics[width=.23\textwidth]{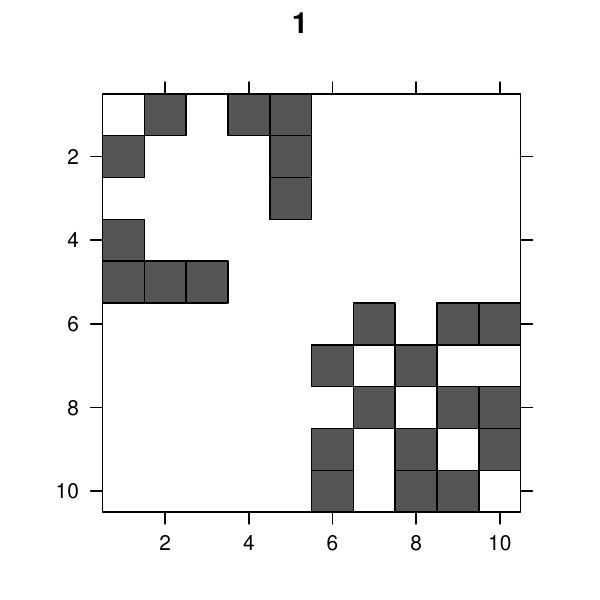}
\includegraphics[width=.23\textwidth]{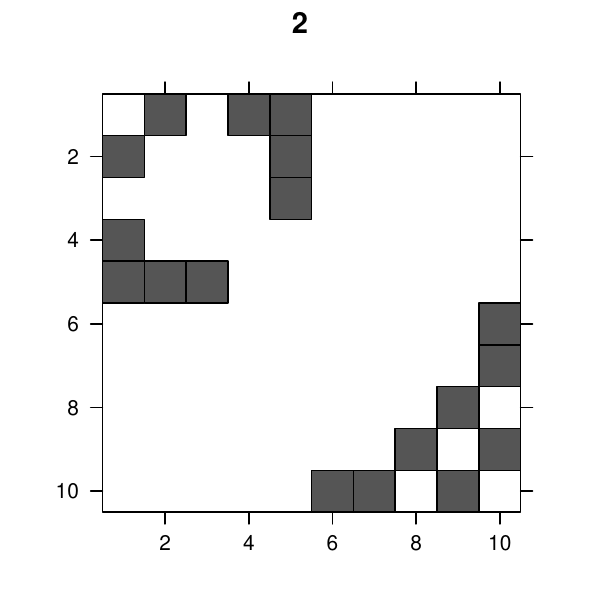}
\includegraphics[width=.23\textwidth]{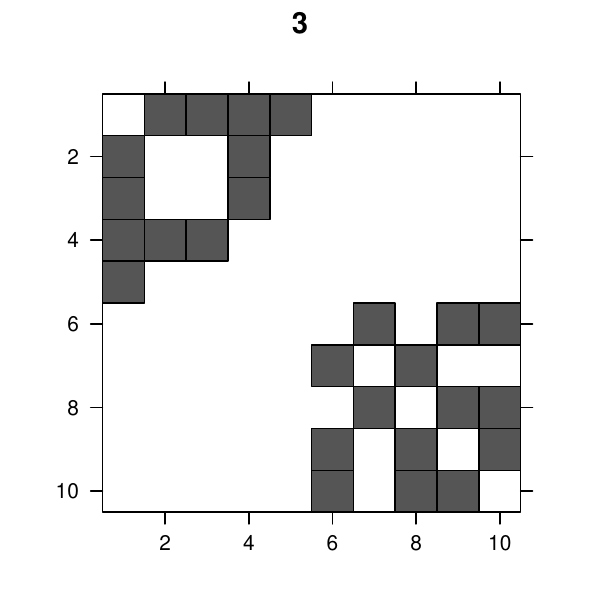}
\includegraphics[width=.23\textwidth]{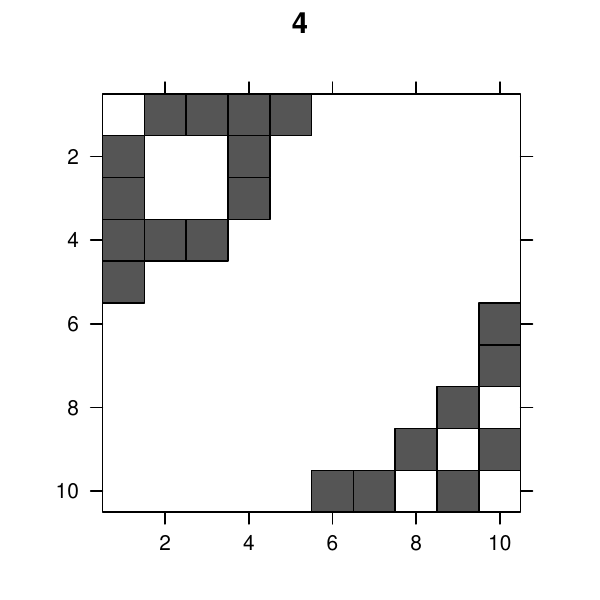}
\caption{Shared sparsity patterns for four $10 \times 10$ precision matrices. for elements $\cG_{x,ii'}$ in the upper $5 \times 5$ block, matrices (1,2) and (3,4) have the same non-zero support, i.e. $\cG_{x,ii'} = \{ (1,2),(3,4)\}$. On the other hand, when $i,i'$ are in the lower block, $\cG_{x,ii'} = \{ (1,3),(2,4)\}$}
\label{fig:jsem-exmaple}
\end{figure}

We assume an element-wise group sparsity pattern over $k$ for the precision matrices $\Omega_x^k$:
\[
\cG_x = \{ \cG_x^{ii'}: i \neq i'; i, i' \in \cI_p \},
\]
where each $\cG_x^{ii'}$ is a partition of $\cI_K$, and consists of non-overlapping index groups $g$ such that $g \subseteq \cI_K, \cup_{g \in \cG_x^{ii'}} g = \cI_K$. First introduced in \citet{MaMichailidis15}, this formulation helps incorporate group structures that are common across some of the precision matrices being modeled. Figure~\ref{fig:jsem-exmaple} illustrates this through a small example. Subsequently, we use the Joint Structural Estimation Method (JSEM, \citet{MaMichailidis15}) to estimate $\Omega_x$, which first uses the group structure given by $\cG_x$ in penalized nodewise regressions \citep{MeisenBuhlmann06} to obtain neighborhood coefficients $\zeta_i = (\bfzeta_i^1, \ldots, \bfzeta_i^K)$ of each variable $X_i, i \in \cI_p$, then fits a maximum likelihood model over the combined support sets to obtain sparse estimates of the precision matrices:
\begin{align}\label{eqn:jsem-model}
\widehat \zeta_i &= \argmin_{\zeta_i} \left\{
\frac{1}{n} \sum_{k=1}^K \| \bfX_i^k - \bfX_{-i}^k \bfzeta_i^k \|^2 +
\sum_{i' \leq i} \sum_{g \in \cG_x^{ii'}} \eta_n \| \bfzeta_{ii'}^{[g]} \| \right\}, \notag\\
\widehat E_x^k &= \{(i,i'): 1 \leq i < i' \leq p, \hat \zeta_{ii'}^k \neq 0 \text{ OR } \hat \zeta_{i'i}^k \neq 0 \}, \notag\\
\widehat \Omega_x^k &= \argmin_{\Omega_x^k \in \BS_+ (\hat E_x^k)}
\left\{ \Tr (\widehat \bfS_x^k \Omega_x^k ) - \log \det (\Omega_x^k) \right\}.
\end{align}
where $\widehat \bfS_x^k := (\bfX^k)^T \bfX^k/n$, $\eta_n$ is a tuning parameter, and $\BS_+ (\hat E_x^k)$ is the set of positive-definite matrices that have non-zero supports restricted to $\hat E_x^k$.

For the precision matrices $\Omega_y^k$, we assume an element-wise sparsity pattern $\cG_y$ defined in a similar manner as $\cG_x$. The sparsity structure $\cH$ for $\cB$ is more general, each non-overlapping group $h \in \cH$ being defined as:
$$
h = \{ (\cS_p, \cS_q, \cS_K): \cS_p \subseteq \cI_p, \cS_q \subseteq \cI_q, \cS_K \subseteq \cI_K \}
; \quad \bigcup_{h \in \cH} h = \cI_p \times \cI_q \times \cI_K.
$$
In other words, any arbitrary partition of $\cI_p \times \cI_q \times \cI_K$ can be specified as the sparsity pattern of $\cB$.

Denote the neighborhood coefficients of the $j^{\Th}$ variable in the lower layer by $\bftheta_j^k$, and $\Theta_j := (\bftheta_j^1, \ldots, \bftheta_j^K), \Theta = \{ \Theta_j \}$. We obtain sparse estimates of $\cB, \Theta$, and subsequently $\Omega_y$, by solving the following group-penalized least square minimization problem that has the tuning parameters $\gamma_n$ and $\lambda_n$ and then refitting:
\begin{align}
\{ \widehat \cB, \widehat \Theta \} &= 
\argmin_{\cB, \Theta} \left\{ \frac{1}{n} \sum_{j=1}^q \sum_{k=1}^K \| \bfY^k_j - (\bfY_{-j}^k - \bfX^k \bfB_{-j}^k) \bftheta_j^k - \bfX^k \bfB_j^k \|^2 \right. \notag\\
& \left. + \sum_{j \neq j'} \sum_{g \in \cG_y^{jj'}} \gamma_n \| \bftheta_{jj'}^{[g]} \| + \sum_{h \in \cH} \lambda_n \| \bfB^{[h]} \| \right\}, \label{eqn:jmmle-objfun}\\
\widehat E_y^k &= \{(j,j'): 1 \leq j < j' \leq q, \hat \theta_{jj'}^k \neq 0 \text{ OR } \hat \theta_{j'j}^k \neq 0 \}, \notag\\
\widehat \Omega_y^k &= \argmin_{\Omega_y^k \in \BS_+ (\hat E_y^k)}
\left\{ \Tr (\widehat \bfS_y^k \Omega_y^k ) - \log \det (\Omega_y^k) \right\}. \label{eqn:omega-y-calc}
\end{align}
%
The outcome of a node in the lower layer is thus modeled using all other nodes in that layer using the neighborhood coefficients $\widehat \bfB_j^k$, {\it and} nodes in the immediate upper layer using the regression coefficients $\widehat \bftheta_j^k$.

\begin{remark}
Common sparsity structures across the same layer are incorporated into the regression by the group penalties over the element-wise groups $\bftheta_{jj'}^{[g]}$, while sparsity pattern overlaps across the different regression matrices $\bfB^k$ are handled by the group penalties over $\bfB^{[h]}$, which denote the collection of elements in $\cB$ that are in $h$. Other kinds of structural assumptions on $\cB$ or $\Theta$ can be handled within the above structure by swapping out the group norms in favor of other appropriate norm-based penalties.
\end{remark}

\begin{remark}
Group sparsity assumptions are not necessary for the JMMLE framework: rather, they help leverage additional information regarding interaction of features in and between the layers in many applications, \textit{as and when that information is available}. In the vertical direction of the model, i.e. given a fixed $k$, a framework agnostic of any structural dependency assumptions amounts to element-wise groups in $\bfB^k$ and $\Omega_y^k$. In JMMLE, this occurs by construction for $\Theta$, and since $\cH$ consists of all possible partitions of $\cI_p \times \cI_q \times \cI_K$, it covers the case of element-wise groups as well. On the other hand, the absence of any horizontal (i.e. across $k$) dependency simply decomposes the problems \eqref{eqn:jmmle-objfun} and \eqref{eqn:omega-y-calc} into $K$ independent sub-problems that can be solved separately either by setting $K=1$ in our framework or by using existing methods, such as \citet{LinEtal16}.
The proposed framework provides signficant more generality and aims at \textit{tight} vertical and horizontal integration based on available prior information.
\end{remark}

\subsubsection{Alternating Block Algorithm}
The objective function in \eqref{eqn:jmmle-objfun} is bi-convex, i.e. convex in $\cB$ for fixed $\Theta$, and vice-versa, but not jointly convex in $\{ \cB, \Theta \}$. Consequently, we use an alternating iterative algorithm to solve for $\{ \cB, \Theta \}$ that minimizes \eqref{eqn:jmmle-objfun} by iteratively cycling between $\cB$ and $\Theta$, i.e. holding one set of parameters fixed and solving for the other, then alternating until convergence.

Choice of initial values plays a crucial role in the performance of this algorithm as discussed in detail in
\citet{LinEtal16}. 
%
%
We choose the initial values $\{ \widehat \bfB^{k (0)} \}$ by fitting separate lasso regression models for each $j$ and $k$:
\begin{align}\label{eqn:init-B}
\widehat \bfB_j^{k (0)} = \argmin_{\bfB_j^k \in \BR^p} \|\bfY_j^k - \bfX^k \bfB_j^k \|^2 + \lambda_n \| \bfB_j^k \|_1; \quad
j \in \cI_q, k \in \cI_K.
\end{align}

We obtain initial estimates of $ \Theta_j, j \in \cI_q$ by performing group-penalized nodewise regression on the residuals $\widehat \bfE^{k (0)} := \bfY^k - \bfX^k \widehat \bfB_j^{k (0)}$:
\begin{align}\label{eqn:init-Theta}
\widehat \Theta_j^{(0)} = \argmin_{\Theta_j} \frac{1}{n} \sum_{k=1}^K \|
\widehat \bfE_j^{k (0)} - \widehat \bfE_{-j}^{k (0)} \bftheta_j^k \|^2
+ \gamma_n \sum_{j \neq j'} \sum_{g \in \cG_y^{jj'}} \| \bftheta_{jj'}^{[g]} \|.
\end{align}

The steps of our full estimation procedure, coined as the {\it Joint Multiple Multi-Layer Estimation} (JMMLE) method, are summarized in Algorithm \ref{algo:jmmle-algo}.

\begin{Algorithm}
(The JMMLE Algorithm)
\label{algo:jmmle-algo}

\noindent 1. Initialize $\widehat \cB$ using \eqref{eqn:init-B}.

\noindent 2. Initialize $\widehat \Theta$ using \eqref{eqn:init-Theta}.

\noindent 3. Update $\widehat \cB$ as:
\begin{align}\label{eqn:update-B}
\widehat \cB^{(t+1)} &= \argmin_{\substack{\bfB^k \in \BM(p,q)\\k \in \cI_K}} \left\{ \frac{1}{n} \sum_{j=1}^q \sum_{k=1}^K \| \bfY^k_j - (\bfY_{-j}^k - \bfX^k \bfB_{-j}^k) \widehat \bftheta_j^{k (t)} - \bfX^k \bfB_j^{k } \|^2
+ \lambda_n \sum_{h \in \cH} \| \bfB^{[h]} \| \right\}
\end{align}

\noindent 4. Obtain $\widehat \bfE^{k (t+1)} := \bfY^k - \bfX^k \bfB_j^{k (t)}, k \in \cI_K$. Update $\widehat \Theta$ as:
\begin{align}\label{eqn:update-Theta}
\widehat \Theta_j^{(t+1)} = \argmin_{\Theta_j \in \BM(q-1, K)}
\left\{ \frac{1}{n} \sum_{k=1}^K
\| \widehat \bfE_j^{k (t+1)} - \widehat \bfE_{-j}^{k (t+1)} \bftheta_j^k \|^2
+ \gamma_n \sum_{j \neq j'} \sum_{g \in \cG_y^{jj'}} \| \bftheta_{jj'}^{[g]} \| \right\}
\end{align}

\noindent 5. Continue till convergence.

\noindent 6. Calculate $\widehat \Omega_y^k, k \in \cI_K$ using \eqref{eqn:omega-y-calc}.
\end{Algorithm}

\subsubsection{Tuning parameter selection}

A number of methods have been proposed in the literature to select regularization tuning parameters in $\ell_1$-penalized problems. Some approaches rely on traditional criteria like cross-validation, Akaike Information Criterion (AIC)~\citep{DanaherEtal14} or the Bayesian Information Criterion (BIC)~\citep{LinEtal16,MaMichailidis15}. A number of studies have proposed their modifications for the case when feature dimensions increase with sample size \citep{FoygelDrton10,GaoEtal12,KimKwonChoi12}.

As a demonstration, to select the tuning parameter $\lambda_n$ we use the High-dimensional BIC (HBIC, \citet{KimKwonChoi12, WangKimLi13}), and for selecting $\gamma_n$ in the node-wise regression step in the JSEM model \eqref{eqn:jsem-model}, employ BIC as in \citet{MaMichailidis15}. Unlike BIC, the penalty term in HBIC scales with the parameter dimensions. As a result, the tuning parameter selected as the minimizer of HBIC asymptotically identifies the oracle estimator in ultra-high dimensional penalized problems \citep{FanTang13, WangKimLi13}. In our case, we train multiple JMMLE models using Algorithm \ref{algo:jmmle-algo} over a finite set of values $\lambda_n \in \cD_n$, and calculate their HBIC:
\begin{align*}
\text{HBIC} (\lambda_n; \Theta) &=
\frac{1}{n} \sum_{j=1}^q \sum_{k=1}^K \| \bfY^k_j - (\bfY_{-j}^k - \bfX^k \widehat \bfB_{-j,\lambda_n}^k ) \bftheta_j^{k } - \bfX^k \widehat \bfB_{j,\lambda_n}^k \|^2 +\\
& \log (\log n) \frac{\log (pq)}{n} \sum_{k=1}^K
\left( \| \bfB^k \|_0 + | \widehat E_{y, \gamma_n^* (\lambda_n)}^k| \right).
\end{align*}
Following this step, we select the optimal $\lambda_n$ as the empirical minimizer of HBIC over $\cD_n$: $
\lambda^* = \argmin_{\lambda_n \in \cD_n} \text{HBIC} (\lambda, \widehat \Theta_{\gamma_n^*(\lambda_n)})
$.

The step for updating $\Theta$ (i.e. \eqref{eqn:update-Theta} in Algorithm~\ref{algo:jmmle-algo}) in our JMMLE algorithm is analogous to the JSEM method \citet{MaMichailidis15}, hence we use BIC to select the penalty parameter $\gamma_n$. In our setting the BIC for a given $\gamma_n$ and fixed $\cB$ is given by:
\begin{align*}
\text{BIC} (\gamma_n; \cB) &=
\Tr \left( \bfS_y^k \widehat \Omega_{y,\gamma_n}^k \right) - \log \det \left( \widehat \Omega_{y,\gamma_n}^k \right) +
\frac{\log n}{n} \sum_{k=1}^K | \widehat E_{y,\gamma_n}^k |
\end{align*}
where $\gamma_n$ in subscript indicates the corresponding quantity is calculated taking $\gamma_n$ as the tuning parameter, and $\bfS_y^k := (\bfY^k - \bfX^k \bfB^k)^T (\bfY^k - \bfX^k \bfB^k)/n$. Every time $\widehat \Theta$ is updated in the JMMLE algorithm, we choose the optimal $\gamma_n$ as the one with the smallest BIC over a fixed set of values $\cC_n$. Thus for a fixed $\lambda_n \equiv \lambda$, our final choice of $\gamma_n$ will be 
$
\gamma_n^* (\lambda) = \argmin_{\gamma_n \in \cC_n} \text{BIC} (\gamma_n; \widehat \cB_{\lambda_n})
$.

%
%
%

%

\subsection{Properties of JMMLE estimators}
\label{sec:jmmle-theory}
We now provide theoretical results ensuring the convergence of our alternating algorithm, as well as the consistency of estimators obtained from the algorithm. We present statements of theorems in the main body of the paper, while detailed proofs and auxiliary results are delegated to the Appendix.

We introduce some additional notation and define technical conditions that help establish the results that follow. Denote the true values of the parameters by $\Omega_{x 0} = \{ \Omega_{x 0}^k \}, \Omega_{y 0} = \{ \Omega_{y 0}^k \}, \Theta_0 = \{ \Theta_{0 j} \}, \cB_0 = \{ \bfB_0^k \}$. Sparsity levels of individual true parameters are indicated by $s_j := | \supp (\Theta_{0j})|, b_k := | \supp(\bfB^k_0) |$. Also define $S := \sum_{j=1}^q s_j, B:= \sum_{k=1}^K b_k, s:= \max_{j \in \cI_q } s_j$, and $\cX := \{ \bfX^k \}_{k=1}^K, \cE := \{ \bfE^k \}_{k=1}^K$.

\begin{definition}[Bounded eigenvalues]
A positive definite matrix $\Sigma \in \BM(b,b)$ is said to have bounded eigenvalues with constants $(c_0, d_0)$ if
\[
0 < 1/c_0 \leq \Lambda_{\min} (\Sigma) \leq \Lambda_{\max} (\Sigma) \leq 1/d_0 < \infty
\]
\end{definition}

\begin{definition}[Diagonal dominance] A matrix $\bfM  \in \BM(b,b)$ is said to be strictly diagonally dominant if for all $a \in \cI_b$,
$$
| (\bfM)_{aa} | > \sum_{a' \neq a} |(\bfM)_{aa'} |
$$
Denote $\Delta_0 (\bfM) = \min_a \{ | (\bfM)_{aa} | - \sum_{a' \neq a} |(\bfM)_{aa'} | \}$.
\end{definition}

Our first result establishes the convergence of Algorithm~\ref{algo:jmmle-algo} for fixed realizations of $(\cX,\cE)$.

\begin{theorem}
\label{thm:algo-convergence}
Suppose for any fixed $(\cX, \cE)$, estimates in each iterate of Algorithm~\ref{algo:jmmle-algo} are uniformly bounded by some quantity dependent on only $p, q$ and $n$:
\begin{align}
\left\| (\widehat \cB^{(t)}, \widehat \Theta_y^{(t)}) - ( \cB_0, \Theta_{y 0}) \right\|_F
\leq R(p,q,n);
\quad t \geq 1
\end{align}
Then, any limit point $(\cB^\infty, \Theta_y^\infty)$ of the algorithm is a stationary point of the objective function, i.e. a point where partial derivatives along all coordinates are non-negative.
\end{theorem}

As established in Theorems \ref{thm:thm-Theta} and \ref{thm:thm-B}, at sub-iterations of Algorithm~\ref{algo:jmmle-algo} (specifically steps 3 and 4) a $O(\sqrt{\log(pq)}/n)$ bound on $\widehat \cB^{(t)}$ leads to $\widehat \Theta^{(t+1)}$ being consistent for $\Theta_0$, and a $O(\sqrt{\log q}/n)$ bound on $\widehat \Theta^{(t)}$ leads to $\widehat \cB^{(t+1)}$ being consistent for $\cB_0$: both with probability approaching 1 as $p,q,n \rightarrow \infty$. Thus, while the constant $R(p,q,n)$ in Theorem~\ref{thm:algo-convergence} above does not need to obey any explicit bounds for Algorithm~\ref{algo:jmmle-algo} to have {\it a} limit point, having a tighter $O(\sqrt{\log(pq)}/n)$ bound ensures that the limit point lies close to the population parameters with high enough probability.

The next steps establish that for random realizations of $\cX$ and $\cE$,
(a) successive iterates lie in this non-expanding ball around the true parameters, and (b) the procedures in \eqref{eqn:init-B} and \eqref{eqn:init-Theta} ensure starting values that lie inside the same ball, both with probability approaching 1 as $(p,q,n) \rightarrow \infty$. To do so, we break down the main problem into two sub-problems. Take as $\bfbeta = (\ve (\bfB^1)^T, \ldots, \ve(\bfB^K)^T)^T$: any subscript or superscript on $\bfB$ being passed on to $\bfbeta$. Denote by $\widehat \Theta$ and $\widehat \bfbeta$ the generic estimators given by
\begin{align}
\widehat \Theta_j &= \argmin_{\Theta_j \in \BM(q-1, K)} \left\{ \frac{1}{n} \sum_{k=1}^K \| \widehat \bfE^k_j - \widehat \bfE^k_{-j} \bftheta_j^k \|^2 + \gamma_n \sum_{j \neq j'} \sum_{g \in \cG_y^{jj'}} \| \bftheta_{jj'}^{[g]} \| \right\};
\quad j \in \cI_q, \label{eqn:EstEqn2}\\
\widehat \bfbeta &= \argmin_{\bfbeta \in \BR^{pqK}} \left\{-2 \bfbeta^T \widehat \bfgamma + \bfbeta^T \widehat \bfGamma \bfbeta + \lambda_n \sum_{h \in \cH} \| \bfbeta^{[h]}  \| \right\}, \label{eqn:EstEqn1}
\end{align}
where
$$
\widehat \bfGamma = \begin{bmatrix}
(\widehat \bfT^1)^2 \otimes \frac{(\bfX^1)^T \bfX^1}{n} & &\\
& \ddots &\\
& & (\widehat \bfT^K)^2 \otimes \frac{(\bfX^K)^T \bfX^K}{n}
\end{bmatrix}; \quad
\widehat \bfgamma = \begin{bmatrix}
(\widehat \bfT^1)^2 \otimes \frac{(\bfX^1)^T}{n}\\
\vdots\\
(\widehat \bfT^K)^2 \otimes \frac{(\bfX^K)^T}{n}
\end{bmatrix}
\begin{bmatrix}
\ve (\bfY^1)\\
\vdots\\
\ve (\bfY^K)
\end{bmatrix},
$$
with 
\begin{align}\label{eqn:define-T}
\hat T_{jj'}^k = \begin{cases}
1 &\text{ if } j = j'\\
- \hat \theta_{jj'}^k &\text{ if } j \neq j'
\end{cases}.
\end{align}
Using matrix algebra it is easy to see that solving for $\cB$ in \eqref{eqn:jmmle-objfun} given a fixed $\widehat \Theta$ is equivalent to solving \eqref{eqn:EstEqn1}.

Next, we assume the following conditions:

\vspace{1em}
\noindent{\bf (E1)} The matrices $\Omega_{y0}^k, k \in \cI_K$ are diagonally dominant,

\noindent{\bf (E2)} The matrices $\Sigma_{y0}^k, k \in \cI_K$ have bounded eigenvalues with constants $(c_y, d_y)$ that are common across $k$.
\vspace{1em}

Now, we are in a position to establish the estimation consistency for \eqref{eqn:EstEqn2}, as well as the consistency of the final estimates $\widehat \Omega_y^k$ using their support sets.

\begin{theorem}\label{thm:thm-Theta}
Consider random $(\cX, \cE)$, any deterministic $\widetilde \cB$ that satisfy the following bound
$$
\| \widetilde \bfB^k - \bfB_0^k \|_1 \leq C_\beta \sqrt{ \frac{ \log(pq)}{n}},
$$
where $C_\beta$ depends only on $\cB_0$. Then, for sample size $n \succsim \log (pq)$ there exist constants $c_1, c_3, c_4 > 0, c_2, c_5 > 1, \tau_1 > 2$ such that with probability at least
$$
1 - K( 1/p^{\tau_1-2} - c_1 \exp [-(c_2^2-1) \log(pq)] - 2 \exp (- c_3 n) - c_4 \exp [-(c_5^2-1) \log(pq)]),
$$
the following bounds hold:

\noindent (I) Denote $|g_{\max}| = \max_{g \in \cG_y} |g|$. Then for the choice of tuning parameter
$$
\gamma_n \geq 4 \sqrt{| g_{\max}|} \BQ_0 \sqrt {\frac{ \log (p q)}{n}},
$$
where $\BQ_0$ depends on the model parameters only, we have
\begin{align}
\| \widehat \Theta_j - \Theta_{0,j} \|_F & \leq 12 \sqrt{s_j} \gamma_n / \psi, \label{eqn:theta-norm-bound-1}\\
\sum_{j \neq j', g \in \cG_y^{jj'}} \| \hat \bftheta_{jj'}^{[g]} - \bftheta_{0,jj'}^{[g]} \| & \leq 48 s_j \gamma_n / \psi. \label{eqn:theta-norm-bound-2}
\end{align}
with $\psi = \min_k \Lambda_{\min} (\Sigma_{x0}^k)/2$.

\noindent (II) For the choice of tuning parameter $\gamma_n = 4 \sqrt{| g_{\max}|} \BQ_0 \sqrt{\log (p q)/n}$,
\begin{align}\label{eqn:OmegaBounds0}
\frac{1}{K} \sum_{k=1}^K \| \widehat \Omega_y^k - \Omega_{y0}^k \|_F \leq
O \left( \BQ_0 \sqrt{\frac{| g_{\max}| S}{K}} 
\sqrt {\frac{ \log (p q)}{n}} \right).
\end{align}
\end{theorem}


Condition (E2) ensures that the lower layer covariance matrices are well-conditioned, so that the precision matrices $\{ \Omega_y^k \}$ exist. The diagonal dominance condition (E1) is a sufficient condition for the convergence bounds of Theorem~\ref{thm:thm-Theta} to hold. Specifically, the upper bounds of the finite-sample error rates of estimates $\widehat \Theta$ and $\widehat \Omega_y$ are controlled by the ratios of off-diagonal to diagonal elements, i.e. $\omega_{y,jj}^k/\sum_{j' \neq j} | \omega_{y,jj'}^k| $ through the multiplier $\BQ_0$. A number of $\ell_1$-penalized problems make Restricted Eigenvalue (RE)-type assumptions \citep{BickelRitovTsybakov09,LohWainwright12,LinEtal16} on the (upper layer) design matrices. Following \citet{BasuMichailidis15,LinEtal16} (see Lemma B.1 and Proposition 1 in respective papers), we utilize the diagonal dominance condition to ensure RE conditions for some key model quantities.

To prove an equivalent result for the solution of \eqref{eqn:EstEqn1}, we need the following conditions on the true parameter versions $(\bfT_0^k)^2$, defined from $\Theta_0$ similarly as \eqref{eqn:define-T}.

\vspace{1em}
\noindent{\bf (E3)} The matrices $(\bfT_0^k)^2, k \in \cI_K$ are diagonally dominant,

\noindent{\bf (E4)} The matrices $\Sigma_{x0}^k, k \in \cI_K$ have bounded eigenvalues with common constants $(c_x, d_x)$.
\vspace{1em}

\noindent Given these, we next establish the required consistency results.

\begin{theorem}\label{thm:thm-B}
Assume random $(\cX, \cE)$, and fixed $\widetilde \Theta$ so that for $j \in \cI_q$,
\[
\| \widetilde \Theta_j - \Theta_{0,j} \|_F \leq C_\Theta \sqrt{\frac{\log q}{n}}
\]
for some $C_\Theta$ dependent on $\Theta_0$ only. Then, given the choice of tuning parameter
$$
\lambda_n \geq 4 \sqrt{| h_{\max} |} \BR_0 \sqrt{ \frac{ \log(pq)}{n}},
$$
where $\BR_0$ depends on the population parameters only, with probability at least
$$ 1 - K( c_1 \exp[-(c_2^2-1) \log(pq)] - 2 \exp( -c_3 n)) $$
the following bounds hold:
\begin{align}
\| \widehat \bfbeta - \bfbeta_0 \|_1 & \leq 48 \sqrt{ | h_{\max} |} B \lambda_n / \psi_*, \label{eqn:BetaThm1}\\
\| \widehat \bfbeta - \bfbeta_0 \| & \leq 12 \sqrt B \lambda_n / \psi_*, \label{eqn:BetaThm2}\\
\sum_{h \in \cH} \| \bfbeta^{[h]} - \bfbeta_0^{[h]} \| & \leq 48 B \lambda_n / \psi_*, \label{eqn:BetaThm3}\\
(\widehat \bfbeta - \bfbeta_0 )^T \widehat \bfGamma (\widehat \bfbeta - \bfbeta_0 ) & \leq
72 B \lambda_n^2 / \psi_*, \label{eqn:BetaThm4}
\end{align}
where $|h_{\max}| = \max_{h \in \cH} |h|$, $d_k$ is the maximum degree $(\bfT_0^k)^2$, and
$$
\psi_*= \frac{1}{2} \min_k \left[ \Lambda_{\min} (\Sigma_{x 0}^k) \left( \Delta_0 ( (\bfT_0^k)^2)
- d_k C_\Theta \sqrt{ \frac{\log (pq)}{n}} \right) \right].
$$
\end{theorem}

\begin{remark}
In an effort to keep the JMMLE framework as general as possible, we do not impose any explicit sparsity conditions on the fixed quantities $\widetilde \cB$ and $\widetilde \Theta$ used to estimate the other parameter inside an iteration of Algorithm~\ref{algo:jmmle-algo}. Since $C_\beta$ (or $C_\Theta$) depends only on the population parameter $\cB_0$ (or $\Theta_0$), when that parameter is actually sparse their corresponding sparsity values can be a part of $C_\beta$ (or $C_\Theta$). When we do obtain the actual estimates (\eqref{eqn:theta-norm-bound-1}--\eqref{eqn:OmegaBounds0} in Theorem~\ref{thm:thm-Theta} and \eqref{eqn:BetaThm1}--\eqref{eqn:BetaThm4} in Theorem~\ref{thm:thm-B}), their finite-sample error bounds scale with the corresponding sparsity parameters $\{s_j\}$ and $B$ at rates that are standard in the literature \citep{BasuMichailidis15,LohWainwright12,RavikumarEtal11}.
\end{remark}

Following the choice of tuning parameters in Theorems \ref{thm:thm-Theta} and \ref{thm:thm-B}, $S = o(n/\log (pq))$ and $B = o(n/\log (pq))$ are sufficient conditions on the sparsity of corresponding parameters for the JMMLE estimators to be consistent. As the last step to establish estimation consistency for the limit points of Algorithm~\ref{algo:jmmle-algo}, we now ensure that the starting values are satisfactory as previously discussed.

\begin{theorem}\label{thm:starting-values}
Consider the starting values as derived in \eqref{eqn:init-B} and \eqref{eqn:init-Theta}. For sample size $n \succsim \log(pq)$, and the choice of the tuning parameter
\[
\lambda_n \geq 4 c_2 \max_{k \in \cI_K} \left\{ [\Lambda_{\max} (\Sigma_{x 0}^k) \Lambda_{\max} (\Sigma_{y 0}^k)]^{1/2} \right\}
\sqrt{ \frac{\log (pq)}{n}},
\]
we have $\| \widehat \bfbeta^{(0)} - \bfbeta_0 \|_1 \leq 64 B \lambda_n /\psi^*$ with probability at least $1 - c_1 \exp( -(c_2^2-1) \log(pq)) - 2 \exp(c_3 n)$. Also, for $\gamma_n \geq 4\sqrt{| g_{\max}|} \BQ_0 \sqrt{ \log (pq)/n}$ we have
\begin{align*}
\| \widehat \Theta_j^{(0)} - \Theta_{0,j} \|_F & \leq 12 \sqrt{s_j} \gamma_n / \psi,\\
\sum_{j \neq j', g \in \cG_y^{jj'}} \| \hat \bftheta_{jj'}^{[g](0)} - \bftheta_{0,jj'}^{[g]} \| & \leq 48 s_j \gamma_n / \psi,
\end{align*}
with probability at least
$$ 1 - K(1/p^{\tau_1-2} - c_1 \exp [-(c_2^2-1) \log(pq)] - 2 \exp (- c_3 n) - c_4 \exp [-(c_5^2-1) \log(pq)]).
$$
\end{theorem}

Putting all the pieces together, the required consistency result given our choice of starting values follows in a straightforward manner.

\begin{corollary}\label{corollary:jmmle-final}
Assume conditions (E1)-(E4), and starting values $\{ \cB^{(0)}, \Theta^{(0)} \}$ obtained using \eqref{eqn:init-B} and \eqref{eqn:init-Theta}, respectively. Then, for random realizations of $\cX, \cE$,

\vspace{1em}
\noindent (I) For the choice of $\lambda_n$
$$
\lambda_n \geq 4 \max \left[ c_2 \max_{k \in \cI_K} \left\{ [\Lambda_{\max} (\Sigma_{x 0}^k) \Lambda_{\max} (\Sigma_{y 0}^k)]^{1/2} \right\}, \sqrt{| h_{\max}|} \BR_0 \right] \sqrt{\frac{\log(pq)}{n}},
$$
we have
$$
\| \widehat \bfbeta - \bfbeta_0 \|_1 \leq \max \left\{ 48 \sqrt{ | h_{\max} |}, 64 \right\} \frac{B \lambda_n}{\psi_*}
$$
with probability at least $1 - 18 c_1 \exp[-(c_2^2-1) \log(pq)] - 4 \exp( -c_3 n)$.

\vspace{1em}
\noindent (II) For the choice of $\gamma_n$
$$
\gamma_n \geq 4 \sqrt{ | g_{\max} |} \BQ_0 \sqrt{\frac{\log(pq)}{n}},
$$
\eqref{eqn:theta-norm-bound-1} and \eqref{eqn:theta-norm-bound-2} hold, while for $\gamma_n = 4 \sqrt{ | g_{\max} |} \BQ_0 \sqrt{ \log (pq)/n}$, \eqref{eqn:OmegaBounds0} holds, both with probability at least
$$
1 - K(2/p^{\tau_1-2} - 2 c_1 \exp [-(c_2^2-1) \log(pq)] - 4 \exp (- c_3 n) - 2 c_4 \exp [-(c_5^2-1) \log(pq)]).
$$
\end{corollary}

\begin{remark}
To save computation time for high data dimensions, an initial screening step, e.g. the debiased lasso procedure of \citet{JavanmardMontanari14}, can be used to first restrict the support set of $\bfB_j^k$ before obtaining the initial estimates using \eqref{eqn:init-B}. The consistency properties of resulting initial and final estimates follow along the lines of the special case $K=1$ discussed in \citet{LinEtal16}, in conjunction with Theorem~\ref{thm:starting-values} and Corollary~\ref{corollary:jmmle-final}, respectively. We leave the details to the reader.
\end{remark}

\begin{remark}
While the proof of the above results follow roughly similar roadmaps as the case with $K=1$ and simple $\ell_1$-penalization \citep{LinEtal16} and the joint structural estimation of \citet{MaMichailidis15} and utilize Gaussian concentration inequalities, generalization to $K>1$ and an optional grouping structure in $\cB$ add significant additional technical complexity to the proofs. More importantly, we work in presence of minimal assumptions, steering clear of conditions used in previous works, like Incoherence \citep{LinEtal16} and Uniform Irrepresentability \citep{MaMichailidis15} that are hard to verify in practice.
\end{remark}

\section{Hypothesis testing in multilayer models}
\label{sec:sec3}
In this section, we lay out a framework for hypothesis testing in our proposed joint multi-layer structure. Present literature in high-dimensional hypothesis testing either focuses on testing for similarities in the within-layer connections of single-layer networks \citep{CaiLiu16,Liu17}, or coefficients of single response penalized regression \citep{vanDeGeerEtal14,ZhangZhang14,MitraZhang16}. However, to our knowledge no method is available in the literature to perform testing for {\it between-layer} connections in a two-layer (or multi-layer) setup.

Denote the $i\Th$ row of the coefficient matrix $\bfB^k$ by $\bfb_i^k$, for $i \in \cI_p$. In this section we are generally interested in obtaining asymptotic sampling distributions of $\widehat \bfb_i^k$, and subsequently formulating testing procedures to detect similarities or differences across $k$ in the full vector $\bfb_i^k$ or its elements. There are two main challenges in doing the above: firstly the need to mitigate the bias of the group-penalized JMMLE estimators, and secondly the dependency among response nodes translating into the need for controlling false discovery rate while simultaneously testing for several element-wise hypotheses concerning the true values $b_{0 ij}^k, j \in \cI_q$. To this end, in Section~\ref{sec:testing-subsec-1} we first propose a debiased estimator for $\bfb_i^k$ that makes use of already computed (using JSEM) node-wise regression coefficients in the upper layer, and establish asymptotic properties of scaled version of them. Section~\ref{sec:testing-subsec-2} is devoted to pairwise testing, where we assume $K=2$, and propose asymptotic global tests for detecting differential effects of a variable in the upper layer, i.e. testing for the null hypothesis $H_0^i: \bfb_{0 i}^1 = \bfb_{0 i}^2$, as well as pairwise simultaneous tests across $j \in \cI_q$ for detecting the element-wise differences $b_{0 ij}^1 - b_{0 ij}^2$.


\subsection{Debiased estimators and asymptotic normality}
\label{sec:testing-subsec-1}
\citet{ZhangZhang14} proposed a debiasing procedure for lasso estimates and subsequently calculate confidence intervals for individual coefficients $\beta_j$ in high-dimensional linear regression: $\bfy = \bfX \bfbeta + \bfepsilon, \bfy \in \BR^n, \bfX \in \BM(n,p)$ and $\epsilon_r \sim N(0,\sigma^2), r \in \cI_n$ for some $\sigma>0$. Given an initial lasso estimate $\widehat \bfbeta^{\text{(init)}} \in \BR^p$ their debiased estimator was defined as:
$$
\hat \beta_j^{(\text{deb})} = \hat \beta_j^{(\text{init})} + \frac{\bfz_j^T ( \bfy - \bfX \hat \bfbeta^{(\text{init})})}{\bfz^T \bfx_j},
$$
where $\bfz_j$ is the vector of residuals from the $\ell_1$-penalized regression of $\bfx_j$ on $\bfX_{-j}$. With centering around the true parameter value, say $\beta_j^0$, and proper scaling this has an asymptotic normal distribution:
$$
\frac{\hat \beta_j^{(\text{deb})} - \beta_j^0}{\| \bfz_j \|/| \bfz_j^T \bfx_j |} \sim N(0, \sigma^2).
$$
Essentially, they obtain the debiasing factor for the $j^{\Th}$ coefficient by taking residuals from the regularized regression and scale them using the projection of $\bfx_j$ onto a space approximately orthogonal to it. \citet{MitraZhang16} later generalized this idea to group lasso estimates. Further, \citet{vanDeGeerEtal14} and \citet{JavanmardMontanari14} performed debiasing on the entire coefficient vectors.

We start off by defining debiased estimates for individual rows of the coefficient matrices $\bfB^k$ in our two-layer model:
\begin{align}\label{eqn:DebiasedBeta}
\widehat \bfc_i^k = \widehat \bfb_i^k + \frac{1}{n t_i^k} \left( \bfX_i^k - \bfX_{-i}^k \widehat \bfzeta_i^k \right)^T (\bfY^k - \bfX^k \widehat \bfB^k )
; \quad i \in \cI_p, k \in \cI_K,
\end{align}
where $\widehat \bfb_i^k$ denotes the $i\Th$ row of $\widehat \bfB^k$, and $t_i^k = ( \bfX_i^k - \bfX_{-i}^k \widehat \bfzeta_i^k )^T \bfX_{i}^k/n$, and $\widehat \bfzeta_i^k, \widehat \bfB^k$ are {\it generic estimators} of the neighborhood coefficient matrices in the upper layer and within-layer coefficient matrices, respectively. By structure this is similar to the proposal of \citet{ZhangZhang14}. However, as seen shortly, minimal conditions need to be imposed on the parameter estimates used in \eqref{eqn:DebiasedBeta} for the asymptotic results based on a scaled version of the debiased estimator to go thorugh, and they continue to hold for arbitrary sparsity patterns over $k$ in all of the parameters.

Present methods of debiasing coefficients from regularized regression require specific assumptions on the regularization structure of the main regression, as well as on how to calculate the debiasing factor. While \citet{ZhangZhang14}, \citet{JavanmardMontanari14} and \citet{vanDeGeerEtal14} work on coefficients from lasso regressions, \citet{MitraZhang16} debias the coefficients of pre-specified groups in the coefficient vector from a group lasso. Current proposals for obtaining the debiasing factor available in the literature include node-wise lasso \citep{ZhangZhang14} and a variance minimization scheme with $\ell_\infty$-constraints \citep{JavanmardMontanari14}. In comparison, we only assume the following generic constraints on the parameter estimates used in our procedure.

\vspace{1em}
\noindent{\bf (T1)} For the upper layer neighborhood coefficients, the following holds for all $k \in \cI_K$:
$$
\| \widehat \bfzeta^k - \bfzeta_0^k \|_1 \leq D_\zeta  = O \left( \sqrt { \frac{\log p}{n}} \right),
$$
where $D_\zeta$ depends only on the true values, i.e. $\{ \zeta^k_0 \}$.

\noindent{\bf (T2)} The lower layer precision matrix estimates satisfy for all $k \in \cI_K$
$$
\| \widehat \Omega_y^k - \Omega_{y0}^k \|_\infty \leq D_\Omega
= O \left( \sqrt { \frac{\log (pq)}{n}} \right),$$
where $D_\Omega$ depends only on $\Omega_{y 0}$.

\noindent{\bf (T3)} For the regression coefficient matrices, the following holds for all $k \in \cI_K$:
$$
\| \widehat \bfB^k - \bfB^k_0 \|_1 \leq D_\beta = O \left( \sqrt { \frac{\log (p q)}{n}} \right),
$$
where $D_\beta$ depends on $\cB_0$ only.
\vspace{1em}

\noindent
The above finite-sample error rates are common in high-dimensional problems, and can pertain to sparse \citep{MaMichailidis15,LinEtal16,LohWainwright12,BasuMichailidis15} or non-sparse estimators \citep{RohdeTsybakov11,BasuEtal19}. Based on the estimators plugged in, additional conditions may be involved in the estimation step. For example, JMMLE estimators satisfy these bounds under conditions (E1)-(E4) following the results in Section~\ref{sec:jmmle-theory}.

Given these conditions, the following result provides the asymptotic joint distribution of a scaled version of the debiased coefficients. A similar result for fixed design in the context of single-response linear regression can be found in \citet{StuckyVandeGeer17}. However, the authors use the nuclear norm as the loss function while obtaining the debiasing factors and employ the resulting Karush-Kuhn-Tucker (KKT) conditions to derive their results, whereas we leverage bounds on generic parameter estimates combined with the sub-Gaussianity of our random design matrices.

\begin{theorem}\label{Thm:ThmTesting}
Define $\widehat s_i^k = \sqrt{\| \bfX_i^k - \bfX_{-i}^k \widehat \bfzeta_i^k \|^2/n}$, and $m_i^k = \sqrt n t_i^k / \widehat s_i^k$. Consider parameter estimates that satisfy conditions (T1)-(T3). Define the following:
\begin{align*}
\widehat \Omega_y &= \diag(\widehat \Omega_y^1, \ldots, \widehat \Omega_y^K),\\
\bfM_i &= \diag(m_i^1, \ldots, m_i^K),\\
\widehat \bfC_i &= \ve(\widehat \bfc_i^1, \ldots, \widehat \bfc_i^K)^T,\\
\bfD_i &= \ve(\bfb_{0i}^1, \ldots, \bfb_{0i}^K)^T.\\
\end{align*}
Also assume that conditions (E2), (E4) hold, and the matrices $\Omega_{x0}^k, k \in \cI_K$ are diagonally dominant. Then, for sample size satisfying 
$\log p = o(n^{1/2}), \log q = o(n^{1/2})$ we have
\begin{align}\label{eqn:ThmTestingEqn}
\widehat \Omega_y^{1/2} \bfM_i (\widehat \bfC_i - \bfD_i) \sim
\cN_{Kq} ({\bf 0}, \bfI) + \bfR_n,
\end{align}
where $\| \bfR_n \|_\infty = o_P(1)$.
\end{theorem}

\subsection{Test formulation}
\label{sec:testing-subsec-2}
We now simply plug in estimators from the JMMLE algorithm in Theorem~\ref{Thm:ThmTesting}. Doing so is fairly straightforward. Condition (T1) is ensured by the JSEM penalized neighborhood estimators in \eqref{eqn:jsem-model} (immediate from Proposition A.1 in \citet{MaMichailidis15}). On the other hand, bounds on total sparsity of the true coefficient matrices: $B = o(\sqrt{n} / \log(pq))$, and lower layer precision matrices: $S = o( n/ \log (pq)$, in conjunction with Corollary~\ref{corollary:jmmle-final}, ensure conditions (T2) and (T3), respectively -all with probability approaching 1 as $(n,p,q) \rightarrow \infty$.

An asymptotic joint distribution of debiased versions of the JMMLE regression estimates can then be obtained immediately.
\begin{corollary}\label{corollary:CorTesting}
Consider the estimates $\widehat \cB$ and $\widehat \Omega_y $ obtained from Algorithm~\ref{algo:jmmle-algo}, and upper layer neighborhood coefficients from solving the node-wise regression in \eqref{eqn:jsem-model}. Suppose that $\log (pq) /\sqrt n \rightarrow 0$, and the sparsity conditions $B = o(\sqrt{n} / \log(pq)), S = o( n / \log(pq))$ are satisfied. Then, with the same notations as in Theorem~\ref{Thm:ThmTesting} we have
\begin{align}\label{eqn:CorTestingEqn}
\widehat \Omega_y^{1/2} \bfM_i (\widehat \bfC_i - \bfD_i) \sim
\cN_{Kq} ({\bf 0}, \bfI) + \bfR_{1n}
\end{align}
where $\| \bfR_{1n} \|_\infty = o_P(1)$.
\end{corollary}

We are now ready to formulate asymptotic global and simultaneous testing procedures based on Corollary~\ref{corollary:CorTesting}. In this paper, we restrict our attention to testing for pairwise differences only. Specifically, we set $K=2$, and are interested in testing whether there are overall and elementwise differences between individual rows of the true coefficient matrices, i.e. $\bfb_{0i}^1$ and $\bfb_{0i}^2$.

When $\bfb_{0 i}^1 = \bfb_{0 i}^2$, it is immediate from Corollary~\ref{corollary:CorTesting} that a scaled version of the vector of estimated differences $\widehat \bfc_i^1 - \widehat \bfc_i^2$ follows a $q$-variate multivariate normal distribution. Consequently, we formulate a global test for detecting differential overall downstream effect of the $i^{\Th}$ covariate in the upper layer.

\begin{Algorithm}\label{algo:AlgoGlobalTest}
(Global test for $H_0^i: \bfb_{0 i}^1 = \bfb_{0 i}^2$ at level $\alpha, 0< \alpha< 1$)

\noindent 1. Obtain the debiased estimators $\widehat \bfc_i^1, \widehat \bfc_i^2$ using \eqref{eqn:DebiasedBeta}.

\noindent 2. Calculate the test statistic
$$
D_i = (\widehat \bfc_i^1 - \widehat \bfc_i^2)^T
\left( \frac{ \widehat \Sigma_y^1}{(m_i^1)^2} +
\frac{\widehat \Sigma_y^2}{(m_i^2)^2} \right)^{-1} (\widehat \bfc_i^1 - \widehat \bfc_i^2)
$$
where $\widehat \Sigma_y^k = (\widehat \Omega_y^k)^{-1}, k = 1,2$.

\noindent 3. Reject $H_0^i$ if $D_i \geq \chi^2_{q, 1-\alpha}$.
\end{Algorithm}

Besides controlling the type-I error at a specified level, the above testing procedure maintains rate optimal power.

\begin{theorem}\label{thm:PowerThm}
Consider the global test given in Algorithm~\ref{algo:AlgoGlobalTest}, performed using parameter estimates satisfying conditions (T1)-(T3). Define $\bfdelta := \bfb_{0 i}^1 - \bfb_{0 i}^2$. Further, assume that either of the following sufficient conditions are satisfied.

\begin{itemize}
\item[(I)] The following bound holds: $D_\Omega \leq \Delta_0 (\Omega_{y0}^k), k \in \cI_K$;

\item[(II)] For every $j \in \cI_q, k \in \cI_K$, we have
$
\sum_{j' = 1}^q | \sigma_{y0,jj'}^k |^q \leq c_0 (p)
$ for some $q \in [0,1)$ and positive-valued function $c_0(\cdot)$.
\end{itemize}

Denote $\sigma_{x0,i,-i}^k = Var( X_i^k - \BX_{-i}^k \bfzeta_{0,i}^k)$. Then, the power of the global test is given by
$$
K_q \left( \chi^2_{q,1-\alpha} + n \bfdelta^T 
\left( \frac{ \Sigma_{y 0}^1}{\sigma_{x0, i,-i}^1} + \frac{\Sigma_{y 0}^2}{\sigma_{x0, i,-i}^2} \right)^{-1} \bfdelta \right) + o(1)
$$
where $K_q$ is the cumulative distribution function of the $\chi^2_q$ distribution. Consequently, for $\| \bfdelta \| > O(n^{-1/2})$, $P( H_0^i \text{ is rejected }) \rightarrow 1$ as $(n,p,q) \rightarrow \infty$.
\end{theorem}

The conditions (I) or (II) above are needed to derive upper bounds for $\| \widehat \Sigma_y^k - \Sigma_{y0}^k \|_\infty$ using those for $\| \widehat \Omega_y^k - \Omega_{y0}^k \|_\infty$. While (I) imposes a potentially more stringent bound on the estimation error of $\Omega_y$, (II) restricts the power calculations to a uniformity class of covariance matrices \citep{BickelLevina08,CaiLiuLuo12}.

\begin{remark}
While the formulation of the testing procedure broadly gives parallel results as  \citet{ZhangZhang14} and \citet{MitraZhang16}, it does so \textit{without assuming any specific penalty function} or (group) sparsity conditions (such as strong group sparsity in \citet{MitraZhang16}). Instead, we only require the standard finite-sample bounds (T1)-(T3), satisfied by existing sparse and non-sparse estimators in a high-dimensional setting.
\end{remark}

\subsection{Control of False Discovery Rate}
Given that the null hypothesis is rejected, we consider the multiple testing problem of simultaneously testing for all entrywise differences, i.e. testing
$$
H_0^{ij}: b_{0 ij}^1 = b_{0ij}^2 \quad \text{vs.} \quad H_1^{ij}: b_{0 ij}^1 \neq b_{0 ij}^2 
$$
for all $j \in \cI_q$. Here we use the test statistic
\begin{align}\label{eqn:PairwiseStatistic}
d_{ij} &= \frac{\widehat c_{ij}^1 - \widehat c_{ij}^2}{\sqrt{\hat \sigma_{jj}^1/ (m_i^1)^2 + \hat \sigma_{jj}^2/ (m_i^2)^2}},
\end{align}
with $\hat \sigma_{jj}^k$ being the $j^{\Th}$ diagonal element of $\widehat \Sigma_y^k, k = 1,2$.

For the purpose of simultaneous testing, we consider tests with a common rejection threshold $\tau$, i.e. for $j \in \cI_q$, $H_0^{ij}$ is rejected if $| d_{ij} | > \tau$. We denote $\cH_0^i = \{ j: b_{0,ij}^1 = b_{0,ij}^2 \}$ and define the False Discovery Proportion (FDP) and False Discovery Rate (FDR) for these tests as follows:
$$
FDP (\tau) = \frac{\sum_{j \in \cH_0^i} \BI( |d_{ij}| \geq \tau)}{\max\left\{
\sum_{j \in \cI_q} \BI( |d_{ij}| \geq \tau), 1\right\} }; \quad
FDR (\tau) = \BE [ FDP (\tau) ].
$$
For a pre-specified level $\alpha$, we choose a threshold that ensures both FDP and FDR $\leq \alpha$ using the Benjamini-Hochberg (BH) procedure. 
The procedure for FDR control is now given by Algorithm \ref{algo:AlgoFDR}.

\begin{Algorithm}\label{algo:AlgoFDR}
(Simultaneous tests for $H_0^{ij}: b_{0 ij}^1 = b_{0 ij}^2$ at level $\alpha, 0< \alpha< 1$)

\noindent 1. Calculate the pairwise test statistics $d_{ij}$ using \eqref{algo:AlgoFDR} for $j \in \cI_q$.

\noindent 2. Obtain the threshold
$$
\hat \tau = \inf \left\{\tau \in \BR: 1 - \Phi(\tau) \leq \frac{\alpha}{2 q}
\max \left( \sum_{j \in \cI_q} \BI( |d_{ij}| \geq \tau), 1 \right) \right\}.
$$

\noindent 3. For $j \in \cI_q$, reject $H_0^{ij}$ if $|d_{ij}| \geq \hat \tau$.
\end{Algorithm}

To ensure that this procedure maintains FDR and FDP asymptotically at a pre-specified level $\alpha \in (0,1)$, we need some dependence conditions on true correlation matrices in the lower layer. Following \citet{LiuShao14}, we consider the following two types of dependencies:

\noindent {\bf (D1)} Define $r_{jj'}^k = \sigma_{y0,jj'}^k /\sqrt{\sigma_{y0,jj}^k \sigma_{y0,j'j'}^k}$ for $j,j' \in \cI_q, k = 1,2$. Suppose there exists $0 < r < 1$ such that $\max_{1 \leq j < j' \leq q} | r_{jj'}^k | \leq r$, and for every $j \in \cI_q$,
$$
\sum_{j'=1}^q \BI \left\{ |r_{jj'}^k| \geq \frac{1}{(\log q)^{2 + \theta}} \right\} \leq O(q^\rho),
$$
for some $\theta > 0$ and $0 < \rho < (1-r)/(1+r)$.

\noindent {\bf (D1*)} Suppose there exists $0 < r < 1$ such that $\max_{1 \leq j < j' \leq q} | r_{jj'}^k | \leq r$, and for every $j \in \cI_q$,
$$
\sum_{j'=1}^q \BI \left\{ |r_{jj'}^k| > 0 \right\} \leq O(q^\rho),
$$
for some $0 < \rho < (1-r)/(1+r)$.

Originally proposed by \citet{LiuShao14}, the above dependency conditions are meant to control the amount of correlation amongst the test statistics. Condition (D1) allows each variable to be highly correlated with at most $O(q^\rho)$ other variables and weakly correlated with others, while (D1*) limits the number of variables to have {\it any} correlation with it to $O(q^\rho)$. Note that (D1*) is a stronger condition, and can be seen as the limiting condition of (D1) as $q \rightarrow \infty$.

\begin{theorem}\label{thm:FDRthm}
Suppose $\mu_j = b_{0,ij}^1 - b_{0,ij}^2, \sigma_j^2 = \sigma_{y0,jj}^1/ \sigma_{x0,i,-i}^1 + \sigma_{y0,jj}^2/ \sigma_{x0,i,-i}^2$. Assume the following holds as $(n,q) \rightarrow \infty$,
\begin{align}\label{eqn:FDRthmEqn1}
\left| \left\{ j \in \cI_q: |\mu_j / \sigma_j | \geq
4 \sqrt{ \log q/n} \right\} \right| \rightarrow \infty.
\end{align}
Next, consider conditions (D1) and (D1*). If (D1) is satisfied, then the following holds when $\log q = O(n^{\xi}), 0 < \xi < 3/23$:
\begin{align}\label{eqn:FDReqn}
\frac{FDP( \hat \tau)}{(| \cH_0^i|/q) \alpha} \stackrel{P}{\rightarrow} 1; \quad
\lim_{n, q \rightarrow \infty} \frac{FDR( \hat \tau)}{(| \cH_0^i|/q) \alpha} = 1.
\end{align}
Further, if (D1*) is satisfied, then \eqref{eqn:FDReqn} holds for $\log q = o(n^{1/3})$.
\end{theorem}
%
The condition \eqref{eqn:FDRthmEqn1} is essential for FDR control in a diverging parameter space \citep{LiuShao14, Liu17}.

\begin{remark}
Based on the FDR control procedure in Algorithm~\ref{algo:AlgoFDR}, we can perform {\it within-row thresholding} in the matrices $\widehat \bfB^k$ to tackle group misspecification.
\begin{align}
& \hat \tau_i^k := \inf \left\{\tau \in \BR: 1 - \Phi(\tau) \leq \frac{\alpha}{2 q}
\max \left( \sum_{j \in \cI_q} \BI( | \sqrt{\hat \omega_{jj}^k} m_i^k \hat c_{ij}^k | \geq \tau), 1 \right) \right\}, \notag\\
& \hat b_{ij}^{k, \text{thr}} =  \hat b_{ij}^k \BI \left(
|\sqrt{\hat \omega_{jj}^k} m_i^k \hat c_{ij}^k | \geq \hat \tau_i^k \right).
\label{eqn:fdr-threshold}
\end{align}
Even without group misspecification, this helps identify directed edges between layers that have high nonzero values. Similar post-estimation thresholdings have been proposed in the context of multitask regression \citep{ObozinskiEtal11,MajumdarChatterjeeStat} and neighborhood selection \citep{MaMichailidis15}. However, our procedure is the first one to provide explicit guarantees on the amount of false discoveries while doing so.
\end{remark}

\begin{remark}
Following \eqref{eqn:FDRthmEqn1}, a sufficient condition on the sparsity of $\cB_0$ for FDR to be asymptotically controlled at some specified level is $B = o(n^\zeta/ \log q)$ if (D1) is satisfied, and $B = o(n^{1/3}/ \log q)$ if (D1*) is satisfied. In comparison, our results for the global testing procedure require $B = o(\sqrt n/ \log (pq))$, and point estimation requires $B = o(n/ \log (pq))$. In finite samples settings, the stricter sparsity requirements translate to higher sample sizes being needed (given the same $(p,q)$) for our testing procedures to have satisfactory performances compared to estimation only (See Sections \ref{sec:eval-jmmle} and \ref{sec:eval-testing}).

In recent work, \citet{JavanmardMontanari18} showed that the $o(\sqrt n/ \log p)$ bound on the sparsity of the true coefficient vector required to construct confidence intervals from debiased lasso coefficient estimates \citep{vanDeGeerEtal14, ZhangZhang14,JavanmardMontanari14} can be weakened to $o(n/ (\log p)^2)$ when the random design precision matrix is known, or is unknown but satisfies certain sparsity assumptions. Similar relaxations may be possible in our case. For example, the machinery in \citet{Liu17}, which performs simultaneous testing in multiple (single layer) GGMs using slightly modified FDR thresholds, can be useful in obtaining \eqref{eqn:FDReqn} for $ \log q = o(n^{1/2})$ under (D1), (D1*) or other suitable dependency assumptions.
\end{remark}

\subsection{Effect of tuning parameter selection}
A topic not adequately addressed in the high-dimensional hypothesis testing literature concerns the effect of the regularization tuning parameter selection methods (HBIC for $\lambda_n$ and BIC for $\gamma_n$ in our case) on the size, power and confidence intervals obtained. Ideally, tuning parameter selection method(s) in the estimation step should ensure that the estimated quantities from the model with optimal tuning parameter choices can be plugged into the debiasing procedure to obtain quantities that obey the correct asymptotic properties and are used in the tests that follow (e.g. Algorithms \ref{algo:AlgoGlobalTest} and \ref{algo:AlgoFDR}).

In our case, the broad-based assumptions (T1)-(T3) allow plugging in estimators with finite-sample error rates that are satisfied by a host of high-dimensional methods, as previously discussed. Given that the tuning parameters $\lambda_n$ and $\gamma_n$ are selected to be above thresholds that scale with feature and sample dimensions, JMMLE estimators adhere to these error rates with high probability (Corollary~\ref{corollary:jmmle-final}), ensuring the correctness of our testing procedures. To empirically make it likely that the correct tuning parameters get selected, in our numerical examples (on synthetic and real data) that follow, we obtain JMMLE estimates over ranges of $\lambda_n$ and $\gamma_n$ that scale with the error rates of estimators. Additional technicalities will be involved for a more rigorous analysis, possibly using technical material from approaches such as \citet{FoygelDrton10,WangKimLi13}. We defer this topic to future work.

%
\section{Numerical experiments}
\label{sec:sec4}
We evaluate the performance of our proposed JMMLE algorithm and the hypothesis testing framework in a two-layer simulation setup (Sections \ref{sec:eval-jmmle} and \ref{sec:eval-testing}, respectively), and also introduce some computational techniques that significantly accelerate calculations for high data dimensions (Section~\ref{sec:tricks-jmmle}).

\subsection{Simulation 1: estimation}
\label{sec:eval-jmmle}
As a first step towards obtaining a two-layer structure with horizontal (across $k$) integration and inter-layer directed edges, we generate the precision matrices $\{ \Omega_{x0}^k \}$ and $\{ \Omega_{y0}^k \}$ using a dependency structure across $k$ that was first used in the simulation study of \citet{MaMichailidis15}. We set $K=5$, and set different shared sparsity patterns across $k$ inside the lower $p/2 \times p/2$ block of the upper layer precision matrices, and outside the block. In our notation, this gives the following elementwise group structure:
$$
\cG_{x,ii'} = \begin{cases}
\{ (1,2),(3,4), 5 \} &\text{ if } i \leq p/2 \text{ or } j \leq p/2,\\
\{ (1,3,5),(2,4) \} &\text{ otherwise}.
\end{cases}
$$

\begin{figure}
\centering
\includegraphics{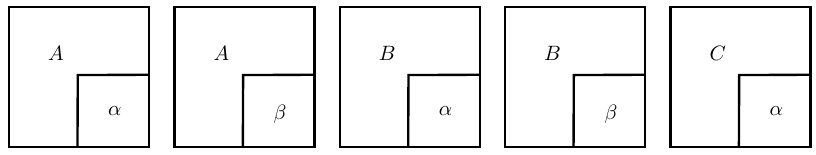}
\caption{Shared sparsity patterns across $k$ for the precision matrices $\{ \Omega_{x0}^k\}$ and $\{ \Omega_{y0}^k\}$}
\label{fig:sim-structure}
\end{figure}

The schematic in Figure~\ref{fig:sim-structure} illustrates this structure. We set an off-diagonal element inside each of these common blocks (i.e. $A,B,C$ and $\alpha, \beta$ in the figure) to be non-zero with probability $\pi_x \in \{ 5/p, 30/p \}$, then generate the values of all non-zero elements independently from the uniform distribution in the interval $[-1, 0.5] \cup [0.5, 1]$. The precision matrices $\Omega_{x0}^k$ are generated by putting together the corresponding common blocks, their positive definiteness ensured by setting all diagonal elements to be $1 + |\Lambda_{\min} (\Omega_{x0}^k)|$. Then, we get elements in the covariance matrix as
$$
\sigma_{x0,ii'}^k = (\bar \Omega_{x0}^k)_{ii'} / \sqrt{(\bar \Omega_{x0}^k)_{ii} (\bar \Omega_{x0}^k)_{i'i'} },
\text{ where } \bar \Omega_{x0}^k = (\Omega_{x0}^k)^{-1},
$$
and generate rows of $\bfX^k$ independently from $\cN(0, \Sigma_{x0}^k)$. We obtain $\Sigma_{y0}^k$ and then $\bfE^k$ using the same setup but with the number of variables being $q$ and setting off-diagonal elements non-zero with probability $\pi_y \in \{ 5/q, 30/q \}$. To obtain the matrices $\bfB_0^k$, for a fixed $(i,j), i \in \cI_p, j \in \cI_q$, we set $b_{0,ij}^k$ non-zero across all $k$ with probability $\pi \in \{ 5/p, 30/p \}$, generate the non-zero groups independently from $\text{Unif} \{ [-1, 0.5] \cup [0.5, 1] \}$, and set $\bfY^k = \bfX^k \bfB_0^k + \bfE^k, k \in \cI_K$. Finally, we generate 150 such independent two-layer datasets for each of the following model settings:

\begin{itemize}
\item Set $\pi_x = \pi = 5/p, \pi_y = 5/q$, and
$$
(p,q,n) \in \{ (60,30,100), (30,60,100), (200,200,150), (300,300,150) \};
$$

\item Set $\pi_x = \pi = 30/p, \pi_y = 30/q$, and $(p,q,n) \in  \{ (200,200,100), (200,200,200) \}$.
\end{itemize}

We use the following scaled arrays of tuning parameters to train Algorithm~\ref{algo:jmmle-algo}-
$$
\gamma_n \in \left\{ 0.3, 0.4, ..., 1 \right\} \sqrt{\frac{\log q}{n}}; \quad
\lambda_n \in \left\{ 0.4, 0.6, ..., 1.8 \right\} \sqrt{\frac{\log p}{n}},
$$
using a one-step version of the algorithm (Section~\ref{sec:tricks-jmmle}) to save computation time.

We use the following performance metrics to evaluate our estimates $\widehat \cB = \{ \widehat \bfB^k \}$:

\begin{itemize}
\item True positive Rate-
\[
\text{TPR}(\widehat \bfB_k) = \frac{| \supp(\hat \bfB^k) \cap \supp (\bfB_0^k)|}{| \supp (\bfB_0^k)|}; \quad
\text{TPR} (\widehat \cB) = \frac{1}{K} \sum_{k=1}^K \text{TP}(\widehat \bfB_k).
\]
\item True negative Rate-
\[
\text{TNR}(\widehat \bfB_k) = \frac{| {\supp}^c (\hat \bfB^k) \cap {\supp}^c (\bfB_0^k)|}{| \supp^c (\bfB_0^k)|}; \quad
\text{TNR} (\widehat \cB) = \frac{1}{K} \sum_{k=1}^K \text{TNR}(\widehat \bfB_k).
\]
\item Matthews Correlation Coefficient-
$$
\text{TP}(\widehat \bfB_k) = | \supp(\hat \bfB^k) \cap \supp (\bfB_0^k)|; \quad
\text{TN}(\widehat \bfB_k) = | {\supp}^c (\hat \bfB^k) \cap {\supp}^c (\bfB_0^k)|,
$$
$$
\text{FP}(\widehat \bfB_k) = | {\supp}^c (\bfB^k_0) | - \text{TN}(\widehat \bfB_k); \quad
\text{FN}(\widehat \bfB_k) = | {\supp} (\bfB^k_0) | - \text{TP}(\widehat \bfB_k), $$
\begin{align*}
& \text{MCC} (\widehat \bfB_k) = \\
& \frac{ \text{TP}(\widehat \bfB_k) \text{TN}(\widehat \bfB_k) -
\text{FP}(\widehat \bfB_k) \text{FN}(\widehat \bfB_k)}
{\sqrt{(\text{TP}(\widehat \bfB_k) + \text{FP}(\widehat \bfB_k))
(\text{TP}(\widehat \bfB_k) + \text{FN}(\widehat \bfB_k))
(\text{TN}(\widehat \bfB_k) + \text{FP}(\widehat \bfB_k))
(\text{TN}(\widehat \bfB_k) + \text{FN}(\widehat \bfB_k)) }}, \\
& \text{MCC} (\widehat \cB) = \frac{1}{K} \sum_{k=1}^K \text{MCC}(\widehat \bfB_k).
\end{align*}
\item Relative error in Frobenius norm-
\[
\text{RF} (\widehat \cB) = \frac{1}{K} \sum_{k=1}^K \frac{\| \hat \bfB^k - \bfB_0^k \|_F}{\| \bfB_0^k \|_F}.
\]
\end{itemize}
We use the same metrics to evaluate the precision matrix estimates $\widehat \Omega_y^k$ as well, with TPR and TNR calculations confined to off-diagonal entries.
\begin{table}[t!]
\centering
\scalebox{.9}{
    \begin{tabular}{ccccccc}
    \hline
    $(\pi_x, \pi_y)$ & $(p,q,n)$   & Method   & TPR            & TNR            & MCC & RF            \\ \hline
    $(5/p, 5/q)$   & (60,30,100)   & JMMLE    & 0.97(0.02)  & 0.99(0.003)  & 0.96(0.014) & 0.24(0.033) \\
    ~              & ~             & Separate & 0.96(0.018) & 0.99(0.004)  & 0.93(0.014) & 0.22(0.029) \\\cline{2-7}
    ~              & (30,60,100)   & JMMLE    & 0.97(0.013) & 0.99(0.002)  & 0.96(0.008) & 0.27(0.024) \\
    ~              & ~             & Separate & 0.99(0.009) & 0.99(0.003)  & 0.93(0.017) & 0.18(0.021) \\\cline{2-7}
    ~              & (200,200,150) & JMMLE    & 0.98(0.011) & 1.0(0)       & 0.99(0.005) & 0.16(0.025) \\
    ~              & ~             & Separate & 0.99(0.001) & 0.99 (0.001) & 0.88(0.009) & 0.18(0.007) \\\cline{2-7}
    ~              & (300,300,150) & JMMLE    & 1.0(0.001)  & 1.0(0)       & 0.99(0.001) & 0.14 (0.015)\\
    ~              & ~             & Separate & 1.0(0.001)  & 0.99(0.001)  & 0.84(0.01)  & 0.21(0.007)\\\hline
    $(30/p, 30/q)$ & (200,200,100) & JMMLE    & 0.97(0.017) & 1.0(0)       & 0.98(0.008) & 0.21(0.032) \\
    ~              & ~             & Separate & 0.32(0.01)  & 0.99(0.001)  & 0.49(0.009) & 0.85(0.06)  \\\cline{2-7}
    ~              & (200,200,200) & JMMLE    & 0.99(0.006) & 1.0(0)       & 0.99(0.007) & 0.13(0.016) \\
    ~              & ~             & Separate & 0.97(0.004) & 0.98(0.001)  & 0.93(0.002) & 0.19(0.07)  \\    \hline
    \end{tabular}
    }
    \caption{Table of outputs for estimation of regression matrices, giving empirical mean and standard deviation (in brackets) of each evaluation metric over 150 replications.}
    \label{table:simtable11}
\end{table}

\begin{table}[t!]
\centering
\scalebox{.9}{
\begin{tabular}{ccccccc}
    \hline
    $(\pi_x, \pi_y)$ & $(p,q,n)$   & Method   & TPR            & TNR             & MCC & RF            \\ \hline
    $(5/p, 5/q)$   & (60,30,100)   & JMMLE    & 0.76(0.018) & 0.90(0.006)  & 0.61(0.024)  & 0.32(0.008) \\
    ~              & ~             & Separate & 0.77(0.031) & 0.92(0.007)  & 0.56(0.03)   & 0.51(0.017) \\
    ~              & ~             & JSEM     & 0.24(0.013) & 0.8(0.003)   & 0.05(0.015)  & 1.03(0.002)\\\cline{2-7}
    ~              & (30,60,100)   & JMMLE    & 0.7(0.018)  & 0.94(0.002)  & 0.55(0.018)  & 0.3(0.005) \\
    ~              & ~             & Separate & 0.76(0.041) & 0.89(0.015)  & 0.59(0.039)  & 0.49(0.014) \\
    ~              & ~             & JSEM     & 0.13(0.005) & 0.9(0.001)   & 0.03(0.007)  & 1.04(0.001) \\\cline{2-7}
    ~              & (200,200,150) & JMMLE    & 0.68(0.017) & 0.98(0)      & 0.48(0.013)  & 0.26(0.002) \\
    ~              & ~             & Separate & 0.78(0.019) & 0.97(0.001)  & 0.55(0.012)  & 0.6(0.007) \\
    ~              & ~             & JSEM     & 0.05(0.002) & 0.97(0)      & 0.02(0.002)  & 1.01(0) \\\cline{2-7}
    ~              & (300,300,150) & JMMLE    & 0.71(0.014) & 0.98(0)      & 0.44(0.008)  & 0.25(0.002) \\
    ~              & ~             & Separate & 0.71(0.017) & 0.98(0.001)  & 0.51(0.011)  & 0.59(0.005) \\
    ~              & ~             & JSEM     & 0.04(0.002) & 0.98(0)      & 0.02(0.002)  & 1.01(0)     \\\hline
    $(30/p, 30/q)$ & (200,200,100) & JMMLE    & 0.77(0.016) & 0.98(0)      & 0.46(0.013)  & 0.31(0.003) \\
    ~              & ~             & Separate & 0.57(0.027) & 0.44(0.007)  & 0.04(0.008)  & 0.84(0.002)\\
    ~              & ~             & JSEM     & 0.05(0.002) & 0.97(0)      & 0.01(0.002)  & 1.01(0)     \\\cline{2-7}
    ~              & (200,200,200) & JMMLE    & 0.76(0.018)  & 0.98(0)     & 0.55(0.015)  & 0.27(0.004) \\
    ~              & ~             & Separate & 0.73(0.023) & 0.94(0.003)  & 0.39(0.017)  & 0.62(0.011)\\
    ~              & ~             & JSEM     & 0.05(0.002) & 0.97(0)      & 0.03(0.003)  & 1.01(0)     \\\hline
    \end{tabular}
}
\caption{Table of outputs for estimation of lower layer precision matrices over 150 replications.}
\label{table:simtable12}
\end{table}

Tables \ref{table:simtable11} and \ref{table:simtable12} summarize the results. For estimation of $\cB$, we compare our results to the method in \citet{LinEtal16} that estimates parameters in each of the $K$ two-layer structure separately, while for estimation of $\Omega_y$, we compare them with the results in \citet{LinEtal16} and using the single-layer JSEM \citep{MaMichailidis15} that estimates $\Omega_y$ assuming structured sparsity patterns and centered matrices $\bfY^k$, but not the data in the upper layer, i.e. $\cX$.

Our joint method has higher average MCC across all data settings than the separate method for the estimation of $\cB$, although TPR and TNR values are similar, except for $p=200, q=200, n=100$ where JMMLE has a much higher average TPR. For estimation of $\Omega_y$, incorporating information from the upper layer vastly improves performance, as demonstrated by the differences in performance between JMMLE and JSEM. For the 4 data settings with lower sparsity $(\pi_x = \pi = 5/p, \pi_y = 5/q)$, JMMLE produces sparser estimates compared to the separate method while estimating $\Omega_y$- as is evident from the lower TPR and MCC values. However, the RF values indicate that the quality of JMMLE estimates is significantly better. This in fact is a common pattern across the estimation of both $\cB$ and $\Omega_y$: JMMLE gives more accurate estimates across the methods, with lower average RF values across all data settings. Finally, for the estimation of $\Omega_y$, JMMLE does better in both of the higher sparsity settings $(\pi_x = \pi = 30/p, \pi_y = 30/q)$ across all metrics.
\begin{scriptsize}
\begin{table}[t!]
\centering
    \begin{tabular}{cccccc}
    \hline
    $(\pi_x, \pi_y)$ & $(p,q,n)$   & TPR$(\widehat \cB)$            & TNR$(\widehat \cB)$             & MCC$(\widehat \cB)$ & RF$(\widehat \cB)$    \\ \hline
    $(5/p, 5/q)$   & (60,30,100)   & 0.98 (0.01)   & 0.99 (0.002)   & 0.89 (0.017)  & 0.29 (0.014) \\
    ~              & (30,60,100)   & 0.94 (0.022)  & 0.99 (0.003)   & 0.93 (0.016)  & 0.31 (0.028) \\
    ~              & (200,200,150) & 0.99 (0.002)  & 0.99 (0)       & 0.98 (0.004)  & 0.17 (0.007) \\
    ~              & (300,300,150) & 0.99 (0.001)  & 1 (0)          & 0.99 (0.002)  & 0.15 (0.006) \\
    $(30/p, 30/q)$ & (200,200,100) & 0.99 (0.006)  & 1 (0)          & 0.98 (0.005)  & 0.2 (0.014)  \\
    ~              & (200,200,200) & 0.99 (0.009)  & 1 (0)          & 0.98 (0.005)  & 0.15 (0.017) \\\hline
    \hline
    $(\pi_x, \pi_y)$ & $(p,q,n)$   & TPR$(\widehat \Omega_y)$            & TNR$(\widehat \Omega_y)$             & MCC$(\widehat \Omega_y)$ & RF$(\widehat \Omega_y)$            \\ \hline
    $(5/p, 5/q)$   & (60,30,100)   & 0.71 (0.024)  & 0.90 (0.005)   & 0.64 (0.024)  & 0.34 (0.008)\\
    ~              & (30,60,100)   & 0.7 (0.019)   & 0.94 (0.002)   & 0.59 (0.014)  & 0.3 (0.004) \\
    ~              & (200,200,150) & 0.62 (0.012)  & 0.98 (0)       & 0.43 (0.009)  & 0.27 (0.003)\\
    ~              & (300,300,150) & 0.69 (0.013)  & 0.98 (0)       & 0.39 (0.008)  & 0.26 (0.02) \\
    $(30/p, 30/q)$ & (200,200,100) & 0.78 (0.024)  & 0.98 (0)       & 0.43 (0.012)  & 0.31 0.003) \\
    ~              & (200,200,200) & 0.69 (0.026)  & 0.98 (0.001)   & 0.5 (0.02)    & 0.29 (0.004)\\\hline
    \end{tabular}
    \caption{Table of outputs for joint estimation in presence of group misspecification}
    \label{table:simtable2}
\end{table}
\end{scriptsize}

\begin{scriptsize}
\begin{table}[t!]
\centering
    \begin{tabular}{ccc}
    \hline
    $(\pi_x, \pi_y)$ & $(p,q,n)$   & FDR          \\\hline
    $(5/p, 5/q)$   & (60,30,100)   & 0.19 (0.077) \\
    ~              & (30,60,100)   & 0.08 (0.064) \\
    ~              & (200,200,150) & 0.04 (0.016) \\
    ~              & (300,300,150) & 0.02 (0.007) \\\hline
    $(30/p, 30/q)$ & (200,200,100) & 0.03 (0.019) \\
    ~              & (200,200,200) & 0.03 (0.016) \\\hline
    \end{tabular}
    \caption{Table of outputs giving empirical FDR for estimating $\cB$ using JMMLE in presence of group misspecification}
    \label{table:simtable22}
\end{table}
\end{scriptsize}

\subsubsection{Effect of heterogeneity}
We repeat the above setups to check the performance of JMMLE in presence of within-group misspecification. For this task, we first set individual elements inside a non-zero group to be zero with probability 0.2 while generating the data, then pass the JMMLE estimates $\widehat \bfB^k$ through the FDR controlling thresholds as given in \eqref{eqn:fdr-threshold}. The results are summarized in Tables \ref{table:simtable2} and \ref{table:simtable22}. Across the simulation settings, values of all metrics are very close to the correctly specified counterparts in Table~\ref{table:simtable11}. Thus, the thresholding step 
proves largely effective. Also, in all cases the empirical FDR for estimating entries in $\cB$ is below 0.2. The performance is slightly worse than the correctly specified cases when estimating $\Omega_y$. This is expected, as the estimates $\widehat \Omega_y$ are obtained from neighborhood coefficients that are calculated based on the {\it pre-thresholding} coefficient estimates.
\begin{table}[t!]
\centering
\scalebox{.9}{
\begin{tabular}{ccccccc}
\hline
$(\pi_x, \pi_y)$  & $(p,q,n)$ & Method & \multicolumn{2}{c}{Global test} & \multicolumn{2}{c}{Simultaneous test} \\\cline{4-7}
& & & Power         & Size           & Power           & FDR \\\hline
    $(5/p, 5/q)$ & (60,30,100) & JMMLE    & 0.98 (0.016)  & 0.07 (0.011)   & 0.94 (0.023)   &  0.24 (0.027) \\
    ~ & ~         & Separate & 0.99 (0.007)  & 0.12 (0.02)    & 0.91 (0.025)   & 0.34(0.038)   \\
    ~ & ~         & SepLasso & 0.99 (0.007)  & 0.11 (0.02)    & 0.91 (0.025)  & 0.33(0.038)     \\\cline{2-7}
    ~ & (60,30,200) & JMMLE    & 0.99 (0.014)  & 0.07 (0.014)   & 0.97  (0.013)   & 0.22 (0.032)  \\
    ~ & ~         & Separate & 0.99 (0.005)  & 0.08 (0.014)   & 0.94 (0.019)   & 0.26(0.031)   \\
    ~ & ~         & SepLasso & 0.99 (0.004)  & 0.08 (0.014)   & 0.94 (0.019)  & 0.26(0.033)     \\\cline{2-7}
    ~ & (30,60,100) & JMMLE    & 0.98 (0.024)  &  0.07 (0.014)  & 0.92  (0.027)   & 0.24 (0.035)  \\
    ~ & ~         & Separate &  1 (0)         & 0.07 (0.015)    & 0.86 (0.036)   & 0.25(0.039)   \\
    ~ & ~         & SepLasso &  1 (0)         & 0.08 (0.014)    & 0.85 (0.036)   & 0.25(0.039)   \\\cline{2-7}
    ~ & (30,60,200) & JMMLE    & 0.99 (0.019)  &  0.08 (0.016)  & 0.96 (0.023)   & 0.24 (0.038)  \\
    ~ & ~         & Separate & 1 (0)          & 0.06 (0.013)   & 0.9 (0.038)   & 0.21(0.035)   \\
    ~ & ~         & SepLasso & 1 (0)          & 0.06 (0.012)   & 0.91 (0.038) & 0.21(0.034)      \\\cline{2-7}
    ~ & (200,200,150) & JMMLE    & 0.99 (0.006)  & 0.06 (0.003)   & 0.84 (0.011)   & 0.22 (0.007)  \\
    ~ & ~         & Separate & 1 (0)          &  0.2 (0.008)  &  0.93 (0.006)  & 0.46(0.009)   \\
    ~ & ~         & SepLasso & 1 (0)          &  0.2 (0.008)  &  0.93 (0.006) & 0.46(0.009)   \\\cline{2-7}
    ~ & (300,300,150) & JMMLE    & 0.99 (0.004)  &  0.07 (0.009)  & 0.54  (0.031)   & 0.34 (0.016)  \\
    ~ & ~         & Separate & 1 (0)          & 0.27 (0.01)    &  0.79 (0.007) & 0.58(0.008)   \\
    ~ & ~         & SepLasso & 1 (0)          & 0.27 (0.01)    &  0.79 (0.007) & 0.58(0.008)    \\\cline{2-7}
    ~ & (300,300,300) & JMMLE    & 0.99 (0.003)  & 0.03 (0.002)   & 0.99 (0.003)   & 0.12 (0.006)  \\
    ~ & ~         & Separate & 1 (0)          & 0.16 (0.005)   & 0.99 (0.004)   & 0.4 (0.007)   \\
    ~ & ~         & SepLasso & 1 (0)          & 0.16 (0.005)   & 0.99 (0.004)  & 0.4 (0.007)     \\\hline
  $(30/p, 30/q)$ & (200,200,100) & JMMLE & 0.99 (0.005)  & 0.112 (0.003)   & 0.41 (0.008)   & 0.52 (0.007)   \\
    ~ & ~         & Separate &  1 (0)         & 0.47 (0.008)   &  0.75 (0.007)  & 0.71(0.004)   \\
    ~ & ~         & SepLasso & 1 (0)          & 0.47 (0.008)   &  0.75 (0.007) & 0.71(0.004)     \\\cline{2-7}
    ~ & (200,200,200) & JMMLE    & 0.99 (0.004)  & 0.09 (0.004)   & 0.96 (0.006)   & 0.27 (0.008)   \\
    ~ & ~         & Separate &  1 (0)         & 0.42 (0.011)   & 0.98 (0.005)   & 0.63(0.006)   \\
    ~ & ~         & SepLasso & 1 (0)          & 0.42 (0.011)   & 0.98 (0.005)  & 0.63(0.006)   \\\cline{2-7}
    ~ & (200,200,300) & JMMLE    & 0.99 (0.002)  &  0.06 (0.003)  & 0.99  (0.004)   & 0.19 (0.008)  \\
    ~ & ~         & Separate & 1 (0)          & 0.27 (0.01)     & 0.99 (0.004)   & 0.52 (0.009)   \\
    ~ & ~         & SepLasso & 1 (0)          & 0.27 (0.01)     & 0.99 (0.004)   & 0.52 (0.009)   \\\hline
\end{tabular}
}
\caption{Table of outputs for global and simultaneous hypothesis testing.}
\label{table:testingtable}
\end{table}

\subsection{Simulation 2: testing}
\label{sec:eval-testing}
We slightly change the data generating model to evaluate our proposed global testing and FDR control procedure. We set $K=2$, then generate the $\bfB_0^1$ by first randomly assigning each of its element to be non-zero with probability $\pi$, then drawing values of those elements from $\text{Unif}\{ [ -1, -0.5] \cup [0.5,1]\}$ independently. After this we generate a matrix of differences $\bfD$, where $(\bfD)_{ij}, i \in \cI_p, j \in \cI_q$ takes values --1, 1, 0 with probabilities 0.1, 0.1 and 0.8, respectively. Finally we set $\bfB_0^2 = \bfB_0^1 + \bfD$. We set identical sparsity structures for the pairs of precision matrices $\{ \Omega_{x0}^1, \Omega_{x0}^2 \}$ and $\{ \Omega_{y0}^1, \Omega_{y0}^2 \}$. We use 150 replications of the above setup to calculate empirical power of global tests, as well as empirical power and FDR of simultaneous tests. To get the empirical sizes of global tests we use estimators obtained from applying JMMLE on a separate set of data generated setting all elements of $\bfD$ to 0. The type-I error of global tests is controlled at level 0.05, while FDR is set at 0.2 obtained by calculating the respective thresholds.

Table~\ref{table:testingtable} reports the empirical mean and standard deviations (in brackets) of all relevant quantities computed from debiased coefficients obtained from JMMLE, separate estimation, as well as from applying the original debiasing technique of \citet{ZhangZhang14} on $qK$ separate lasso estimates of row-level coefficient vectors, i.e. $\hat \bfb_i^k$. We report outputs for all combinations of data dimensions and sparsity used in Section~\ref{sec:eval-jmmle}, and also for increased sample sizes in each setting until a satisfactory FDR is reached. As expected from the theoretical analysis, higher sample sizes than those used in Section~\ref{sec:eval-jmmle} result in increased power for both global and simultaneous tests, and decreased size and FDR for all but one ($p=30, q=60$) of the settings. While separate estimation has slightly higher power in global testing, our joint method gives better results everywhere else. The empirical size of the JMMLE-based global tests remain slightly higher than the nominal level across the settings considered. This is in all likelihood a consequence of the higher sample size requirements in testing than estimation, as nominal sizes for JMMLE estimates tend to go down when $p,q$ are kept constant and $n$ is increased (for example, check setting 6 vs. setting 7 ($p = 300, ,q = 300, n = 150, 300)$, and setting 8 vs. 9 vs. 10 ($p = 200, ,q = 200, n = 100,200,300)$. This pattern is similar to empirical results in past proposals of high-dimensional testing methodology \citep{ref:JASA151658_WangPengLi,aispu}. However, the size estimates of JMMLE are much closer to the nominal level across data settings compared to either separate estimation or separate lasso, owing the fact that only JMMLE is able to leverage information across the different multilayer networks.

\subsection{Computation}
\label{sec:tricks-jmmle}
Next, we discuss some observations and strategies that speed up the JMMLE algorithm and reduce computation time significantly, especially for higher number of features in either layer.

\paragraph{Block update and refit $\bfB^k$ in each iteration.} Similar to the case of $K=1$ \citep{LinEtal16}, we use block coordinate descent {\it within} each $\bfB^k$. This means instead of the full update step \eqref{eqn:update-B} we perform the following steps in each iteration to speed up convergence:
$$
\left\{\widehat \bfB^{k (t+1)}_j \right\}_{k=1}^K =
\argmin_{\substack{\bfb_j^k \in \BR^p\\k \in \cI_K}} \left\{ \frac{1}{n} \sum_{j=1}^q \sum_{k=1}^K \| \bfY^k_j + \bfr_j^{k (t)} - \bfX^k \bfB_j^{k } \|^2
+ \lambda \sum_{h \in \cH} \| \bfB_j^{[h]} \| \right\},
$$
where $\bfr_1^{k (t)} = \widehat \bfE_{-1}^{k (t)} \widehat \bftheta_1^{k (t)}$, and
$$
\bfr_j^{k (t)} = \sum_{j'=1}^{j-1} \hat \bfe_j^{k (t+1)} \hat \theta_{jj'}^{k (t)} +
\sum_{j'=j+1}^{q} \hat \bfe_j^{k (t)} \hat \theta_{jj'}^{k (t)}
$$
for $j \geq 2$. Further, when starting from the initializer of the coefficient matrix given in \eqref{eqn:init-B}, the support set of coefficient estimates becomes constant after only a few ($< 10$) iterations of our algorithm, after which it refines the values inside the same support until overall convergence. This process speeds up significantly if a refitting step is added {\it inside each iteration} after the matrices $\widehat \bfB^k$ are updated:
\begin{align*}
\left\{\widetilde \bfB^{k (t+1)}_j \right\}_{k=1}^K &=
\argmin_{\substack{\bfb^k \in \BR^p\\k \in \cI_K}} \left\{ \frac{1}{n} \sum_{j=1}^q \sum_{k=1}^K \| \bfY^k_j + \bfr_j^{k (t)} - \bfX^k \bfB_j^{k } \|^2
+ \lambda \sum_{h \in \cH} \| \bfB^{[h]}_{-j} \| \right\}; \\
\widehat \bfB^{k (t+1)}_j &= \left[ (\bfX_{\cS_{jk}}^k)^T (\bfX_{\cS_{jk}}^k) \right]^- (\bfX_{\cS_{jk}}^k)^T \bfY_j^k,
\end{align*}
where $\cS_{jk} = \supp(\widetilde \bfB^{k (t+1)}_j)$.

\paragraph{One-step estimator.} Algorithm~\ref{algo:jmmle-algo}, even after the above modifications, is computation-intensive. The reason behind this is the full tuning and updating of the lower layer neighborhood estimates $\{ \widehat \Theta_j \}$ in each iteration. In practice, the algorithm speeds up significantly without compromising on estimation accuracy if we dispense of the $\Theta$ update step in all, but the last iteration. More precisely, we consider the following one-step version of the original algorithm.

\begin{Algorithm}
(The one-step JMMLE Algorithm)
\label{algo:jmmle-algo-1step}

\noindent 1. Initialize $\widehat \cB$ using \eqref{eqn:init-B}.

\noindent 2. Initialize $\widehat \Theta$ using \eqref{eqn:init-Theta}.

\noindent 3. Update $\widehat \cB$ as:
\begin{align*}
\widehat \cB^{(t+1)} &= \argmin_{\substack{\bfB^k \in \BM(p,q)\\k \in \cI_K}} \left\{ \frac{1}{n} \sum_{j=1}^q \sum_{k=1}^K \| \bfY^k_j - (\bfY_{-j}^k - \bfX^k \bfB_{-j}^k) \widehat \bftheta_j^{k (0)} - \bfX^k \bfB_j^{k } \|^2
+ \lambda_n \sum_{h \in \cH} \| \bfB^{[h]} \| \right\}
\end{align*}

\noindent 4. Continue till convergence to obtain $\widehat \cB = \{ \widehat \bfB^k \}$.

\noindent 5. Obtain $\widehat \bfE^k := \bfY^k - \bfX^k \widehat \bfB^k, k \in \cI_K$. Update $\widehat \Theta$ as:
\begin{align*}
\widehat \Theta_j = \argmin_{\Theta_j \in \BM(q-1, K)}
\left\{ \frac{1}{n} \sum_{k=1}^K
\| \widehat \bfE_j^k - \widehat \bfE_{-j}^k \bftheta_j^k \|^2
+ \gamma \sum_{j \neq j'} \sum_{g \in \cG_y^{jj'}} \| \bftheta_{jj'}^{[g]} \| \right\}
\end{align*}

\noindent 6. Calculate $\widehat \Omega_y^k, k \in \cI_K$ using \eqref{eqn:omega-y-calc}.
\end{Algorithm}

Compared to one-step algorithms based on first order approximation of the objective function \citep{ZouLi08,Taddy17}, we let $\cB$ converge completely, then use these solutions to recover the support set of the precision matrices. The estimation accuracy of $\Omega_y$ depends on the solution $\widehat \cB$ used to solve the sub-problem \eqref{eqn:EstEqn2} (Theorem~\ref{thm:thm-Theta} and Lemmas \ref{thm:ThetaThm} and \ref{prop:ErrorRE}). Thus, letting $\cB$ converge first ensures that the solutions $\widehat \Theta$ and $\widehat \Omega_y$ obtained subsequently are of a better quality compared to a simple early stopping of the JMMLE algorithm.
\begin{table}[t!]
\centering
    \begin{tabular}{cccccc}
    \hline
    $(p,q,n)$     & Method         & TPR$(\widehat \cB)$            & TNR$(\widehat \cB)$            & MCC$(\widehat \cB)$ & RF$(\widehat \cB)$             \\\hline
    (60,30,100) & Full         & 0.982 (0.013) & 0.994 (0.003) & 0.959 (0.014)   & 0.23 (0.021)  \\
    ~           & One step     & 0.971 (0.02)  & 0.996 (0.003) & 0.965 (0.014)   & 0.242 (0.033) \\
    (30,60,100) & Full         & 0.966 (0.015) & 0.991 (0.003) & 0.954 (0.008)   & 0.269 (0.026) \\
    ~           & One step     & 0.968 (0.013) & 0.992 (0.002) & 0.957 (0.008)   & 0.265 (0.024) \\ \hline
    \hline
    $(p,q,n)$     & Method         & TPR$(\widehat \Omega_y)$            & TNR$(\widehat \Omega_y)$            & MCC$(\widehat \Omega_y)$ & RF$(\widehat \Omega_y)$            \\\hline
    (60,30,100) & Full         & 0.756 (0.019) & 0.907 (0.005)  & 0.616 (0.021)   & 0.318 (0.007) \\
    ~           & One step     & 0.764 (0.018) & 0.904 (0.006)  & 0.678 (0.024)   & 0.321 (0.008) \\
    (30,60,100) & Full         & 0.695 (0.016) & 0.943 (0.002)  & 0.552 (0.015)   & 0.304 (0.005) \\
    ~           & One step     & 0.696 (0.018) & 0.943 (0.002)  & 0.552 (0.018)   & 0.304 (0.005) \\\hline
    \end{tabular}
    \caption{Comaprison of evaluation metrics for full and one-step versions of the JMMLE algorithm.}
    \label{table:simtable41}
\end{table}
\begin{table}[h!]
\centering
  \begin{tabular}{ccc}
    \hline
    $(p,q,n)$     & Method   & Comp. time (min) \\ \hline
    (60,30,100) & Full     & 6.1              \\ 
    ~           & One-step & 0.7              \\ \hline
    (30,60,100) & Full     & 22.4             \\ 
    ~           & One-step & 2.7              \\ \hline
    \end{tabular}
    \caption{Comaprison of computation times (averaged over 150 replications) for full and one-step versions of the JMMLE algorithm.}
    \label{table:simtable42}
\end{table}

We compared the performance of both versions of our algorithm for the two data settings with smaller feature dimensions. Computations were performed on the HiperGator supercomputer\footnote{\url{https://www.rc.ufl.edu/services/hipergator}}, in parallel across 8 cores of an Intel E5-2698v3 2.3GHz processor with 2GB RAM per core, the parallelization being done across the range of values for $\lambda_n$ within each replication. As seen in Table~\ref{table:simtable41}, performance is  indistinguishable across all the metrics, but the one-step algorithm saves a significant amount of computation time compared to the full version (Table~\ref{table:simtable42}).



\section{Real data example}
\label{sec:secreal}
We now apply the proposed methodology to breast cancer data obtained from The Cancer Genome Atlas\footnote{\url{https://www.genome.gov/Funded-Programs-Projects/Cancer-Genome-Atlas}}. The data set consists of mRNA and RNAseq expression values for $3980$ genes, divided into 88 pathways, for $n_1 = 262$ estrogen receptor positive or ER-positive (ER+) and $n_2 = 76$ ER-negative (ER--) breast cancer patients. 
As preprocessing steps, we consider single-pathway genes, and fit coordinate-wise lasso models to each column of the log-transformed response matrices $\bfY^k$, with $\bfX^k$ as predictors (say Lasso$(\bfy_{j}^k \sim \bfX^k)$). We then take the top 100 columns of each $\bfY^k$ that have lowest prediction errors (say $\cS^k$), and take unions of these column indices (i.e. $\cS := \cS^1 \cup \cS^2$) to construct the final response matrices $\bfY^k \rightarrow \bfY^k_\cS$. This gives us the final response dimension as $q = 166$. To select columns of $\bfX^k$, we take the top 200 predictor indices that have the highest mean absolute coefficient values across the lasso models on the selected response indices, i.e. Lasso$(\bfy_{j}^k \sim \bfX^k)$ where $j \in \cS$, and take the union of these indices. The resulting predictor dimension is $p = 339$.

Our objective here is to (a) obtain mRNA-mRNA, mRNA-RNAseq and RNAseq-RNAseq networks for the ER+ and ER-- groups while incorporating pathway information, (b) test for differential strengths of mRNA-RNAseq connections between the two sample groups. To this end, we take mRNA and RNAseq expression data as the top and bottom layers ($X$ and $Y$ in our nomenclature), respectively, and consider pathway-wise groups. Note that gene expression (data in the $Y$ layer) is controlled on two levels. First, transcription is controlled by limiting the amount of mRNA (data in the $X$ layer) that is produced from a particular gene. The second level of control is through post-transcriptional events that regulate the translation of mRNA into proteins. For comparison purposes, we apply JMMLE and the separate estimation method \citet{LinEtal16} for estimating $\cB$ and $\Omega_y$, and JSEM for estimating $\Omega_y$. For comparing estimation performances of the methods, we use the following performance metrics calculated over 100 random 80:20 train-test splits of samples within each group:
\begin{itemize}
\item Root Mean Squared Scaled Prediction Error:
$$ \text{RMSSPE}(\widehat \cB, \widehat \Omega_y) = \left[
\sum_{k=1}^K \frac{1}{n_k} \Tr \left(\bfY^k - \bfX^k \widehat \bfB^k)^T (\bfY^k - \bfX^k \widehat \bfB^k)
\widehat\Omega_y^k \right).
\right]^{1/2} $$

\item Proportion of non-zero coefficients in $\widehat \cB$:
$$ \text{NZ}(\widehat \bfB^k) = \frac{| \supp(\hat \bfB^k) |}{pq}; \quad
\text{NZ}(\widehat \cB) = \frac{1}{K} \sum_{k=1}^K \text{NZ}(\widehat \bfB^k). $$

\item Proportion of non-zero coefficients in off-diagonal entries of $\widehat \Omega_y$:
$$ \text{NZ}(\widehat \Omega_y^k) = \frac{| \supp(\hat \Omega_y^k) - q|}{q^2}; \quad
\text{NZ}(\widehat \Omega_y) = \frac{1}{K} \sum_{k=1}^K \text{NZ}(\widehat \Omega_y^k). $$
\end{itemize}

\begin{table}[t]
\centering
    \begin{tabular}{llll}
    \hline
~        & RMSSPE        & NZ$(\widehat \cB)$  & NZ$(\widehat \Omega_y)$  \\\hline
JMMLE    & 14.38 (3.27) & 0.014 (0.004) & 0.077 (0.009) \\
Separate & 17.1 (2.26)  & 8.8 $\times 10^{-5}$ (3.5 $\times 10^{-4}$) & 0.085 (0.077) \\
JSEM     & 18.19 (4.04) & 0 (0)             & 0.09 (0.002) \\\hline
    \end{tabular}
    \caption{Performance metric comparison over 100 random splits of the real data}
    \label{table:real-compare}
\end{table}
Table~\ref{table:real-compare} presents the comparison results. JMMLE and the separate estimation procedure obtain about the same amount of non-zero coefficients in $\Omega_y$ on average. Estimation of only the lower layer coefficients detects the most entries in the precision matrices $\widehat \Omega_y$, but has the highest prediction errors (calculated using $\widehat \cB = 0$). Separate estimation also hardly detects any non-zero elements in $\cB$, while JMMLE detects around 1.4\% of the inter-layer connection as non-zero. As a result, prediction errors are much lower for JMMLE.

\begin{table}[t!]
\centering
\scalebox{.95}{
\begin{tabular}{l|lll|lll}
    \hline
Sample group & \multicolumn{3}{c}{ER+ $(k=1)$} & \multicolumn{3}{c}{ER- $(k=2)$} \\\hline
    ~ & Value & mRNA      &  RNAseq   & Value  & mRNA      &  RNAseq   \\\cline{2-7}
    ~ & -5.87 & TAF9\_3   & TRA2B\_1  & -11.32 & TAF9\_3   & TRA2B\_1  \\
    ~ & -5.7  & KCNN3\_3  & THOC7\_1  & -6.01  & TAF9\_3   & UQCRQ\_1  \\
    ~ & 4.9   & SQRDL\_3  & COX6A1\_1 & -5.38  & TAF9\_3   & TAF9\_1   \\
    ~ & 4.35  & SQRDL\_3  & ATP5G3\_1 & 5.17   & SQRDL\_3  & COX6A1\_1 \\
Conections & -4.34 & KCNN3\_3  & PABPN1\_1 & 5.14   & SQRDL\_3  & ACTR3\_1  \\
in $\widehat \cB$ & 4.31  & SQRDL\_3  & ACTR3\_1  & 4.55   & SQRDL\_3  & SSU72\_1  \\
    ~ & -4.21 & KCNN3\_3  & SNRPD2\_1 & -4.52  & KCNN3\_3  & THOC7\_1  \\
    ~ & 3.98  & CYP7B1\_3 & ECH1\_1   & 4.4    & UNG\_3    & COX6A1\_1 \\
    ~ & 3.88  & SQRDL\_3  & SSU72\_1  & -4.18  & TAF9\_3   & ATP5J\_1  \\
    ~ & 3.87  & CYP7B1\_3 & FTH1\_1   & 4.17   & CYP7B1\_3 & FTH1\_1   \\ \hline
    \hline
    ~ & Value &    RNASeq1 &  RNAseq2  & Value &  RNAseq1 &  RNAseq2     \\\cline{2-7}
    ~ & -0.27 & ECH1\_1    & PIGY\_1   & -0.16 & THOC7\_1 & PABPN1\_1    \\
    ~ & -0.25 & THOC7\_1   & PABPN1\_1 & -0.12 & RBBP4\_1 & PABPN1\_1    \\
    ~ & -0.22 & COX6A1\_1  & SF3B5\_1  & -0.11 & NAPA\_1  & CD63\_1      \\
    ~ & -0.21 & PCBP1\_1   & SH3GL1\_1 & -0.11 & SOD1\_1  & SNRPD3\_1    \\
Conections & -0.21 & ECH1\_1    & DDX42\_1  & -0.1  & EIF3I\_1 & TXNL4A\_1    \\
in $\widehat \Omega_y$ & -0.19 & EXOSC2\_1  & QARS\_1   & -0.1  & PCBP1\_1 & SH3GL1\_1    \\
    ~ & -0.19 & QARS\_1    & PIGY\_1   & -0.1  & TAF9\_1  & COX7C\_1     \\
    ~ & -0.18 & ECH1\_1    & SDHC\_1   & -0.1  & ECH1\_1  & HNRNPA1L2\_1 \\
    ~ & -0.18 & PABPN1\_1  & VAMP8\_1  & -0.1  & KARS\_1  & FUNDC1\_1    \\
    ~ & -0.18 & EIF3I\_1   & TXNL4A\_1 & -0.1  & ECH1\_1  & PIGY\_1      \\ \hline
    \hline
    ~ & Value &    mRNA1 & mRNA2     & Value & mRNA1    & mRNA2     \\\cline{2-7}
    ~ & -0.32 & GP1BB\_3 & COX6A2\_3 & -0.19 & PTPRC\_3 & ITGAL\_3  \\
    ~ & -0.32 & PTTG1\_3 & PTTG2\_3  & -0.17 & PTPRC\_3 & CD2\_3    \\
    ~ & -0.3  & ABCA8\_3 & C7\_3     & -0.16 & PDCD1\_3 & ICOS\_3   \\
    ~ & -0.3  & PTPRC\_3 & CD2\_3    & -0.16 & PDCD1\_3 & CD2\_3    \\
Conections & -0.29 & ABCA8\_3 & FXYD1\_3  & -0.16 & PTPRC\_3 & CYBB\_3   \\
in $\widehat \Omega_x$ & -0.28 & PDCD1\_3 & ICOS\_3   & -0.15 & GP1BB\_3 & COX6A2\_3 \\
    ~ & -0.26 & PDCD1\_3 & CD2\_3    & -0.15 & PTPRC\_3 & IL2RG\_3  \\
    ~ & -0.25 & CD6\_3   & PDCD1\_3  & -0.15 & PTPRC\_3 & CTSS\_3   \\
    ~ & -0.25 & PTPRC\_3 & PTGER4\_3 & -0.14 & PTPRC\_3 & PTGER4\_3 \\
    ~ & -0.25 & LAT\_3   & PDCD1\_3  & -0.14 & LCP2\_3  & CYBB\_3   \\ \hline
    \end{tabular}
}
\caption{Top 10 within-layer and between-layer connections obtained by JMMLE.}
\label{table:top10}
\end{table}

To summarize within-layer and between-layer interactions, we consider the 10 highest entries in $\widehat \bfB^k, \widehat \Omega_y^k; k = 1,2$ in terms of absolute value. Table \ref{table:top10} gives their magnitudes, as well as the corresponding mRNA-RNAseq and RNAseq-RNAseq pairs. For the sake of comparison, we also report the same numbers and mRNA-mRNA pairs from the analysis of only the top layer using JSEM~\citep{MaMichailidis15}. According to our findings, the mRNA SQRDL\_3 is involved in downregulation of a number of RNA sequences in both groups of samples. Expression of the SQRDL gene is positively associated with high macrophage activity \citep{LyonsEtal17}, and its lower expression has been associated with breast cancer \citep{LiuEtal07,PiresEtal18}. In the ER+ group, KCNN3\_3 seems to be more heavily involved in doing so than ER--. This is also the case for TAF9\_3, but in the ER-- group vs. ER+. Considering the crucial roles of these genes in cancer cell migration, drug resistance (KCNN3, \citet{kcnn3}) and estrogen signalling (TAF9, \citet{taf9}), the evidence of differential expression may be significant in developing subtype-specific therapeutic targets.

\begin{table}[t!]

\begin{minipage}{.5\linewidth}
\caption*{(a)}
\centering
\begin{tabular}{ll}
\hline mRNA & Statistic \\\hline
DCTN2\_3   & 17015.2   \\ 
ST8SIA1\_3 & 13514.6   \\ 
FUT5\_3    &  8315.7   \\ 
XPA\_3     &  7194.2   \\ 
RETSAT\_3  &  5676.0   \\ 
TAF4B\_3   &  5385.8   \\ 
CYP7B1\_3  &  4189.6   \\ 
UNG\_3     &  3709.1   \\ 
RAD23A\_3  &  2793.8   \\ 
TAF9\_3    &  2427.1   \\ 
\hline
\end{tabular}
\end{minipage}
\begin{minipage}{.5\linewidth}
\caption*{(b)}
\centering
\begin{tabular}{lll}
\hline
mRNA & RNAseq & Statistic \\\hline
DCTN2\_3   & EIF4A1\_1  & 2560.6    \\ 
DCTN2\_3   & ARPC4\_1   & 2021.8    \\ 
ST8SIA1\_3 & EIF4A1\_1  & 1948.2    \\ 
DCTN2\_3   & PAIP1\_1   & 1922.3    \\ 
DCTN2\_3   & SNX5\_1    & 1825.1    \\ 
DCTN2\_3   & SUMO3\_1   & 1817.6    \\ 
DCTN2\_3   & CETN2\_1   & 1779.2    \\ 
DCTN2\_3   & SF3B4\_1   & 1755.8    \\ 
ST8SIA1\_3 & ARPC4\_1   & 1516.5    \\ 
ST8SIA1\_3 & PAIP1\_1   & 1453.9    \\\hline
\end{tabular}
\end{minipage}
\caption{Hypothesis testing outputs from real data analysis: (a) top-10 mRNAs and their global test statistic ($D_i$) values, (b) top-10 mRNA-RNAseq pairs and their simultaneous test statistic ($d_{ij}$) values}
\label{table:realtesting}
\end{table}

After applying our debiasing procedure and performing the global test, 23 mRNA-s were determined to have significant differences in the corresponding rows across sample groups, i.e. between $\widehat \bfb_i^1$ and $\widehat \bfb_i^2$. Within connections of these mRNAs, 957 total mRNA-RNAseq connections were determined by the simultaneous testing procedure to have significant differences between their corresponding coefficients, i.e. between $\hat b_{ij}^1$ and $\hat b_{ij}^2$. Table~\ref{table:realtesting} summarizes the top-10 statistic values in each situation. The DCTN2\_3 mRNA shows significant differential interactions with a number of RNA sequences---up-regulation of the DCTN2 gene has previously been found to be associated with chemotherapy resistance in breast cancer patients \citep{dctn2}. 

\section{Discussion}
\label{sec:sec5}
This work introduces an integrative framework for knowledge discovery in multiple multi-layer Gaussian Graphical Models. We exploit {\it a priori} known structural similarities across parameters of the multiple models to achieve estimation gains compared to separate estimation. More importantly, we derive results on the asymptotic distributions of generic estimates of the multiple regression coefficient matrices in this complex setup, and perform global and simultaneous testing for pairwise differences within the between-layer edges.

\subsection{Performance improvement}
The JMMLE algorithm due to the incorporation of prior information about sparsity patterns improves on the theoretical convergence rates of the estimation method for {\it single} multi-layer GGMs (i.e. $K=1$) introduced in \citet{LinEtal16}. With our initial estimates, the method of \citet{LinEtal16} achieves the following convergence rates for the estimation of $\cB$ and $\Omega_y$, respectively (using Corollary 4 therein):
\begin{align*}
\| \widehat \bfbeta - \bfbeta_0 \|_F & \leq \sum_{k=1}^K O \left( \sqrt{ \frac{ b_k \log (p q)}{n}} \right), \\
\sum_{k=1}^K \| \widehat \Omega_y^k - \Omega_{y0}^k \|_F & \leq O \left( K \sqrt{ \frac{ (S + q) \log (pq)}{n}} \right).
\end{align*}
In comparison, JMMLE has the following rates:
\begin{align*}
\| \widehat \bfbeta - \bfbeta_0 \|_F & \leq O \left( \sqrt{ \frac{ |h_{\max}| B \log (p q)}{n}} \right), \\
\sum_{k=1}^K \| \widehat \Omega_y^k - \Omega_{y0}^k \|_F & \leq O \left( \sqrt{ \frac{ KS |g_{\max}| \log (p q)}{n}} \right).
\end{align*}
For $\cB$, joint estimation outperforms separate estimation when group sizes are small, so that $(| h_{\max}| B)^{1/2} < \sum_k b_k^{1/2}$. The estimation gain for $\Omega_y$ is more substantial, especially for higher values of $q$. This is corroborated by our simulation outputs (Tables \ref{table:simtable11} and \ref{table:simtable12}), where the joint estimates perform better for both sets of parameters, but the differences between RF errors obtained from joint and separate estimates tend to be lower for $\widehat \Omega_y$ than $\widehat \cB$.

\subsection{Remaining challenges}
Our proposed framework for inference in complex multilayer networks still presents a couple of challenges, which can benefit from theoretical work pursued in the following directions. First, the use of tuning parameter selection criteria requires a rigorous analysis, not only to select parameter estimates that exhibit good finite sample performance, but also to ensure that the selected estimates when plugged into the hypothesis testing procedures result in tests and confidence intervals with accurate size, power or coverage guarantees. A number of existing methods have given consistency results of the (extended) BIC based tuning parameter selection procedures in high-dimensional regression \citep{FanTang13, WangKimLi13} or graphical model \citep{FoygelDrton10,GaoEtal12} setups. Using theoretical tools provided therein and generalizing or adapting their conditions to multilayer settings is a possible avenue that can be explored. To maintain focus on the current problem, we defer this to future work. Secondly, extending JMMLE to include overlapping within- or between-layer groups is of interest to tackle practical situations like multiple pathways sharing a number of common genes. The current algorithm involves fitting multiple group lasso models (using the R package \texttt{grpreg}), which can be replaced by alternative methods that can handle overlapping groups.


\subsection{Extensions}
There are two immediate extensions of our hypothesis testing framework.

\vspace{1em}
\noindent (I) In recent work, \citet{Liu17} proposed a framework to test for structural similarities and differences across multiple {\it single layer} GGMs. For $K$ GGMs with precision matrices $\Omega^k = (\omega_{ii'}^k)_{i,i' \in \cI_p}$, a test for the partial correlation coefficients $\rho_{ii'}^{k} = - \omega_{ii'}^{k} / \sqrt{\omega_{ii}^{k} \omega_{i'i'}^{k}}$ using residuals from $pK$ separate penalized neighborhood regressions is developed, one for each variable of each GGM. To incorporate structured sparsity across $k$, our simultaneous regression techniques for all neighborhood coefficients (i.e. \eqref{eqn:jsem-model} and \eqref{eqn:EstEqn2}) can be used instead, to perform testing on the between-layer edges. Theoretical properties of this procedure can be derived using results in \citet{Liu17}, possibly with adjustments for our neighborhood estimates to adhere to the rate conditions for the constants $a_{n1}, a_{n2}$ therein to account for a diverging $(p,q,n)$ setup.

\noindent (II) For $K > 2$, detection of the following sets of inter-layer edges can be scientifically significant:
\begin{align*}
\cB_1 &= \left\{ (i,j): \sum_{1 \leq k < k' \leq K} \left(b_{0,ij}^k - b_{0,ij}^{k'} \right)^2 > 0; i \in \cI_p, j \in \cI_q \right\}\\
\cB_2 &= \left\{ (i,j): b_{0,ij}^1 = \cdots, b_{0,ij}^K \neq 0 \right\}\\
\cB_3 &= \left\{ (i,j): b_{0,ij}^1 = \cdots, b_{0,ij}^K = 0 \right\}
\end{align*}
e.g. detection of gene-protein interactions that are present, but may have different or same weights across $k$ ($\cB_1$ and $\cB_2$, respectively), and that are absent for all $k$ ($\cB_3$). The asymptotic result in Theorem~\ref{Thm:ThmTesting} continues to hold in this situation, and an extension of the global test (Algorithm~\ref{algo:AlgoGlobalTest}) is immediate. However, extending the FDR control procedure requires a technically more involved approach.

\paragraph*{}
The strength of our proposed debiased estimator \eqref{eqn:DebiasedBeta} is that only generic estimates of relevant model parameters that satisfy general rate conditions are necessary to obtain a valid asymptotic distribution. This translates to a high degree of flexibility in choosing the method of estimation. Our formulation based on sparsity assumptions (Section~\ref{sec:algosection}) is a specific way (motivated by applications in Omics data integration) to obtain the necessary estimates. Sparsity may not be an assumption that is required or even valid in complex hierarchical structures from different domains of application. For different two-layer components in such multi-layer setups, low-rank, group-sparse or sparse methods (or a combination thereof) can be plugged into our alternating algorithm. Results analogous to those in Section~\ref{sec:jmmle-theory} need to be established for the corresponding estimators. However, as long as these estimators adhere to the convergence conditions (T1)-(T3), Theorem~\ref{Thm:ThmTesting} can be used to derive the asymptotic distributions of between-layer edges.

Finally, extending our framework to non-Gaussian data and graph Laplacian structures is of interest. As seen for the $K=1$ case in \citet{LinEtal16}, their alternating block algorithm continues to give comparable results under shrunken or truncated empirical distributions of Gaussian errors. Similar results may be possible in the general case, and improvements can come from modifying different parts of the estimation algorithm. For example, the estimation of the precision matrices based on restricted support sets using log-likelihoods in \eqref{eqn:omega-y-calc} can be replaced by methods like nonparanormal estimation \citep{LiuLaffertyWasserman09} or regularized score matching \citep{LinDrtonShojaie16}. For graph Laplacian structures, generalization of recent work on multilayer models \citep{BayramEtal20,KumarEtal20} in the lines of the JMMLE framework may be explored.

\bibliography{jmmle-draft-JMLR} 

\newpage
\appendix
\section*{Appendix}
\section{Proofs of main results}

\begin{proof}[Proof of Theorem~\ref{thm:algo-convergence}]
The theorem is a generalization of Theorem 1 in \citet{LinEtal16}. The proof follows directly from the proof of that theorem, substituting $( \widehat B^{(k)}, \widehat \Theta_\epsilon^{(k)}), (B^*, \Theta_\epsilon^*)$ and $ (B^\infty, \Theta_\epsilon^\infty)$ therein with $(\widehat \cB^{(t)}, \widehat \Theta_y^{(t)}), ( \cB_0, \Theta_{y 0})$ and $ (\cB^\infty, \Theta_y^\infty)$, respectively, and their corresponding variations as required.

\end{proof}

We use the following condition extensively while deriving the results that follow.

\vspace{1em}
\noindent {\bf Condition 3} (Restricted eigenvalues). A symmetric matrix $\bfM \in \BM(b,b)$ is said to satisfy the restricted eigenvalue or RE condition with parameters $\psi, \phi >0$, denoted as curvature and tolerance, respectively, if
$$
\bftheta^T \bfM \bftheta \geq \psi \| \bftheta \|^2 - \phi \| \bftheta \|_1^2
$$
for all $\bftheta \in \BR^b$. In short, this is denoted by $\bfM \sim RE(\psi, \phi)$.

Starting from \citet{BickelRitovTsybakov09}, different versions of the RE conditions have been proposed and used in high-dimensional analysis \citep{LohWainwright12,BasuMichailidis15,MaMichailidis15,vandeGeerBuhlmann09} to ensure that a covariance matrix satisfies a somewhat relaxed positive-definiteness condition.

\begin{proof}[Proof of Theorem~\ref{thm:thm-Theta}]
The proof strategy is as follows. We first show that given fixed $(\cX,\cE)$, and some conditions on $\widetilde \bfE^k := \bfY^k - \bfX^k \widetilde \bfB^k, k \in \cI_K$, the bounds in Theorem~\ref{thm:thm-Theta} hold. We then show that for random $(\cX,\cE)$, those conditions hold with probability approaching 1.

\begin{lemma}\label{thm:ThetaThm}
Assume fixed $\cX, \cE$ and deterministic $\widetilde \cB = \{ \widetilde \bfB^k \}$, and the following conditions.

\noindent{\bf(A1)} For $k \in \cI_K$,
$$
\| \widetilde \bfB^k - \bfB^k_0 \|_1 \leq C_\beta \sqrt{\frac{\log (pq)}{n}}
$$
with $C_\beta$ is non-negative and depends on $\cB_0$ only.


\noindent{\bf(A2)} For all $j \in \cI_q$,
$$
\frac{1}{n} \left\| (\widetilde \bfE_{-j}^k)^T \widetilde \bfE^k \bfT_{0,j}^k \right\|_\infty \leq
\BQ \left(C_\beta, \Sigma_{x0}^k, \Sigma_{y0}^k \right) \sqrt {\frac{ \log (p q)}{n}},
$$
where $\BQ \left(C_\beta, \Sigma_{x0}^k, \Sigma_{y0}^k \right)$ is non-negative and depends on $\cB_0, \Sigma_{x0}^k$ and $\Sigma_{y0}^k$ only.

\noindent{\bf(A3)} Denote $\widetilde \bfS^k = (\widetilde \bfE^k)^T \widetilde \bfE^k/n$. Then $\widehat \bfS^k \sim RE(\psi^k, \phi^k)$ with $Kq \phi \leq \psi/2$ where $ \psi = \min_k \psi^k, \phi = \max_k \phi^k $.


Then the following hold

\noindent (I) Given the choice of tuning parameter
$$
\gamma_n \geq 4 \sqrt{| g_{\max}|} \BQ_0 \sqrt {\frac{ \log (p q)}{n}}; \quad
\BQ_0 := \max_{k \in \cI_K} \BQ \left(C_\beta, \Sigma_{x0}^k, \Sigma_{y0}^k  \right)
$$
\begin{align}
\| \widehat \Theta_j - \Theta_{0,j} \|_F & \leq 12 \sqrt{s_j} \gamma_n / \psi, \label{eqn:theta-norm-bound-1a}\\
\sum_{j \neq j', g \in \cG_y^{jj'}} \| \hat \bftheta_{jj'}^{[g]} - \bftheta_{0,jj'}^{[g]} \| & \leq 48 s_j \gamma_n / \psi. \label{eqn:theta-norm-bound-2a}\\
| \supp (\widehat \Theta_j) | & \leq 128 s_j / \psi
\end{align}

\noindent (II) For the choice of tuning parameter $\gamma_n = 4 \sqrt{| g_{\max}|} \BQ_0 \sqrt{\log (p q)/n}$,
\begin{align}\label{eqn:OmegaBounds0a}
\frac{1}{K} \sum_{k=1}^K \| \widehat \Omega_y^k - \Omega_{y0}^k \|_F \leq
O \left( \BQ_0 \sqrt{\frac{| g_{\max}| S}{K}}
\sqrt {\frac{ \log (p q)}{n}} \right)
\end{align}
%
%
\end{lemma}

Condition (A1) holds by assumption. When $\cX$ and $\cE$ are random, the following proposition ensures that (A2) and (A3) hold with probabilities approaching to 1.

\begin{lemma}\label{prop:ErrorRE}
Consider deterministic $\widetilde \cB$ satisfying assumption (A1), and conditions (E1), (E2) from the main paper. Then for sample size $n \succsim \log (pq)$ and $k \in \cI_K$,

\begin{enumerate}
\item $\widetilde \bfS^k$ satisfies the RE condition: $ \widetilde \bfS^k \sim RE (\psi^k, \phi^k)$, where 
$$
\psi^k = \frac{ \Lambda_{\min} (\Sigma_{x0}^k)}{2}; \quad \phi^k = \frac{ \psi^k \log p}{n} + 2 C_\beta c_2 [ \Lambda_{\max} (\Sigma_{x0}^k) \Lambda_{\max} (\Sigma_{y0}^k) ]^{1/2} \frac{ \log(pq)}{n}
$$
with probability $\geq 1 - 6c_1 \exp [-(c_2^2-1) \log(pq)] - 2 \exp (- c_3 n), c_1, c_3 > 0, c_2 > 1$.
\item The following deviation bound is satisfied for any $j \in \cI_q$
$$
\left\|\frac{1}{n} (\widetilde \bfE_{-j}^k)^T \widetilde \bfE^k \bfT_{0,j}^k \right\|_\infty \leq \BQ \left(C_\beta, \Sigma_{x0}^k, \Sigma_{y0}^k \right) \sqrt {\frac{ \log (p q)}{n}}
$$

with probability $\geq 1 - 1/p^{\tau_1-2} - 12 c_1 \exp [-(c_2^2-1) \log(pq)] - 6c_4 \exp [-(c_5^2-1) \log(pq)], c_4 > 0, c_5 > 1, \tau_1 > 2$, where
\begin{align*}
\BQ \left(C_\beta, \Sigma_{x0}^k, \Sigma_{y0}^k \right) &=
\left[ 2 C_\beta^2 V_x^k + 4 C_\beta c_2 [ \Lambda_{\max} (\Sigma_{x0}^k) \Lambda_{\max} (\Sigma_{y0}^k) ]^{1/2} \right] \sqrt{\frac{ \log(pq)}{n}} +\\
& c_5 \left[ \Lambda_{\max} ( \Sigma_{y0}^k) \sigma_{y0,j,-j}^k \right]^{1/2} \sqrt{\frac{\log q}{\log (pq)}}
\end{align*}
with $\sigma_{y0,j,-j}^k = Var( E_j - \BE_{-j} \bftheta_{0,j})$, and 
$$
V_x^k = \sqrt{ \frac{ \log 4 + \tau_1 \log p}{c_x^k n}}; \quad
c_x^k = \left[ 128 ( 1 + 4 \Lambda_{\max} (\Sigma_{x0}^k)  )^2 \max_i (\sigma_{x0,ii}^k)^2 \right]^{-1}
$$
\end{enumerate} 
\end{lemma}

\noindent We prove the main theorem by putting together Lemma~\ref{thm:ThetaThm} and Lemma~\ref{prop:ErrorRE}, and simplifying the constants $c_1 := 12 c_1, c_4 := 6c_4$.
\end{proof}

\begin{proof}[Proof of Theorem~\ref{thm:thm-B}]
The strategy is the same as in Theorem~\ref{thm:thm-Theta}. We first establish the theorem statements hold for fixed $\cX, \cE$ in the presence of certain regularity conditions, and then show that those conditions are satisfied with probability approaching 1 when $\cX$ and $\cE$ are random.


\begin{lemma}\label{thm:BetaThm}
Assume fixed $(\cX, \cE)$, and deterministic $\widetilde \Theta = \{ \widetilde \Theta_j \}$, so that

\noindent{\bf(B1)} For $j \in \cI_q$,
$$
\| \widetilde \Theta_j - \Theta_{0,j} \|_F \leq C_\Theta \sqrt{\frac{\log q}{n}},
$$
for some $C_\Theta$ dependent on $\Theta_0$ only.

\noindent{\bf(B2)} Denote $\widetilde \bfGamma^k = (\widetilde \bfT^k)^2 \otimes (\bfX^k)^T \bfX^k/n, \widetilde \bfgamma^k = (\widetilde \bfT^k)^2 \otimes (\bfX^k)^T \bfY^k/n$. Then the deviation bound holds:
$$
\left\| \widetilde \bfgamma^k - \widetilde \bfGamma^k \bfbeta_0 \right\|_\infty \leq \BR( C_\Theta, \Sigma_{x0}^k, \Sigma_{y0}^k) \sqrt{ \frac{ \log(pq)}{n}}.
$$
where $\BR (C_\Theta, \Sigma_{x0}^k, \Sigma_{y0}^k )$ depends on $\Theta_0, \Sigma_{x0}^k$ and $\Sigma_{y0}^k$ only, and $\{ \widetilde \bfT^k \}$ are defined using $\widetilde \Theta$ similar to \eqref{eqn:define-T}.

\noindent{\bf(B3)} $\widetilde \bfGamma \sim RE(\psi_*, \phi_*)$ with $Kpq \phi_* \leq \psi_*/2$.

Then, given the choice of the tuning parameter
$$
\lambda_n \geq 4 \sqrt{| h_{\max} |} \BR_0 \sqrt{ \frac{ \log(pq)}{n}}; \quad 
\BR_0 := \max_{k \in \cI_K} \BR \left(C_\Theta, \Sigma_{x0}^k, \Sigma_{y0}^k \right)
$$
the following holds
\begin{align}
\| \widehat \bfbeta - \bfbeta_0 \|_1 & \leq 48 \sqrt{ | h_{\max} |} B \lambda_n / \psi^* \label{eqn:BetaThmEqn1}\\
\| \widehat \bfbeta - \bfbeta_0 \| & \leq 12 \sqrt B \lambda_n / \psi^* \label{eqn:BetaThmEqn2}\\
\sum_{h \in \cH} \| \bfbeta^{[h]} - \bfbeta_0^{[h]} \| & \leq 48 B \lambda_n / \psi^* \label{eqn:BetaThmEqn3}\\
(\widehat \bfbeta - \bfbeta_0 )^T \widehat \bfGamma (\widehat \bfbeta - \bfbeta_0 ) & \leq
72 B \lambda_n^2 / \psi^* \label{eqn:BetaThmEqn4}
\end{align}
\end{lemma}

Condition (B1) holds by assumption. Next, we verify that conditions (B2) and (B3) hold with high probability given fixed $\widehat \Theta$.

\begin{lemma}\label{prop:ThmBetaRE}
Consider deterministic $\widehat \Theta$ satisfying assumption (B1). Also assume conditions (E3), (E4) from the main body of the paper. Then, for sample size $n \succsim \log (pq)$,

\begin{enumerate}
\item $\widetilde \bfGamma$ satisfies the RE condition: $ \widetilde \bfGamma \sim RE (\psi_*, \phi_*)$, where 
$$
\psi_* = \min_k \psi^k \left( \min_i \psi_k^j - d_k C_\Theta \sqrt{ \frac{ \log (pq)}{n}}\right), 
\phi_* = \max_k \phi^k \left( \min_i \phi_k^j + d_k C_\Theta \sqrt{ \frac{ \log (pq)}{n}}\right)
$$
with probability $\geq 1 - 2 \exp(c_3 n), c_3>0$.
\item The deviation bound in (B2) is satisfied with probability $ \geq 1 - 12 c_1 \exp[ (c_2^2-1) \log (pq)]$, where
$$
\BR \left(C_\Theta, \Sigma_{x0}^k, \Sigma_{y0}^k \right) =
c_2 \left\{ d_k C_\Theta \sqrt{ \frac{ \log (pq)}{n}}
[ \Lambda_{\max} (\Sigma_{x0}^k) \Lambda_{\max} (\Sigma_{y0}^k)]^{1/2} +
\left[ \frac{\Lambda_{\max} (\Sigma_{x0}^k)}{\Lambda_{\min} (\Sigma_{y0}^k) } \right]^{1/2} \right\}
$$
\end{enumerate}
\end{lemma}

The theorem follows straighforwardly by putting together the results from Lemmas~\ref{thm:BetaThm} and ~\ref{prop:ThmBetaRE}, and simplifying the constants $c_1 := 12 c_1, c_4 := 6c_4$.
\end{proof}

\begin{proof}[Proof of Theorem~\ref{thm:starting-values}]
The first part is immediate from the proof of part I of Theorem 4 in \citet{LinEtal16}. By choice of $\lambda_n$, we now have
$$
\| \widehat \bfB^{k (0)} - \bfB^k_0 \|_1 = O\left( \sqrt{ \frac{ \log (pq)}{n}} \right),
$$
so we can apply Theorem~\ref{thm:thm-Theta} to prove the bounds on $\{\widehat \Theta_j^{(0)} \}$.
\end{proof}

\begin{proof}[Proof of Theorem~\ref{Thm:ThmTesting}]
Define the following:
$$
\widehat \bfD_i = \ve(\widehat \bfb^1_i, \ldots, \widehat \bfb^K_i); \quad
\bfR_i^k = \bfX_i^k - \bfX_{-i}^k \widehat \bfzeta_i^k; k \in \cI_K
$$
Then, from \eqref{eqn:DebiasedBeta} we have
\begin{align}\label{eqn:ThmTestingProofeq1}
\bfM_i ( \widehat \bfC_i - \widehat \bfD_i )^T &= \frac{1}{\sqrt n}
\begin{bmatrix}
\frac{1}{\widehat s^1_i} (\bfR^1_i)^T \widehat \bfE^1 \\
\vdots\\
\frac{1}{\widehat s^K_i} (\bfR^K_i)^T \widehat \bfE^K
\end{bmatrix}
\end{align}
We now decompose $\widehat \bfE^k:$
\begin{align*}
\widehat \bfE^k &= \bfY^k - \bfX^k \widehat \bfB^k \\
&= \bfE^k + \bfX^k (\bfB_0^k - \widehat \bfB^k)\\
&= \bfE^k + \bfX_i^k (\bfb_{0i}^k - \widehat \bfb_i^k) + \bfX_{-i}^k (\bfB_{0,-i}^k - \widehat \bfB_{-i}^k)
\end{align*}
Putting them back in \eqref{eqn:ThmTestingProofeq1} and using $t_i^k = (\bfR_i^k)^T \bfX_i^k/n$, we get
\begin{align}
\bfM_i ( \widehat \bfC_i - \widehat \bfD_i)^T &= \frac{1}{\sqrt n}
\begin{bmatrix}
\frac{1}{\widehat s^1_i} (\bfR^1_i)^T \bfE^1 \\
\vdots\\
\frac{1}{\widehat s^K_i} (\bfR^K_i)^T \bfE^K
\end{bmatrix} +
\bfM_i (\bfD_i - \widehat \bfD_i )^T \notag\\
& + \frac{1}{\sqrt n}
\begin{bmatrix}
\frac{1}{\widehat s^1_i} (\bfR^1_i)^T \bfX_{-i}^1 (\bfB_{0,-i}^1 - \widehat \bfB_{-i}^1) \\
\vdots\\
\frac{1}{\widehat s^K_i} (\bfR^K_i)^T \bfX_{-i}^K (\bfB_{0,-i}^K - \widehat \bfB_{-i}^K)
\end{bmatrix} \notag\\
\Rightarrow
\widehat \Omega_y^{1/2} \bfM_i ( \widehat \bfC_i - \bfD_i)^T &=
\frac{\widehat \Omega_y^{1/2}}{\sqrt n}
\begin{bmatrix}
\frac{1}{\widehat s^1} (\bfR^1_i)^T \bfE^1 \\
\vdots\\
\frac{1}{\widehat s^K} (\bfR^K_i)^T \bfE^K
\end{bmatrix} +
\frac{\widehat \Omega_y^{1/2}}{\sqrt n}
\begin{bmatrix}
\frac{1}{\widehat s^1_i} (\bfR^1_i)^T \bfX_{-i}^1 (\bfB_{0,-i}^1 - \widehat \bfB_{-i}^1) \\
\vdots\\
\frac{1}{\widehat s^K_i} (\bfR^K_i)^T \bfX_{-i}^K (\bfB_{0,-i}^K - \widehat \bfB_{-i}^K)
\end{bmatrix}\label{eqn:ThmTestingProofeq2}
\end{align}

At this point, we drop $k$ and 0 in the subscripts since there is no ambiguity, and establish the following:

\begin{lemma}\label{Lemma:ThmTestingLemma}
Given conditions (T1) and (T2), the following holds for sample size $n$ such that $n \succsim \log (pq)$:
$$
\frac{1}{\sqrt n \widehat s_i}  \widehat \Omega_y^{1/2} \bfE^T \bfR_i \sim
\cN_q ({\bf 0}, \bfI) + \bfS_{1n};
$$
\begin{align}\label{eqn:ThmTestingProofeq3}
\| \bfS_{1n} \|_\infty & \leq 
\frac{D_\Omega^{1/2} (2 + D_\zeta) c_2 [ \Lambda_{\max} (\Sigma_x) \Lambda_{\max} (\Sigma_e) ]^{1/2} \sqrt{ \log(pq)}}{\sqrt{ \sigma_{x,i,-i}} - n^{-1/4} - D_\zeta \sqrt{V_x}} =
O \left( \frac{ \log(pq)}{\sqrt n} \right)
\end{align}
with probability $\geq 1 - 6c_1 \exp[-(c_2^2-1) \log (p q)] - 1/p^{\tau_1-2} - \kappa_i/\sqrt n$, where $\kappa_i := Var [(X_i - \BX_{-i} \bfzeta_{0,-i})^2]$.

Additionally, given condition (T3) we have
\begin{align}\label{eqn:ThmTestingProofeq4}
& \left\| \frac{1}{\sqrt n \widehat s_i} \bfR_i^T \bfX_{-i} (\bfB_{-i} - \widehat \bfB_{-i})
\widehat \Omega_y^{1/2} \right\|_\infty \notag\\
& \leq
\frac{D_\beta ( \Lambda_{\min} (\Sigma_y)^{1/2} + D_\Omega^{1/2})}{\sigma_{x,i,-i} - n^{-1/2} - D_\zeta \sqrt{V_x}} 
\left[ c_{7} \sqrt{ (\sqrt{ \sigma_{x,i,-i}} \Lambda_{\max} (\Sigma_{x, -i}) ) \log p} + \sqrt n D_\zeta V_x \right] =
O \left( \frac{ \log(pq)}{\sqrt n} \right)
\end{align}
holds with probability $\geq 1 - 6c_6 \exp[-(c_7^2-1) \log (p q)] - 1/p^{\tau_1-2} - \kappa_i/\sqrt n$ for some $c_6 > 0, c_7 > 1$.
\end{lemma}

Given Lemma~\ref{Lemma:ThmTestingLemma}, the first and second summands on the right hand side of \eqref{eqn:ThmTestingProofeq2} are bounded above by applying each of \eqref{eqn:ThmTestingProofeq3} and \eqref{eqn:ThmTestingProofeq4} $K$ times. This completes the proof.
\end{proof}

\begin{proof}[Proof of Theorem~\ref{thm:PowerThm}]
From \eqref{eqn:ThmTestingProofeq2} and Lemma~\ref{Lemma:ThmTestingLemma} we have that
\begin{align}
(\widehat \Omega_y^k)^{1/2} m_i^k ( \widehat \bfc_i^k - \bfb_{0i}^k) \sim \cN_q ({\bf 0}, \bfI) + \bfS_{2n}^k,
\end{align}
where $\| \bfS_{2n}^k \|_\infty = o_P(1)$. We next obtain the following lemma:
\begin{lemma}\label{Lemma:PowerThmLemma}
Drop $k$ in superscripts and 0 in subscripts. Given condition (T1), the following holds with probability $\geq 1 - 6c_{6} \exp [-(c_{7}^2-1) \log (p-1)] - 1/p^{\tau_2-2} - \kappa_i / \sqrt n, \tau_2 > 2$:
\begin{align}
\left| \frac{m_i}{\sqrt n} - \sqrt{ \sigma_{x,i,-i}} \right| & \leq
\delta_i := \sqrt{ \frac{ \log 4 + \tau_2}{c_i n}} +
\frac{D_\zeta + 1 }{\sqrt{ \sigma_{x,i,-i}} - n^{-1/2} - D_\zeta \sqrt{V_x}} \times \notag\\
& \left[ c_{7} [(\sigma_{x,i,-i} \Lambda_{\max} (\Sigma_{x, -i})]^{1/2} \sqrt{ \frac{\log p}{n}} + D_\zeta V_x \right], \label{eqn:PowerThmLemmaEqn}
\end{align}
where $c_i = [ 128 (1 + 4 \sigma_{x,i,-i})^2 (\sigma_{x,i,-i})^2 ]^{-1}$, and the sample size satisfies $n \succsim \log p$.
\end{lemma}

We also provide the following general result:

\begin{lemma}\label{lemma:omega-diff}
Consider two positive definite matrices $\bfA, \bfA_1 \in \BM(a,a)$. Then, for $\delta > 0$, we have
$$
\| \bfA  -\bfA_1 \|_\infty \leq \delta \Rightarrow \| \bfA^{1/2} - \bfA_1^{1/2} \|_\infty \leq \delta^{1/2}.
$$
\end{lemma}

\noindent Applying Lemma~\ref{lemma:omega-diff} it follows immediately from assumption (T2) that
\begin{align}\label{eqn:omega-sqrt-bound}
\left\| \widehat \Omega_y^{1/2} - \Omega_y^{1/2} \right\|_\infty \leq D_{\Omega}^{1/2}
\end{align}

Using Lemma~\ref{Lemma:PowerThmLemma} in conjunction with \eqref{eqn:omega-sqrt-bound} we now have
\begin{align}
\sqrt n (\Omega_{y0}^k)^{1/2} \sqrt{\sigma_{x0,i,-i}^k} (\widehat \bfc_i^k - \bfb_{0i}^k)  \sim
\cN_q ( {\bf 0}, \bfI) + \bfS_{3n}^k \notag\\
\Rightarrow \sqrt n \Sigma_i^{-1/2} (\widehat \bfc_i^1 - \widehat \bfc_i^2 - \bfdelta) \sim
\cN_q \left( {\bf 0}, \bfI \right) + \bfS_{3n},
\label{eqn:PowerThmProofEqn1}
\end{align}
where $\Sigma_i := \Sigma_{y0}^1/ \sigma_{x0,i,-i}^1 + \Sigma_{y0}^2/ \sigma_{x0,i,-i}^2$ and $\bfS_{3n} = \bfS_{3n}^1 - \bfS_{3n}^2, \|\bfS_{3n}^k\|_\infty = o_P(1)$. We now break down the left hand side above as
\begin{align}
\sqrt n \Sigma_i^{-1/2} (\widehat \bfc_i^1 - \widehat \bfc_i^2 - \bfdelta) &=
\sqrt n \Sigma_i^{-1/2} \widehat \Sigma_i^{1/2} \widehat \Sigma_i^{-1/2} (\widehat \bfc_i^1 - \widehat \bfc_i^2) - \sqrt n \Sigma_i^{-1/2} \bfdelta \notag\\
&= (\Sigma_i^{-1/2} \widehat \Sigma_i^{1/2} - \bfI) . \sqrt n \widehat \Sigma_i^{-1/2} (\widehat \bfc_i^1 - \widehat \bfc_i^2) + \notag\\
& \sqrt n \widehat \Sigma_i^{-1/2} (\widehat \bfc_i^1 - \widehat \bfc_i^2) - \sqrt n \Sigma_i^{-1/2} \bfdelta,
\label{eqn:PowerThmProofEqn2}
\end{align}
with
$$
\widehat \Sigma_i :=
\frac{ n \widehat \Sigma_y^1}{(m_i^1)^2} + \frac{n \widehat \Sigma_y^2}{(m_i^2)^2}.
$$
Next, we obtain the following lemma:
\begin{lemma}\label{Lemma:PowerThmLemma2}
Given conditions (T1) and (T2), for the pooled covariance matrix estimate $\widehat \Sigma_i$, we have
$$
\left\| \widehat \Sigma_i - \Sigma_i \right\|_\infty = o(1),
$$
for sample size $n \succsim \log p$.
\end{lemma}
Lemma~\ref{lemma:omega-diff} now implies that $\| \widehat \Sigma_i^{1/2} - \Sigma_i^{1/2} \|_\infty = o(1)$. Putting this in the first summand of \eqref{eqn:PowerThmProofEqn2}, then using \eqref{eqn:PowerThmProofEqn1} we get
$$
\sqrt n \widehat \Sigma_i^{-1/2} (\widehat \bfc_i^1 - \widehat \bfc_i^2) - \sqrt n \Sigma_i^{-1/2} \bfdelta
\sim \cN_q \left( {\bf 0}, \bfI \right) + \bfS_{4n},
$$
with $\| \bfS_{4n} \|_\infty = o_P(1)$. The power of the global test follows as a consequence. Finally, the lower bound on the order of $\| \bfdelta \|$ holds because $n \bfdelta^T \Sigma_i^{-1} \bfdelta \geq n \| \bfdelta \|^2 \Lambda_{\min} (\Sigma_i^{-1}) $, and
$$
\Lambda_{\min} (\Sigma_i^{-1}) = \frac{ \Lambda_{\max} (\Sigma_{y0}^1)}{\sigma_{x,i,-i}^1} +
\frac{ \Lambda_{\max} (\Sigma_{y0}^2)}{\sigma_{x,i,-i}^2}.
$$
\end{proof}

\begin{proof}[Proof of Theorem~\ref{thm:FDRthm}]
The proof follows the general structure of Theorem 4.1 in \citet{LiuShao14}, with two modifications. Firstly, we replace the bound in equation (12) of \citet{LiuShao14} by a new deviation bound
$$
P \left( \left| d_{ij} - \frac{\mu_j}{\sigma_j} \right| \geq t \right) = (1 - \Phi(t))(1 + o(1))
$$
for any $t$, since $(d_{ij} - \mu_j)/\sigma_j \sim N(0,1) + o_P(1)$ from Corollary~\ref{corollary:CorTesting}. We replace $G_\kappa(t)$ in all following calculations in \citet{LiuShao14} with $1 - \Phi(t)$. Secondly, we need to ensure that given both $\Sigma_{y0}^1$ and $\Sigma_{y0}^2$ satisfy the condition (D1) or (D1*), the pooled covariance matrix $\Sigma_{y0}^1/ \sigma_{x0,i,-i}^1 + \Sigma_{y0}^2/  \sigma_{x0,i,-i}^2$ also does so.

For this, denote $c_k =  \sigma_{x0,i,-i}^k, k = 1,2$. Notice that for any $C_1, C_2 > 0$,
\begin{align*}
r_{jj'}^k \geq C_k & \Rightarrow \sigma_{y0,jj'}^k \geq (\sigma_{y0,jj}^k \sigma_{y0,j'j'}^k)^{1/2} C_k\\
& \Rightarrow \frac{\sigma_{y0,jj'}^1}{c_1} + \frac{\sigma_{y0,jj'}^2}{c_2} \geq
\frac{ (\sigma_{y0,jj}^1 \sigma_{y0,j'j'}^1)^{1/2} C_1}{c_1} +
\frac{ (\sigma_{y0,jj}^2 \sigma_{y0,j'j'}^2)^{1/2} C_2}{c_2}\\
& \Rightarrow \frac{ \sigma_{y0,jj'}^1/c_1 + \sigma_{y0,jj'}^2/c_2}
{ (\sigma_{y0,jj}^1 \sigma_{y0,j'j'}^1)^{1/2}/ c_1 + (\sigma_{y0,jj}^2 \sigma_{y0,j'j'}^2)^{1/2}/ c_2}
\geq \min \{ C_1, C_2 \}.
\end{align*}
It now follows that (D1) or (D1*) holds for the pooled covariance matrices.
\end{proof}

\section{Proofs of auxiliary results}

\begin{proof}[Proof of Lemma~\ref{thm:ThetaThm}]
The proof has the same structure as the proof of Theorem 1 in \citet{MaMichailidis15}, where consistency of the (single layer) JSEM estimates are established. Part (I) is analogous to part A.1 therein, but the proof strategy is completely different, which we provide in detail next. Our part (II) follows along similar lines as parts A.2 and A.3, incorporating the updated quantities from the first part (A.1). For this part of the proof, we provide an outline and leave details to the reader.

\paragraph{Proof of part (I).}
%
In its reparametrized version, (\ref{eqn:EstEqn2}) becomes
\begin{align}
\widehat \bfT_j = \argmin_{\bfT_j} \left\{ \frac{1}{n} \sum_{k=1}^K \| (\bfY^k - \bfX^k \widehat \bfB^k) \bfT_j^k \|^2 + \gamma_n \sum_{j \neq j', g \in \cG_y^{jj'}} \| \bfT_{jj'}^{[g]} \| \right\}
\end{align}
with $\bfT_{jj'}^{[g]} := (T_{jj'}^k)_{k \in g}$. Now for any $\bfT_j \in \BM(q, K)$ we have
$$
\frac{1}{n} \sum_{k=1}^K \| (\bfY^k - \bfX^k \widehat \bfB^k) \widehat \bfT_j^k \|^2 + \gamma_n \sum_{j \neq j', g \in \cG_y^{jj'}} \| \widehat \bfT_{jj'}^{[g]} \| \leq
\frac{1}{n} \sum_{k=1}^K \| (\bfY^k - \bfX^k \widehat \bfB^k) \bfT_j^k \|^2 + \gamma_n \sum_{j \neq j', g \in \cG_y^{jj'}} \| \bfT_{jj'}^{[g]} \|
$$
For $\bfT_j = \bfT_{0,j}$ this reduces to
\begin{align}\label{eqn::Thm1ProofProp2Eqn1}
\sum_{k=1}^K (\bfd_j^k)^T \widehat \bfS^k \bfd_j^k & \leq - 2 \sum_{k=1}^K (\bfd_j^k)^T \widehat \bfS^k \bfT_{0,j}^k + \gamma_n \sum_{j \neq j', g \in \cG_y^{jj'}} \left( \| \bfT_{jj'}^{[g]} \| -  \| \bfT_{jj'}^{[g]} + \bfd_{jj'}^{[g]}\| \right)
\end{align}
with $\bfd_{j}^k := \widehat \bfT_j^k - \bfT_{0,j}^k$ etc. For the $k^\text{th}$ summand in the first term on the right hand side, since $d_{jj}^k = 0$, $\widehat \bfE^k \bfd_j^k = \widehat \bfE_{-j}^k \bfd_{-j}^k$. Thus
\begin{align*}
\sum_{k=1}^K \left| (\bfd_j^k)^T \widehat \bfS^k \bfT_{0,j}^k \right| &=
\sum_{k=1}^K \left| \bfd_j^k. \frac{1}{n} (\widehat \bfE^k)^T \widehat \bfE^k \bfT_{0,j}^k \right| \\
& \leq \sum_{k=1}^K \left\| \frac{1}{n} (\widehat \bfE_{-j}^k)^T \widehat \bfE^k \bfT_{0,j}^k \right\|_\infty \| \bfd_{-j}^k \|_1 \\
& \leq \left[ \sum_{j \neq j', g \in \cG_y^{jj'}} \| \bfd_{jj'}^{[g]} \| \right]
\BQ_0 \sqrt {| g_{\max}|} \sqrt{ \frac{\log (pq)}{n}}
\end{align*}
by assumption (A2). For the second term, suppose $\cS_{0,j}$ is the support of $\Theta_{0,j}$, i.e. $\cS_{0,j} = \{ (j',g): \bftheta_{jj'}^{[g]} \neq 0 \}$. Then
\begin{align*}
\sum_{j \neq j', g \in \cG_y^{jj'}} \left( \| \bfT_{jj'}^{[g]} \| -  \| \bfT_{jj'}^{[g]} + \bfd_{jj'}^{[g]}\| \right) & \leq
\sum_{(j',g) \in \cS_{0,j}} \left( \| \bfT_{jj'}^{[g]} \| -  \| \bfT_{jj'}^{[g]} + \bfd_{jj'}^{[g]}\| \right) -
\sum_{(j',g) \notin \cS_{0,j}} \| \bfd_{jj'}^{[g]} \|\\
& \leq \sum_{(j',g) \in \cS_{0,j}} \| \bfd_{jj'}^{[g]} \| - \sum_{(j',g) \notin \cS_{0,j}} \| \bfd_{jj'}^{[g]} \|
\end{align*}
so that by choice of $\gamma_n$, (\ref{eqn::Thm1ProofProp2Eqn1}) reduces to
\begin{align}\label{eqn:Thm1ProofProp2Eqn3}
\sum_{k=1}^K (\bfd_j^k)^T \widehat \bfS^k \bfd_j^k & \leq 
\frac{\gamma_n}{2} \left[ \sum_{(j',g) \in \cS_{0,j}} \| \bfd_{jj'}^{[g]} \| + \sum_{(j',g) \notin \cS_{0,j}} \| \bfd_{jj'}^{[g]} \| \right] +
\gamma_n \left[ \sum_{(j',g) \in \cS_{0,j}} \| \bfd_{jj'}^{[g]} \| - \sum_{(j',g) \notin \cS_{0,j}} \| \bfd_{jj'}^{[g]} \| \right] \notag\\
& = \frac{3 \gamma_n}{2} \sum_{(j',g) \in \cS_{0,j}} \| \bfd_{jj'}^{[g]} \| - \frac{\gamma_n}{2} \sum_{(j',g) \notin \cS_{0,j}} \| \bfd_{jj'}^{[g]} \| \notag\\
& \leq \frac{3 \gamma_n}{2} \sum_{j \neq j', g \in \cG_y^{jj'}} \| \bfd_{jj'}^{[g]} \|
\end{align}
Since the left hand side is $\geq 0$, this also implies
$$
\sum_{(j',g) \notin \cS_{0,j}} \| \bfd_{jj'}^{[g]} \| \leq 3 \sum_{(j',g) \in \cS_{0,j}} \| \bfd_{jj'}^{[g]} \| \quad \Rightarrow
\sum_{j \neq j', g \in \cG_y^{jj'}} \| \bfd_{jj'}^{[g]} \| \leq
4 \sum_{(j',g) \in \cS_{0,j}} \| \bfd_{jj'}^{[g]} \| \leq 4 \sqrt{s_j} \| \bfD_j \|_F
$$
with $\bfD_j = (\bfd_j^1, \ldots, \bfd_j^K)$. Now the RE condition on $\widehat \bfS^k$ means that
$$
\sum_{k=1}^K (\bfd_j^k)^T \widehat \bfS^k \bfd_j^k \geq 
\sum_{k=1}^K \left( \psi_k \| \bfd_j^k \|^2 - \phi_k \| \bfd_j^k \|_1^2 \right) \geq
\psi \| \bfD_j \|_F^2 - \phi \| \bfD_j \|_1^2 \geq 
(\psi - Kq \phi ) \| \bfD_j \|_F^2 \geq \frac{\psi}{2}  \| \bfD_j \|_F^2
$$
by assumption (A3).

Combining the above with \eqref{eqn:Thm1ProofProp2Eqn3}, we finally have
\begin{align}\label{eqn:Thm1ProofProp2Eqn2}
\frac{\psi}{3} \| \bfD_j \|_F^2 & \leq
\gamma_n \sum_{j \neq j', g \in \cG_y^{jj'}} \| \bfd_{jj'}^{[g]} \| \leq
4 \gamma_n \sqrt{s_j} \| \bfD_j\|_F
\end{align}
Since
$$
(\bfD_j)_{j',k} = \widehat T_{jj'}^k - T_{0,jj'}^k = \begin{cases}
0 \text{ if } j = j'\\
-(\widehat \theta_{jj'}^k - \theta_{0,jj'}^k ) \text{ if } j \neq j'
\end{cases}
$$
The bounds in \eqref{eqn:theta-norm-bound-1a} and \eqref{eqn:theta-norm-bound-2a} are obtained by replacing the corresponding elements in (\ref{eqn:Thm1ProofProp2Eqn2}).

For the bound on $| \widehat \cS_j| := |\supp( \widehat \Theta_j)|$, notice that if $\hat \bftheta_{jj'}^{[g]} \neq 0$ for some $(j',g)$,
\begin{align*}
\frac{1}{n} \sum_{k \in g} \left| ((\widehat \bfE_{-j}^k)^T \widehat \bfE^k ( \widehat \bfT_j^k - \bfT_{0,j}^k ))^{j'} \right| & \geq
\frac{1}{n} \sum_{k \in g} \left| ((\widehat \bfE_{-j}^k)^T \widehat \bfE^k \widehat \bfT_j^k )^{j'} \right| - \frac{1}{n} \sum_{k \in g} \left| ((\widehat \bfE_{-j}^k)^T \widehat \bfE^k \bfT_{0,j}^k )^{j'} \right|\\
& \geq |g| \gamma_n - \sum_{k \in g} \BQ ( C_\beta, \Sigma_x^k, \Sigma_y^k ) \sqrt{ \frac{ \log (pq)}{n}}
\end{align*}
using the KKT condition for (\ref{eqn:EstEqn2}) and assumption (A2). The choice of $\gamma_n$ now ensures that the right hand side is $\geq 3|g| \gamma_n / 4$. Hence,
\begin{align*}
| \hat \cS_j| & \leq \sum_{(j',g) \in \widehat \cS_j} \frac{16}{9 n^2 |g|^2 \gamma_n^2 } \sum_{k \in g} \left| ((\widehat \bfE_{-j}^k)^T \widehat \bfE^k ( \widehat \bfT_j^k - \bfT_{0,j}^k ))^{j'} \right|^2\\
& \leq \frac{16}{9 \gamma_n^2} \sum_{k=1}^K \frac{1}{n} \left\| (\widehat \bfE_{-j}^k)^T \widehat \bfE^k ( \widehat \bfT_j^k - \bfT_{0,j}^k ) \right\|^2 \\
& = \frac{16}{9 \gamma_n^2} \sum_{k=1}^K (\bfd_j^k)^T \widehat \bfS^k \bfd_j^k \\
& \leq \frac{8}{3 \gamma_n} \sum_{j \neq j', g \in \cG_y^{jj'}} \| \bfd_{jj'}^{[g]} \| \leq \frac{128 s_j}{\psi} 
\end{align*}
using (\ref{eqn:Thm1ProofProp2Eqn3}) and (\ref{eqn:Thm1ProofProp2Eqn2}).

\paragraph{Proof of part (II).}
We denote the selected edge set for the $k^\text{th}$ Y-network by $\hat E^k$. Denote its population version by $E_0^k$. Further, let
$$
\tilde \Omega_y^k = \diag (\Omega_{y0}^k) + \Omega_{y, E_0^k \cap \hat E^k}^k
$$
Based on similar derivations as in the proof of Corollary A.1 in \citet{MaMichailidis15}, the following two upper bounds can be established:
\begin{align}
| \hat E^k | \leq \frac{ 128 S }{\psi} \label{eqn:Thm1ProofProp2Bd4}\\
\frac{1}{K} \sum_{k=1}^K \| \tilde \Omega_y^k - \Omega_{y0}^k \|_F \leq
\frac{12 c_y \sqrt{S} \gamma_n} {\sqrt K \psi} \label{eqn:Thm1ProofProp2Bd5}
\end{align}
following which, taking $\gamma_n = 4 \sqrt{| g_{\max}|} \BQ_0 \sqrt{ \log (pq)/ n}$,
\begin{align}
\Lambda_{\min} ( \tilde \Omega_y^k) \geq d_y - \frac{48 c_y \BQ_0 \sqrt{| g_{\max}| S}}{ \psi}
\sqrt{ \frac{ \log (pq)}{n}} \geq (1 - t_1) d_y > 0 \label{eqn:Thm1ProofProp2Bd6}\\
\Lambda_{\max} ( \tilde \Omega_y^k) \leq c_y + \frac{48 c_y \BQ_0 \sqrt{| g_{\max}| S}}{ \psi}
\sqrt{ \frac{ \log (pq)}{n}} \leq c_y + t_1 d_y < \infty \label{eqn:Thm1ProofProp2Bd7}
\end{align}
with $0 < t_1 < 1$, and the sample size $n$ satisfying
$$
n \geq | g_{\max}| S \left[ \frac{48 c_y \BQ_0}{\psi t_1 d_y} \right]^2 \log (pq).
$$

Following the same steps as part A.3 in the proof of Theorem 4.1 in \citet{MaMichailidis15}, it can be proven using \eqref{eqn:Thm1ProofProp2Bd4}--\eqref{eqn:Thm1ProofProp2Bd7} that
$$
\sum_{k=1}^K \left\| \widehat \Omega_y^k - \tilde \Omega_y^k \right\|_F^2 \leq
O \left( \BQ_0^2 | g_{\max}| S \frac{ \log (pq)}{n} \right)
$$
The proof is now complete by combining this with \eqref{eqn:Thm1ProofProp2Bd5}  and then applying the Cauchy-Schwarz inequality and the triangle inequality.
\end{proof}

\begin{proof}[Proof of Lemma~\ref{prop:ErrorRE}]


We drop the subscript 0 for true values and the superscript $k$ since there is no scope of ambiguity. For part 1, we start with an auxiliary lemma:
\begin{lemma}\label{lemma:ErrorRElemma1}
For a sub-Gaussian design matrix $\bfX \in \BM(n,p)$ with columns having mean ${\bf 0}_p$ and covariance matrix $\Sigma_x$, the sample covariance matrix $\widehat \Sigma_x = \bfX^T \bfX/n$ satisfies the RE condition
$$
\widehat \Sigma_x \sim RE \left( \frac{\Lambda_{\min} ( \Sigma_x) }{2}, \frac{\Lambda_{\min} ( \Sigma_x) \log p }{2 n} \right)
$$
with probability $\geq 1 - 2 \exp(-c_3 n)$ for some $c_3 > 0$.
\end{lemma}
Denote $\widehat \bfE = \bfY - \bfX \widehat \bfB$. For $\bfv \in \BR^q$, we have
\begin{align}\label{eqn:ErrorREeqn3}
\bfv^T \widehat \bfS \bfv &= \frac{1}{n} \| \widehat \bfE \bfv \|^2 \notag\\
&= \frac{1}{n} \| (\bfE + \bfX ( \bfB_0 - \widehat \bfB ))\bfv \|^2 \notag\\
&= \bfv^T \bfS \bfv + \frac{1}{n} \| \bfX (\bfB_0 - \widehat \bfB) \bfv \|^2 + 2 \bfv^T (\bfB_0 - \widehat \bfB)^T \left( \frac {(\bfX)^T \bfE}{n} \right) \bfv
\end{align}
For the first summand, $ \bfv^T \bfS^k \bfv \geq \psi_y \| \bfv \|^2 - \phi_y \| \bfv \|_1^2$ with $\psi_y = \Lambda_{\min} (\Sigma_y)/2, \phi_y = \psi_y \log p/n$ by applying Lemma \ref{lemma:ErrorRElemma1} on $\bfS$. The second summand is greater than or equal to 0. For the third summand,
$$
2 \bfv^T (\bfB_0 - \widehat \bfB)^T \left( \frac {(\bfX)^T \bfE}{n} \right) \bfv \geq
-2 C_\beta \left\| \frac {(\bfX)^T \bfE}{n} \right\|_\infty \| \bfv \|_1^2
\sqrt{ \frac{ \log (pq)}{n}}
$$
by assumption (A1). Now, we use another lemma:
\begin{lemma}\label{lemma:ErrorRElemma2}
For zero-mean independent sub-gaussian matrices $\bfX \in \BM(n,p), \bfE \in \BM(n,q)$ with parameters $(\Sigma_x, \sigma_x^2)$ and $(\Sigma_e, \sigma_e^2)$ respectively, given that $n \succsim \log(pq)$ the following holds with probability $\geq 1 - 6c_1 \exp [-(c_2^2-1) \log(pq)]$ for some $c_1 >0, c_2 > 1$:
$$
\frac{1}{n} \| \bfX^T \bfE \|_\infty \leq c_2 [ \Lambda_{\max} (\Sigma_x) \Lambda_{\max} (\Sigma_e) ]^{1/2} \sqrt{\frac{ \log(pq)}{n}}
$$
\end{lemma}
Subsequently we collect all summands in (\ref{eqn:ErrorREeqn3}) and get
$$
\bfv^T \widehat{ \bfS} \bfv \geq \psi_y \| \bfv \|^2 - \left( \phi_y + 2 C_\beta c_2 [ \Lambda_{\max} (\Sigma_x) \Lambda_{\max} (\Sigma_y) ]^{1/2} \frac{ \log(pq)}{n} \right) \| \bfv \|_1^2
$$
with probability $\geq 1 - 2\exp(- c_3 n) - 6c_1 \exp [-(c_2^2-1) \log(pq)]$. This concludes the proof of part 1.

To prove part 2, we decompose the quantity in question:
\begin{align}\label{eqn:ErrorRElemma2maineqn}
\left\| \frac{1}{n} \widehat \bfE_{-j}^T \widehat \bfE \bfT_{0,j} \right\|_\infty &=
\left\| \frac{1}{n} \left[ \bfE_{-j} + \bfX (\bfB_{0,j} - \widehat \bfB_j) \right]^T \left[ \bfE + \bfX (\bfB_0 - \widehat \bfB) \right] \bfT_{0,j} \right\|_\infty \notag\\
& \leq \left\| \frac{1}{n} \bfE_{-j}^T \bfE \bfT_{0,j} \right\|_\infty +
\left\| \frac{1}{n} \bfE_{-j}^T \bfX (\bfB_0 - \widehat \bfB) \bfT_{0,j} \right\|_\infty \notag\\
& + \left\| \frac{1}{n} (\bfB_{0,j} - \widehat \bfB_j)^T \bfX^T \bfX (\bfB_0 - \widehat \bfB) \bfT_{0,j} \right\|_\infty +
\left\| \frac{1}{n} (\bfB_{0,j} - \widehat \bfB_j)^T \bfX^T \bfE \bfT_{0,j} \right\|_\infty \notag\\
&= \| \bfW_1 \|_\infty + \| \bfW_2 \|_\infty + \| \bfW_3 \|_\infty + \| \bfW_4 \|_\infty
\end{align}
Now
$$
\bfW_1 = \frac{1}{n} \bfE_{-j}^T ( \bfE_j - \bfE_{-j} \bftheta_{0,j})
$$
For node $j$ in the $y$-network, $\BE_{-j}$ and $E_j - \BE_{-j} \bftheta_{0,j}$ are the neighborhood regression coefficients and residuals, respectively. Thus they are orthogonal, so we can apply Lemma \ref{lemma:ErrorRElemma2} on $\bfE_{-j}$ and $\bfE_j - \bfE_{-j} \bftheta_{0,j}$ to obtain that for $n \succsim \log (q-1)$,
\begin{align}\label{eqn:ErrorRElemma2eqn5}
\| \bfW_1 \|_\infty & \leq c_5 \left[ \Lambda_{\max} ( \Sigma_{y,-j}) \sigma_{y,j,-j} \right]^{1/2} \sqrt{\frac{\log(q-1)}{n}}
\end{align}
holds with probability $\geq 1 - 6c_4 \exp [-(c_5^2-1) \log(pq)]$ for some $c_4 > 0, c_5 > 1$.

For $\bfW_2$ and $\bfW_4$, identical bounds hold:
\begin{align*}
\| \bfW_2 \|_\infty & \leq \left\| \frac{1}{n} \bfE_{-j}^T \bfX (\bfB_0 - \widehat \bfB) \right\|_\infty \| \bfT_{0,j} \|_1 \leq
\left\| \frac{1}{n} \bfE^T \bfX \right\|_\infty \| \bfB_0 - \widehat \bfB \|_1 \| \bfT_{0,j} \|_1\\
\| \bfW_4 \|_\infty & \leq \left\| \frac{1}{n} (\bfB_{0,j} - \widehat \bfB_j)^T \bfX^T \bfE \right\|_\infty \| \bfT_{0,j} \|_1 \leq
\left\| \frac{1}{n} \bfE^T \bfX \right\|_\infty \| \bfB_0 - \widehat \bfB \|_1 \| \bfT_{0,j} \|_1\\
\end{align*}
Since $\Omega_y$ is diagonally dominant, $|\omega_{y,jj}| \geq \sum_{j \neq j'} |\omega_{y,jj'}|$ for any $j \in \cI_q$. Hence
$$
\| \bfT_{0,j} \|_1 = \sum_{j'=1}^q | T_{jj'} | = 1 + \sum_{j \neq j'} | \theta_{jj'} | = 1 + \frac{1}{\omega_{y,jj}} \sum_{j \neq j'} | \omega_{y,jj'} | \leq 2
$$
so that for $n \succsim \log (pq)$,
\begin{align}\label{eqn:ErrorRElemma2eqn6}
\| \bfW_2 \|_\infty + \| \bfW_4 \|_\infty  & \leq
4 C_\beta c_2 [ \Lambda_{\max} (\Sigma_x) \Lambda_{\max} (\Sigma_y) ]^{1/2} \frac{ \log(pq)}{n}
\end{align}
with probability $\geq 1 - 12 c_1 \exp [-(c_2^2-1) \log(pq)]$ by applying Lemma~\ref{lemma:ErrorRElemma2} and assumption (A1).

Finally, for $\bfW_3$, we apply Lemma 8 of \citet{RavikumarEtal11} on the (sub-gaussian) design matrix $\bfX$ to obtain that for sample size
\begin{align}\label{eqn:ErrorRElemma2eqn7}
n \geq 512 ( 1 + 4 \Lambda_{\max} (\Sigma_x^k))^4 \max_i (\sigma_{x,ii}^k )^4 \log (4p^{\tau_1})
\end{align}
we get that with probability $ \geq 1 - 1/p^{\tau_1-2}, \tau_1 > 2$,
$$
\left\| \frac{\bfX^T \bfX}{n} \right\|_\infty \leq \sqrt{ \frac{ \log 4 + \tau_1 \log p}{c_x n}} + \max_i \sigma_{x,ii} = V_x; \quad
c_x = \left[ 128 ( 1 + 4 \Lambda_{\max} (\Sigma_x)  )^2 \max_i (\sigma_{x,ii})^2 \right]^{-1}
$$
Thus, with the same probability,
\begin{align}\label{eqn:ErrorRElemma2eqn4}
\| \bfW_4 \|_\infty \leq \left\| \frac{\bfX^T \bfX}{n} \right\|_\infty \| \widehat \bfB - \bfB_0 \|_1^2 \| \bfT_{0,j} \|_1 
\leq 2 C_\beta^2 V_x \frac{ \log(pq)}{n}
\end{align}
We now bound the right hand side of (\ref{eqn:ErrorRElemma2maineqn}) using (\ref{eqn:ErrorRElemma2eqn5}), (\ref{eqn:ErrorRElemma2eqn6}) and (\ref{eqn:ErrorRElemma2eqn4}) to complete the proof, with the leading term of the sample size requirement being $n \succsim \log(pq)$.
\end{proof}

\begin{proof}[Proof of Lemma~\ref{thm:BetaThm}]
The proof follows that of part (I) of Lemma~\ref{thm:ThetaThm}, with a different group norm structure. We only point out the differences.

Putting $\bfbeta = \bfbeta_0$ in (\ref{eqn:EstEqn1}) we get
$$
-2 \widehat \bfbeta^T \widehat \bfgamma + \bfbeta^T \widehat \bfGamma \widehat \bfbeta + \lambda_n \sum_{h \in \cH} \| \widehat \bfbeta^{[h]}  \| \leq
-2 \bfbeta_0^T \widehat \bfgamma + \bfbeta_0^T \widehat \bfGamma \bfbeta_0 + \lambda_n \sum_{h \in \cH} \| \bfbeta_0^{[h]}  \|
$$
Denote $\bfb = \widehat \bfbeta - \bfbeta_0$. Then we have
$$
\bfb^T \widehat \bfGamma \bfb \leq 2 \bfb^T ( \widehat \bfgamma - \widehat \bfGamma \bfbeta_0 ) + \lambda_n
\sum_{h \in \cH} ( \| \bfbeta_0^{[h]} \| - \| \bfbeta_0^{[h]} + \bfb^{[h]} \|)
$$
Proceeding similarly as the proof of part (I) of Lemma~\ref{thm:ThetaThm}, with a different deviation bound and choice of $\lambda_n$, we get expressions equivalent to (\ref{eqn:Thm1ProofProp2Eqn3}) and (\ref{eqn:Thm1ProofProp2Eqn2}) respectively:
\begin{align}
\bfb^T \widehat \bfGamma \bfb & \leq \frac{3}{2} \sum_{h \in \cH} \| \bfb^{[h]} \| \\
\frac{\psi^*}{3} \| \bfb \|^2 & \leq \lambda_n \sum_{h \in \cH} \| \bfb^{[h]} \| \leq 4 \lambda_n \sqrt{B} \| \bfb \|
\end{align}
Furthermore, $\| \bfb \|_1 \leq \sqrt{ | h_{\max} |} \sum_{h \in \cH} \| \bfb^{[h]} \| $. The bounds in (\ref{eqn:BetaThmEqn1}), (\ref{eqn:BetaThmEqn2}), (\ref{eqn:BetaThmEqn3}) and (\ref{eqn:BetaThmEqn4}) now
follow.

\end{proof}

\begin{proof}[Proof of Lemma~\ref{prop:ThmBetaRE}]
For part 1 it is enough to prove that with $ \widehat \Sigma_x^k := (\bfX^k)^T \bfX^k/n$,
\begin{align}\label{eqn:ThmBetaREProofEqn1}
\widehat \bfT_k^2 \otimes \widehat \Sigma_x^k & \sim RE (\psi_*^k, \phi_*^k)
\end{align}
with high enough probability. because then we can take $\psi_* = \min_k \psi_*^k, \phi_* = \max_k \phi_*^k$. The proof of (\ref{eqn:ThmBetaREProofEqn1}) follows similar lines of the proof of Proposition 1 in \citet{LinEtal16}, only replacing $\Theta_\epsilon, \widehat \Theta_\epsilon, \bfX$ therein with $(\bfT^k)^2, (\widehat \bfT^k)^2, \bfX^k$, respectively. We omit the details.

Part 2 follows the proof of Proposition 2 in \citet{LinEtal16}.
\end{proof}

\begin{proof}[Proof of Lemma~\ref{Lemma:ThmTestingLemma}]
To show \eqref{eqn:ThmTestingProofeq3} we have
\begin{align*}
\frac{1}{\sqrt n \widehat s_i}  \widehat \Omega_y^{1/2} \bfE^T \bfR_i =
\frac{1}{\sqrt n \widehat s_i}  (\widehat \Omega_y^{1/2} - \Omega_y^{1/2}) \bfE^T \bfR_i +
\frac{1}{\sqrt n \widehat s_i}  \Omega_y^{1/2} \bfE^T \bfR_i
\end{align*}
The second summand is distributed as $\cN_q ({\bf 0}, \bfI)$. For the first summand,
\begin{align*}
\frac{1}{\sqrt n}  \left\| (\widehat \Omega_y^{1/2} - \Omega_y^{1/2}) \bfE^T \bfR_i \right\|_\infty & \leq
\frac{1}{\sqrt n}  \left\| \widehat \Omega_y^{1/2} - \Omega_y^{1/2} \right\|_\infty  \left\| \bfE^T \bfR_i \right\|_1 \\
& \leq \sqrt{n D_\Omega} \frac{1}{n} \left[ \| \bfE^T (\bfX_i -  \bfX_{-i} \bfzeta_i ) \|_1 + \| \bfE^T \bfX_{-i} (\widehat \bfzeta_i - \bfzeta_{i} ) \|_1 \right] \\
& \leq \sqrt{n D_\Omega} \frac{1}{n} \left[ \| \bfE^T \bfX_i \|_\infty + \| \bfE^T \bfX_{-i} \|_\infty
\left\{ \| \bfzeta_i  \|_1 + \| \widehat \bfzeta_i - \bfzeta_i  \|_1 \right\} \right] \notag\\
& \leq \sqrt{n D_\Omega} \left[ \frac{1}{n} \| \bfE^T \bfX_i \|_\infty + 
\frac{1 + D_\zeta}{n} \| \bfE^T \bfX_{-i} \|_\infty \right] \\
& \leq \sqrt{n D_\Omega} (2 + D_\zeta) .\frac{1}{n} \| \bfE^T \bfX \|_\infty
\end{align*}
because $\Omega_x$ is diagonally dominant implies $\| \bfzeta_i \|_1 = \sum_{i' \neq i} |\omega_{x,ii'}|/ \omega_{x,ii} \leq 1$, and using assumption (T1) and \eqref{eqn:omega-sqrt-bound}. Applying Lemma~\ref{lemma:ErrorRElemma2}, the following holds for $n \succsim \log (pq)$:
\begin{align}\label{eqn:ThmTestingProofeq31}
\frac{1}{\sqrt n}  \left\| (\widehat \Omega_y^{1/2} - \Omega_y^{1/2}) \bfE^T \bfR_i \right\|_\infty & \leq \sqrt{ D_\Omega} (2 + D_\zeta) c_2 [ \Lambda_{\max} (\Sigma_x) \Lambda_{\max} (\Sigma_e) ]^{1/2} \sqrt{ \log(pq)}
\end{align}
with probability $ \geq 1 - 6c_1 \exp [-(c_2^2-1) \log(pq)]$.

On the other hand,
\begin{align*}
s_i^2 := \frac{1}{n} \left\| \bfX_i - \bfX_{-i}  \bfzeta_{0,i} \right\|^2 & \leq 
\widehat s_i^2 + \frac{1}{n} \left\| \bfX_{-i} (\widehat \bfzeta_i - \bfzeta_{0,i} ) \right\|^2
\leq \widehat s_i^2 + \| \widehat \bfzeta_i - \bfzeta_{0i} \|_1^2 \left\| \frac{1}{n} \bfX_{-i}^T \bfX_{-i} \right\|_\infty
\end{align*}
which implies $s_i \leq \widehat s_i + D_\zeta \sqrt{ V_x}$. By applying Lemma 8 of \citet{RavikumarEtal11},
\begin{align}\label{eqn:ThmTestingProofeq32}
\left\| \frac{1}{n} \bfX_{-i}^T \bfX_{-i} \right\|_\infty \leq
\left\| \frac{1}{n} \bfX^T \bfX \right\|_\infty \leq V_x
\end{align}
with probability $ \geq 1 - 1/p^{\tau_1-2}, \tau_1>2$, and
\begin{align}\label{eqn:ThmTestingProofeq33}
n \geq 512 ( 1 + 4 \Lambda_{\max} (\Sigma_{x}))^4 \max_i (\sigma_{x,ii} )^4 \log (4p^{\tau_1})
\end{align}
On the other hand, by Chebyshev's inequality, for any $\epsilon>0$
$$
P\left( | s_i - \sqrt{ \sigma_{x,i,-i}} | \geq \epsilon \right) \leq \frac{Var (s_i)}{\epsilon^2} =
\frac{\kappa_i}{n \epsilon^2}
$$
Taking $\epsilon = n^{-1/4}$, we have $s_i \geq \sqrt{ \sigma_{x,i,-i}} - n^{-1/4}$ with probability $\geq 1 - \kappa_i n^{-1/2}$. Then, for $n$ satisfying \eqref{eqn:ThmTestingProofeq33} and $\sqrt{ \sigma_{x,i,-i}} - n^{-1/4} >  D_\zeta \sqrt{ V_x} $, we get the bound with the above probability:
\begin{align}\label{eqn:ThmTestingProofeq34}
\frac{1}{\widehat s_i} \leq \frac{1}{\sqrt{ \sigma_{x,i,-i}} - n^{-1/4} - D_\zeta \sqrt{ V_x}}
\end{align}
Combining \eqref{eqn:ThmTestingProofeq31} and \eqref{eqn:ThmTestingProofeq34} gives the upper bound for the right hand side of \eqref{eqn:ThmTestingProofeq3} with the requisite probability and sample size conditions.

To prove \eqref{eqn:ThmTestingProofeq4} we have
\begin{align}\label{eqn:ThmTestingProofeq41}
\frac{1}{n} \| \bfR_i^T \bfX_{-i} \|_\infty & \leq
\frac{1}{n} \| (\bfX_i - \bfX_{-i} \bfzeta_{0,i})^T \bfX_{-i} \|_\infty +
\frac{1}{n} \| \bfX_{-i}^T \bfX_{-i} ( \widehat \bfzeta_i - \bfzeta_{0,i}) \|_\infty
\end{align}
Applying Lemma~\ref{lemma:ErrorRElemma2}, for $n \succsim \log(p-1)$ we have
\begin{align}
\frac{1}{n} \| (\bfX_i -  \bfX_{-i} \bfzeta_i )^T \bfX_{-i} \|_\infty \leq 
c_{7} [ \sigma_{x,i,-i} \Lambda_{\max} (\Sigma_{x, -i}) ]^{1/2} \sqrt{\frac{ \log (p-1)}{n}}
\end{align}
with probability $\geq 1 - 6c_{6} \exp [-(c_{7}^2-1) \log (p-1)]$ for some $c_{6} >0, c_{7} > 1$. By \eqref{eqn:ThmTestingProofeq32}, the second term on the right side of \eqref{eqn:ThmTestingProofeq41} is bounded above by $D_\zeta V_x$ with probability $ \geq 1 - 1/p^{\tau_1-2}$ and $n$ satisfying \eqref{eqn:ThmTestingProofeq33}. The bound of \eqref{eqn:ThmTestingProofeq4} now follows by conditions (T2), (T3) and \eqref{eqn:ThmTestingProofeq34}. Since $\sqrt{ \sigma_{x,i,-i}} - n^{-1/4} >  D_\zeta \sqrt{ V_x} $ implies $\sqrt{ \sigma_{x,i,-i}} >  D_\zeta \sqrt{ V_x} $, and $D_\zeta = O(\sqrt{ \log p/n})$, the leading term of the overall sample size requirement is $n \succsim \log (pq)$.
\end{proof}

\begin{proof}[Proof of Lemma~\ref{Lemma:PowerThmLemma}] We drop $k$ in the superscripts. By definition,
\begin{align}
\frac{m_i}{\sqrt n} &= \frac{1}{\widehat s_i} \frac{ ( \bfX_i - \bfX_{-i} \widehat \bfzeta_i)^T \bfX_i}{n} \notag\\
&= \frac{1}{\widehat s_i} \left[ \frac{\| \bfX_i - \bfX_{-i} \widehat \bfzeta_i \|^2}{n} +
 \frac{(\bfX_i - \bfX_{-i} \widehat \bfzeta_i)^T \bfX_{-i} \widehat \bfzeta_i}{n} \right] \notag\\
& \leq \widehat s_i + \frac{1}{\widehat s_i} .\frac{1}{n} \| \bfR_i^T \bfX_{-i} \|_\infty
\left (\| \widehat \bfzeta_i - \bfzeta_{0i} \|_1 + \| \bfzeta_{0i} \|_1 \right) \notag\\
\Rightarrow \left| \frac{m_i}{\sqrt n} - \sqrt{\sigma_{x,i,-i}} \right| & \leq
| \widehat s_i - \sqrt{\sigma_{x,i,-i}} | + \frac{1}{\widehat s_i} .\frac{1}{n} \| \bfR_i^T \bfX_{-i} \|_\infty
\left (\| \widehat \bfzeta_i - \bfzeta_{i} \|_1 + \| \bfzeta_{i} \|_1 \right) \label{eqn:PowerThmLemmaProofEqn1}
\end{align}
By applying Lemma 8 in \citet{RavikumarEtal11}, we have a bound for the first summand on the right hand side:
$$
| \widehat s_i - \sqrt{\sigma_{x,i,-i}} | \leq \sqrt{ \frac{ \log 4 + \tau_2}{c_i n}}; \quad
c_i = \left[ 128 (1 + 4 \sigma_{x,i,-i})^2 \sigma_{x,i,-i}^2 \right]^{-1},
$$
with probability $1 - 1/p^{\tau_2 - 2}$ for some $\tau_2>2$, and $n \geq 512 (1 + 4 \sigma_{x,i,-i})^42 \sigma_{x,i,-i}^4 \log(4) $. For the second summand in the right-hand side of \eqref{eqn:PowerThmLemmaProofEqn1}, $1/\widehat s_i$ can be bounded using \eqref{eqn:ThmTestingProofeq34}, $(1/n) \| \bfR_i^T \bfX_{-i} \|_\infty$ can be bounded using derivations following \eqref{eqn:ThmTestingProofeq41}. Finally, $\| \widehat \bfzeta_i - \bfzeta_{i} \|_1 \leq D_\zeta$ from assumption (T1), and $\| \bfzeta_i \|_1 \leq 1$ because $\Omega_x$ is diagonally dominant and $|\zeta_{ii'}| = |\omega_{x,ii'}|/ \omega_{x,ii}$ for $i' \neq i$. The lemma now follows by putting everything back together in \eqref{eqn:PowerThmLemmaProofEqn1}.
\end{proof}

\begin{proof}[Proof of Lemma~\ref{lemma:omega-diff}]
$\| \bfA - \bfA_1 \|_\infty \leq \delta$ implies that $\bfA_1 + \delta \bfJ_a \geq \bfA$ and $\bfA + \delta \bfJ_a \geq \bfA_1$, where $\bfJ_a \in \BM(a,a)$ has all entries 1, and for positive definite matrices $\bfP, \bfQ$, $\bfP \geq \bfQ$ means $\bfP - \bfQ$ is positive definite. Now applying Theorem 1 part (a) in \citet{Bellman68} we have
$$
(\bfA + \delta \bfJ_a)^{1/2} \geq \bfA_1^{1/2}; \quad
(\bfA_1 + \delta \bfJ_a)^{1/2} \geq \bfA^{1/2}.
$$
Using the same result, it is easy to prove that
$$ \bfA^{1/2} + \sqrt \delta \bfJ_a \geq (\bfA + \delta \bfJ_a)^{1/2}, $$
and the same for $\bfA_1$. The lemma follows.
\end{proof}

\begin{proof}[Proof of Lemma~\ref{Lemma:PowerThmLemma2}]
We drop $k$ in the superscripts and 0 in subscripts. Note that it is enough to prove
$$
\left\| \frac{n \widehat \Sigma_y}{(m_i)^2} - \frac{ \Sigma_y}{\sigma_{x,i,-i}} \right\|_\infty = o_P(1).
$$
For this, consider the decomposition
\begin{align*}
\frac{n \widehat \Sigma_y}{(m_i)^2} &= \frac{\widehat \Sigma_y - \Sigma_y + \Sigma_y}{\sigma_{x,i,-i}}.
\frac{\sigma_{x,i,-i}}{(m_i)^2/n}\\
\Rightarrow \frac{n \widehat \Sigma_y}{(m_i)^2} - \frac{ \Sigma_y}{\sigma_{x,i,-i}} &=
\frac{\widehat \Sigma_y - \Sigma_y}{(m_i)^2/n} +
\frac{\Sigma_y}{\sigma_{x,i,-i}} \left[ 1 - \frac{\sigma_{x,i,-i}}{(m_i)^2/n} \right]\\
& = \frac{n}{(m_i)^2} \left[ \widehat \Sigma_y - \Sigma_y +
\frac{\Sigma_y}{\sigma_{x,i,-i}} \left( \frac{(m_i)^2}{n} - \sigma_{x,i,-i} \right) \right].
\end{align*}
From Lemma~\ref{Lemma:PowerThmLemma} we now have
$$
\frac{m_i}{\sqrt n} \geq \sqrt{\sigma_{x,i,-i}} - \delta_i \quad \Rightarrow \quad
\frac{m_i^2}{n} \geq (\sqrt{\sigma_{x,i,-i}} - \delta)^2 \geq \sigma_{x,i,-i} - \delta_i^2,
$$
so that
\begin{align}\label{eqn:PowerThmLemma2ProofEqn1}
\left\| \frac{n \widehat \Sigma_y}{(m_i)^2} - \frac{ \Sigma_y}{\sigma_{x,i,-i}} \right\|_\infty \leq
\frac{ \| \widehat \Sigma_y - \Sigma_y \|_\infty + \sigma_{x,i,-i}^{-1} \delta_i^2 \| \Sigma_y \|_\infty }{\sigma_{x,i,-i} - \delta_i^2},
\end{align}
with probability $\geq 1 - 6c_{6} \exp [-(c_{7}^2-1) \log (p-1)] - 1/p^{\tau_2-2} - \kappa_i / \sqrt n$ and for sample size satisfying $n \succsim \log p$,  $n \geq 512 (1 + 4 \sigma_{x,i,-i})^42 (\sigma_{x,i,-i})^4 \log(4) $ and $\sqrt{ \sigma_{x,i,-i}} > \max \{ \delta_i, n^{-1/4} - D_\zeta \sqrt{V_x} \}$. For the $\ell_\infty$ norms on the right-hand side, we have
\begin{align}\label{eqn:PowerThmLemma2ProofEqn2}
\| \Sigma_y \|_\infty = \| \Omega_y^{-1} \|_\infty \leq (\Delta_0 (\Omega_y))^{-1}
\end{align}
following \citet{Varah75}. For a bound on $\| \widehat \Sigma_y - \Sigma_y \|_\infty$, if condition (II) of Theorem~\ref{thm:PowerThm} is satisfied then we have
\begin{align}\label{eqn:PowerThmLemma2ProofEqn31}
\| \widehat \Sigma_y - \Sigma_y \|_\infty \leq \tilde D_\Omega
\end{align}
where $\tilde D_\Omega = O(D_\Omega)$ and $D_\Omega = O(\tilde D_\Omega)$ \citet{BickelLevina08}. If condition (I) is satisfied, denote $\epsilon = D_\Omega / \Delta_0 (\Omega_y)$. Then
\begin{align}
\| \widehat \Sigma_y - \Sigma_y \|_\infty &= \| \widehat \Sigma_y (\Omega_y - \widehat \Omega_y) \Sigma_y \|_\infty \notag\\
& \leq \| \widehat \Sigma_y \|_\infty \| \Omega_y - \widehat \Omega_y \|_\infty
\| \Sigma_y \|_\infty \notag\\
& \leq \| (\bfI + (\Omega_y - \widehat \Omega_y) \Sigma_y)^{-1}\|_\infty \| \Sigma_y \|_\infty \epsilon \notag\\
& \leq \frac{\epsilon}{\Delta_0( \Omega_y)}
\left[ 1 + \sum_{t=1}^\infty (\| (\Omega_y - \widehat \Omega_y) \Sigma_y \|_\infty)^t \right] \notag\\
& \leq \frac{\epsilon}{(1- \epsilon) \Delta_0( \Omega_y)} \notag\\
& = \frac{D_\Omega}{(\Delta_0 (\Omega_y) - D_\Omega ) \Delta_0 (\Omega_y)}\label{eqn:PowerThmLemma2ProofEqn32}
\end{align}
Combining \eqref{eqn:PowerThmLemma2ProofEqn2} with \eqref{eqn:PowerThmLemma2ProofEqn31} or \eqref{eqn:PowerThmLemma2ProofEqn32} as required and putting them back in the right-hand side of \eqref{eqn:PowerThmLemma2ProofEqn1}, we get the needed.
\end{proof}

\begin{proof}[Proof of Lemma~\ref{lemma:ErrorRElemma1}]
This is the same as in Lemma 2 in Appendix B of \citet{LinEtal16} and its proof can be found there.
\end{proof}

\begin{proof}[Proof of Lemma~\ref{lemma:ErrorRElemma2}]
This is a part of Lemma 3 of Appendix B in \citet{LinEtal16}, and is proved therein.
\end{proof}


\end{document}